\documentclass[10pt,twocolumn,letterpaper]{article}

\usepackage[pagenumbers]{cvpr} 

\definecolor{cvprblue}{rgb}{0.21,0.49,0.74}
\usepackage[pagebackref,breaklinks,colorlinks,citecolor=cvprblue]{hyperref}
\usepackage{booktabs}
\usepackage{multirow}
\usepackage{makecell}
\usepackage{comment}
\usepackage{colortbl}
\usepackage[accsupp]{axessibility} 
\usepackage{amssymb,enumitem}
\usepackage{dirtytalk}
\usepackage{longtable}
\usepackage{pifont}
\usepackage{float}
\usepackage{psfrag}
\usepackage[percent]{overpic}

\usepackage[utf8]{inputenc}

\usepackage{times}
\usepackage{epsfig}
\usepackage{graphicx}
\usepackage{subcaption}
\usepackage{caption}
\usepackage{amsmath}
\usepackage{amssymb}

\usepackage[usestackEOL]{stackengine}

\definecolor{cvprblue}{rgb}{0.21,0.49,0.74}
\usepackage[pagebackref,breaklinks,colorlinks,citecolor=cvprblue]{hyperref}

\usepackage[T1]{fontenc}    
\usepackage{url}            
\usepackage{booktabs}       
\usepackage{amsfonts}       
\usepackage{nicefrac}       
\usepackage{microtype}      
\usepackage{graphicx}
\usepackage{diagbox} 
\usepackage{siunitx}
\usepackage{enumitem}
\usepackage{arydshln} 
\usepackage{tabularx}
\usepackage{booktabs} 
\usepackage{xspace} 
\usepackage{pifont} 
\usepackage[dvipsnames]{xcolor} 
\usepackage{dsfont} 
\usepackage{bm}

\DeclareMathOperator{\diag}{diag}
\newcommand{\norm}[1]{\left\lVert#1\right\rVert}

\usepackage{stfloats}

\colorlet{colorFst}{red!30}         
\colorlet{colorSnd}{orange!30} 		
\colorlet{colorTrd}{yellow!30} 		

\colorlet{colorLow}{darkgray!30}    
\newcommand{\fst}{\cellcolor{colorFst}\bf}   
\newcommand{\nd}{\cellcolor{colorSnd}}      
\newcommand{\rd}{\cellcolor{colorTrd}}      

\newcommand{\ours}{DROID-Splat\xspace}

\definecolor{gray}{rgb}{0.65,0.65,0.65}
\definecolor{mycol}{rgb}{0.90,0.95,1.0}

\renewcommand\midrule{\specialrule{0.4pt}{0pt}{1pt}}

\def\paperID{16598} 
\def\confName{CVPR}
\def\confYear{2025}

\title{DROID-Splat \\ Combining end-to-end SLAM with 3D Gaussian Splatting}

\author{Christian Homeyer$^1$ \quad Leon Begiristain$^1$ \quad Christoph Schnörr$^1$\\
$^1$Image and Pattern Analysis Group, Heidelberg University, Germany \\
{\tt\small homeyer@math.uni-heidelberg.de},
{\tt\small lbegirri7@alumnes.ub.edu},
{\tt\small schnoerr@math.uni-heidelberg.de}
}

\begin{document}
\maketitle


 \begin{abstract}
 	Recent progress in scene synthesis makes standalone SLAM systems purely based on optimizing hyperprimitives with a Rendering objective possible \cite{monogs}. 
	However, the tracking performance still lacks behind traditional \cite{orbslam} and end-to-end SLAM systems \cite{droid}.
	An optimal trade-off between robustness, speed and accuracy has not yet been reached, especially for monocular video.
 	In this paper, we introduce a SLAM system based on an end-to-end Tracker and extend it with a Renderer based on recent 3D Gaussian Splatting techniques. 
 	Our framework \textbf{DroidSplat} achieves both SotA tracking and rendering results on common SLAM benchmarks. 
	We implemented multiple building blocks of modern SLAM systems to run in parallel, allowing for fast inference on common consumer GPU's. 
	Recent progress in monocular depth prediction and camera calibration allows our system to achieve strong results even on in-the-wild data without known camera intrinsics. 
 	Code will be available at \url{https://github.com/ChenHoy/DROID-Splat}.
 \end{abstract}
\section{Introduction}
Simultaneous Localization and Mapping (SLAM) has been a longstanding problem in Computer Vision, fundamental to applications in robotics, autonomous driving and augmented reality. 
While traditional systems focus on reconstruction of accurate odometry and geometry from hand-crafted features, they usually result in sparse or semi-dense representations of the environment. 
End-to-end SLAM systems \cite{droid, dpvo, dpvs} improved robustness and accuracy by using learned features and a dense reconstruction objective, however they often lack the ability to optimize a photo-realistic scene.
Recent progress in scene synthesis makes standalone SLAM systems purely based on optimizing hyperprimitives with a Rendering objective possible \cite{monogs}. 
However, the tracking performance still lacks behind traditional \cite{orbslam} and end-to-end SLAM systems \cite{droid}. 
We aim to close this gap by combining the best of both worlds. 
\begin{figure}[h!]
	\centering
	\includegraphics[width=1.0\linewidth]{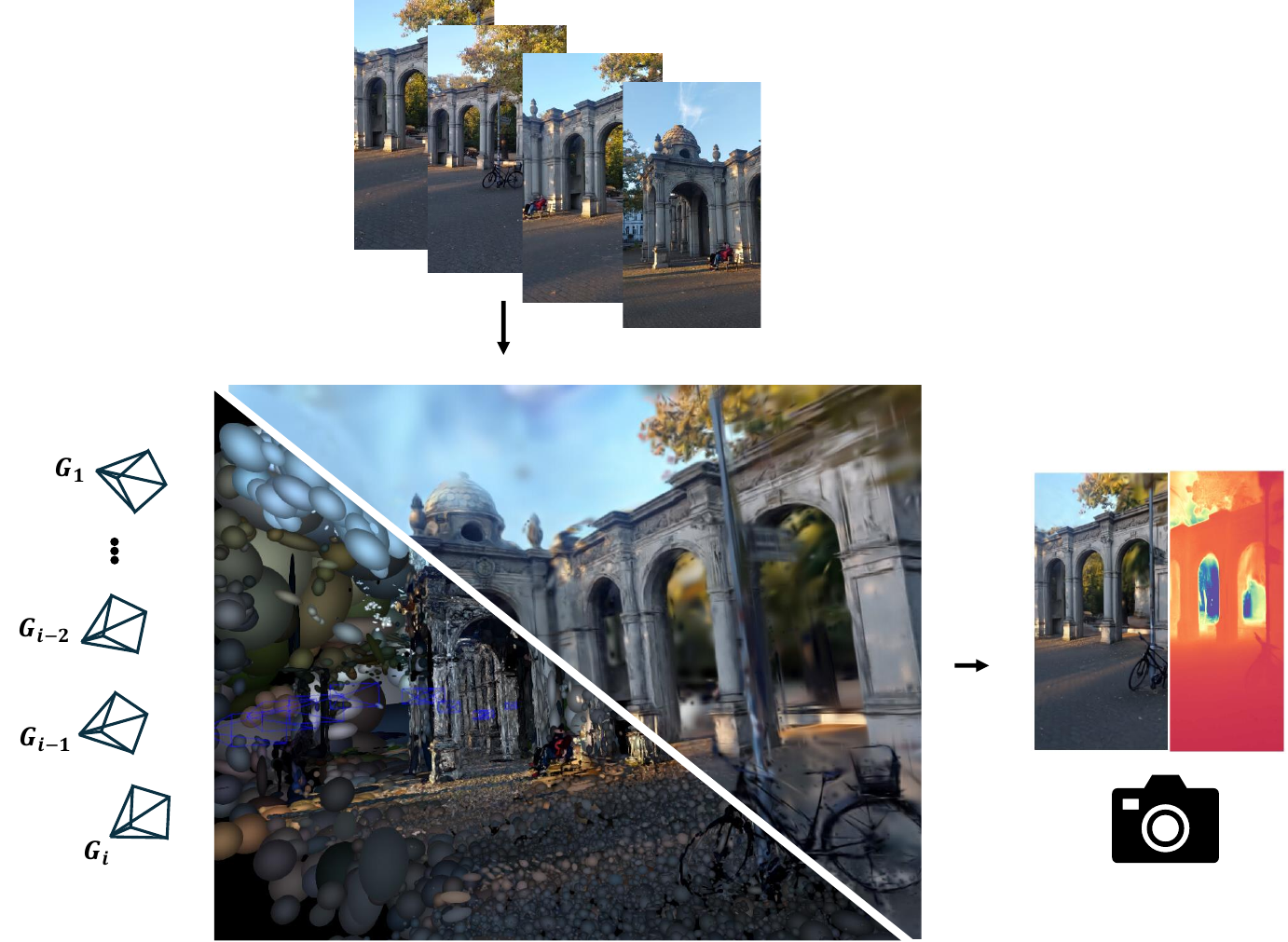}
	\caption{
		\textbf{DROID-Splat} allows to reconstruct a scene with known/unknown intrinsics. By combining an optical flow tracking objective and a fast, dense Renderer, we can achieve photo-realistic Reconstructions while optimizing accurate odometry.
	}
	\label{fig:eyecatcher}
\end{figure}

In this paper, we introduce DROID-Splat: A SotA SLAM system based on dense, end-to-end optical flow and a dense Rendering objective using 3D Gaussian Splatting \cite{3dgs}. 
Our system offers the same flexibility as it's parent system \cite{droid}: We support monocular and rgbd inference for different camera models 
(since we focus on single camera reconstruction, we neglect the stereo \cite{droid} or multi-view case \cite{r3d3}). 
By combining the best of both worlds, we achieve fast tracking inference on consumer GPU's and can quickly optimize a photo-realistic scene reconstruction. 
Our framework consists of a i) local frontend ii) global backend iii) loop closure detector iv) dense renderer. 
With this work we aim to systematically analyze the interplay of individual components and optimization objectives in more detail than previous work.
A lot of recent SLAM frameworks have emerged, that focus on a single component. Our work aims to provide a comprehensive tool, 
which allows to easily reconstruct a scene from a video.

Monocular Video is notoriously difficult to reconstruct. For this reason we additionally allow integration of SotA monocular depth prediction \cite{metric3d, zoedepth, depthanything} priors similar to \cite{hi-slam, glorieslam} and concurrent work \cite{splat-slam}. 
We show that with recent advances, it is possible to robustly handle in-the-wild data with unknown camera intrinsics. Using a depth prior and an additional camera calibration objective \cite{droidcalib}, we achieve strong reconstruction performance even on cellphone videos.

Our contributions are:
\begin{itemize}
	\item We propose a dense SLAM system, which combines a dense end-to-end tracker with dense hyperprimitives. 
	\item We combine common building blocks of modern SLAM systems in a fast parallel implementation. Our comprehensive ablations show which components really matter.
	\item We show SotA results on common SLAM benchmarks for both tracking and rendering in near real-time.
	\item Our framework is flexible with regards to input and works even on in-the-wild data with unknown intrinsics.
\end{itemize}

\section{Related Work}
\label{sec:related-work}
\textbf{Visual SLAM.} Traditional SLAM systems can be categorized into \textit{direct} or \textit{indirect} \cite{dso} systems depending on their intermediate representation and objective function. Indirect approaches \cite{orbslam} make use of sparse feature descriptors for matching and then solve a geometric bundle adjustment problem. Direct approaches \cite{dso, lsdslam} optimize a photometric error directly and operate on semi-dense pixel representations. However, direct approaches usually lead to more difficult optimization problems. Overcoming the limitation of both hand-crafted features and ill-behaved optimization, end-to-end SLAM systems \cite{droid, dpvo, dpvs} were proposed which allow a dense representation with well-behaved tracking. Long-term tracking requires a loop closure mechanism \cite{orbslam, dpvs, loopy-slam, loopsplat}. Common frameworks memoize features of past frames to find similarities of new incoming frames to start a loop closure optimization, e.g. a Pose Graph Optimization \cite{g2o}. 
On top of good odometry, common systems are concerned with a dense scene reconstruction. Traditional SLAM approaches either relied on voxel \cite{voxelhashing, infiniTAM, bundlefusion} or point \cite{badslam, keller2013real, elasticfusion} based map representations. These representations can allow a dense reconstruction, but are not photo-realistic.

\textbf{Differentiable rendering.} Neural Radiance Fields (NeRFs) \cite{nerf} opened the gate to achieve photorealistic volume rendering, but training was initially slow. Using a multi-resolution hash encoding \cite{instantngp} enabled for the first time to use neural radiance fields in a SLAM context \cite{nerf-slam}. Recently, 3D Gaussian Splatting (3DGS) \cite{3dgs} has revolutionized the field. The real-time rendering and training quickly enabled numerous works to taylor a \textit{direct} SLAM system based on the Rendering objective \cite{splatam, gs-slam, gaussian-slam, monogs}. However tracking remains to be behind the traditional counterparts. Combined hybrid systems \cite{photoslam, glorieslam, splat-slam} resolve this issue by combining the best of both worlds. Similar in fashion, we make use of a dense, robust end-to-end system \cite{droid} and combine it with a renderer. We refer to \cite{slam-survey} about more details on Rendering in SLAM.
  
Numerous works improve the original GS with new techniques \cite{mvsgs, mcmc, hf-slam, glorieslam, photoslam, 2dgs, gaussianopacityfields, mipsplatting, dnsplatter, pixelgs}, some including better densification and pruning strategies \cite{mipsplatting, gaussianopacityfields, mvsgs} or better loss supervision \cite{dnsplatter}. We aim to leverage recent advances in diff. rendering in a SLAM context. 

\textbf{Concurrent Work.} The most related work to us is \cite{splat-slam}, which similar to us, is based off DROID-SLAM \cite{droid} and 3DGS \cite{3dgs}. In the same manner, they make use of a previously proposed monocular prior integration \cite{hi-slam}. However, we go beyond the monocular use case and analyze different input modes, renderer and additional camera calibration \cite{droidcalib}. Moreover, we use a different loop closure mechanism and dive deeper into the interplay of tracker and renderer.

\section{Our Approach}
\label{sec:approach}
\begin{figure*}[h!]
	\centering
	\begin{overpic}[width=0.98\linewidth, height=8.5cm, tics=0, clip]
		{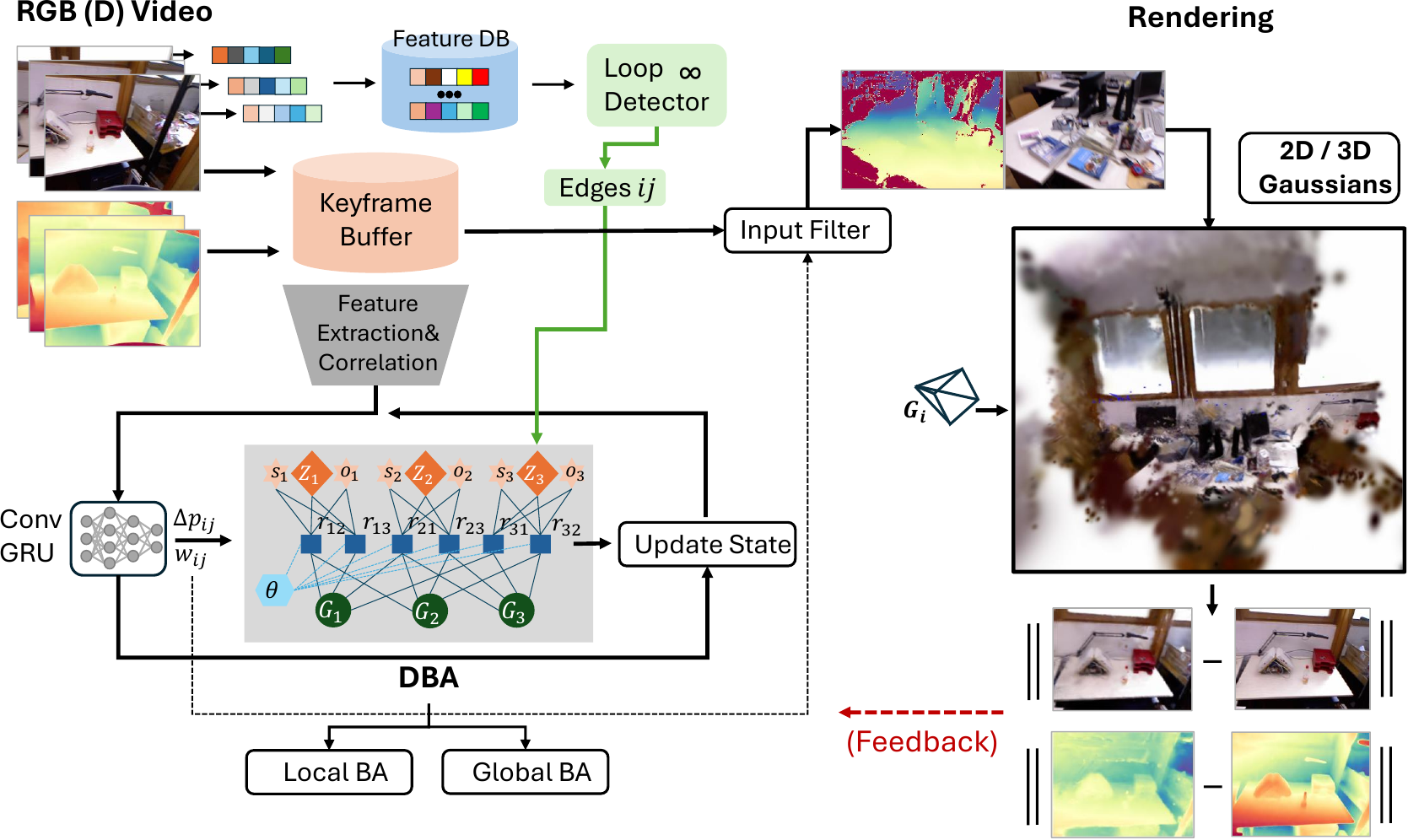}		
	\end{overpic}
	\caption{\textbf{DROID-Splat.} We make use of an end-to-end SLAM system with an optical flow based objective to perform tracking and reconstruct odometry and a dense initial map. The tracking objective is flexible, which allows us to optimize intrinsics or prior scale and shift as well if wanted. We make use of SotA Gaussian Splatting techniques to learn a photo-realistic reconstruction based on a Rendering objective. Since all components are differentiable and run in parallel, we can let parts interact flexibly.}
	\label{fig:framework}
\end{figure*}

Since our goal is a photo-realistic dense scene reconstruction, we use a dense end-to-end tracker, which can provide reliable depth (or disparity) for each pixel. After filtering this map for only covisible points or areas of high confidence, 
we feed it into a rendering module, which optimizes Gaussian hyperprimitives for each pixel and densifies the scene based on a rendering objective. Due to the lightweight nature of Gaussian Splatting \cite{3dgs}, we can run this rendering objective in real-time in parallel to our tracking system. An overview of our system can be seen in Figure \ref{fig:framework}. We systematically build up our system from common SLAM components. By unifying these techniques under one umbrella, we can reach state-of-the-art online photo-realistic reconstruction.

\subsection{End-to-end Tracking}
We base our tracker on online end-to-end system DROID-SLAM \cite{droid}. A frame-graph $\left(\mathcal{V}, \mathcal{E}\right)$ is build from an incoming ordered stream of images $ \{I_{1},\; \cdots,\; I_{n}\} \in \mathbb{R}^{N \times H\times W \times 3}$. This structure is in practice a keyframe buffer, storing our tracking state variables disparity $ d_{i} \in \mathbb{R}^{H\times W} $ and camera pose $g_{i} \in SE\left(3\right)$. Dense optical flow is estimated by a recurrent neural network \cite{raft}. Given enough motion in the scene, a keyframe is inserted into the graph. An edge $(i,\; j) $ signifies covisibility between the frames $ i $ and $ j $. As this graph is dynamically build and maintained over the incoming stream, we perform differentiable Bundle Adjustment over the graph. Given the current state of poses and disparity, we can compute correspondences 
\begin{equation}\label{projection}
	p_{ij} = \Pi_{c}\left( G_{ij} \circ \Pi_{c}^{-1}\left( p_{i},\; d_{i} \right) \right)
\end{equation} with the camera projection function $\Pi $. We use a pinhole camera model in all of our experiments, however similar to \cite{droidcalib}, we support multiple camera models in theory for this function. A correlation volume as in \cite{raft} can be indexed given $ p_{ij} $, so we retrieve correlation features along an edge $\left( i,\; j\right)$. The features, along with image context and a hidden state are input to a convolutional GRU to produce an update. The GRU produces i) the residual field $ r_{ij} \in \mathbb{R}^{H\times W\times 2} $ and an associated confidence $w_{ij}\in \mathbb{R}^{H\times W\times 2}$. The residuals guide the current correspondences as $p_{ij}^{*}= r_{ij}+ p_{ij}$. Together with the learned pose estimation confidence this powers a differentiable bundle adjustment optimization. Tracking is based on the reprojection based loss:  
\begin{align}\label{eqn:objective}
\mathbf E(\mathbf{G}', \mathbf{d}') &= \sum_{(i,j) \in \mathcal{E}} \norm{\mathbf{p}_{ij}^* - \Pi_c(\mathbf{G}'_{ij} \circ \Pi_c^{-1}(\mathbf{p}_i, \mathbf{d}'_i)) }_{\Sigma_{ij}}^2 \nonumber \\
\Sigma_{ij} &= \diag \mathbf{w}_{ij} \quad. 
\end{align}
This generic loss function can be used flexibly to not only supervise disparity $d'$ and pose $G'$, but as shown in \cite{droidcalib}, we can also directly optimize the calibration of the camera with intrinsics $\theta$:
\begin{align}\label{eqn:calib}
	\mathbf E(\mathbf{G}', \mathbf{d}', \mathbf{\theta}) &= \sum_{(i,j) \in \mathcal{E}} \norm{\mathbf{p}_{ij}^* - \Pi_c\left( \mathbf{G}'_{ij} \circ \mathbf{P}_{i}, \mathbf{\theta} \right) }_{\Sigma_{ij}}^2 \nonumber \\
	\text{with} \quad \mathbf{P}_{i} &= \Pi_c^{-1} \left( \mathbf{p}_i, \mathbf{d}'_i, \mathbf{\theta} \right) 
\end{align}

Now \cite{droid} supports \textit{RGBD}-SLAM by regularizing this with a prior term:
\begin{equation}\label{eqn:prior}
\mathbf E_{reg}\left(\mathbf{d}^{*}, \mathbf{d}'\right) = \sum_{i \in \mathcal{V}} \norm{ \mathbf{d}_{i}^{*} - \mathbf{d}'_{i} }^2
\end{equation} over a given input depth $\mathbf{d}^{*}$ from an external sensor. Since we want to reconstruct any video, we make use of monocular depth prediction priors \cite{metric3d, zoedepth, depthanything}. Even though monocular depth prediction has made progress to predict accurate metric predictions \cite{metric3d, zoedepth}, there is considerable temporal fluctuations across the board of SotA monocular models. For this reason we optimize in what we call the \textit{Pseudo-RGBD} mode, similar as \cite{hi-slam, glorieslam, splat-slam}:
\begin{equation}\label{eqn:mono_prior}
	\mathbf E_{reg,m}\left(\mathbf{d}^{*}, \mathbf{d}', \mathbf{s}', \mathbf{o}'\right) = \sum_{i \in \mathcal{V}} \norm{ \mathbf{d}_{i}^{*} - \left(\mathbf{s}'_{i} \cdot \mathbf{d}'_{i} + \mathbf{o}'_{i}\right) }^2
\end{equation}
After solving this bundle adjustment problem for a fixed number of iterations over the graph, we can update our state variables and continue until the next recurrence. In \textit{P-RGBD} mode, we must be careful as an ambiguity between $\mathbf{s},\mathbf{o} $ and $\mathbf{G}$ exists. For this reason like \cite{hi-slam}, we perform this in a block-coordinate descent manner, where we first fix scales and offsets and optimize poses. Afterwards we fix the pose graph and optimize structure, scales and offsets. We observe a similar ambiguity between intrinsics $\mathbf{\theta}$ and the monocular variables. For this reason we operate in two stages on in-the-wild video inspired by \cite{robudyn}: 1. Fix the prior and use Eq.\ref{eqn:calib} together with Eq. \ref{eqn:prior} to calibrate the camera. 2. Use the calibrated camera to run in \textit{P-RGBD} mode with Eq. \ref{eqn:mono_prior}. 

Modern SLAM systems \cite{lsdslam, droid, orbslam} perform bundle adjustment normally on different parts of the map: i) A local frontend optimizes small-scale graphs for incoming keyframe windows ii) A global backend optimizes large-scale graphs with long-term connections over the whole map. While the original implementation \cite{droid} performs this on two separate GPU's, we run both Processes on a single GPU and perform these two optimizations synchronized in paralell. Monocular prior integration is performed on local frontend windows before the adjusted map is put into the backend. Camera intrinsics $\mathbf{\theta}$ are treated as a global variable, that is optimized in the backend. 

\subsection{Loop Closure}
We observe, that Visual Odometry accuracy and robustness depends not only on the optimization itself, but in particular on the graph structure of front- and backend. Accumulated drift can be compensated by running the Update operator on long-term connections of potential loop candidates. While \cite{go-slam, splat-slam} detect candidates based on low apparent motion detected by the recurrent flow network \cite{droid}, we had more success by using direct visual similarity. While systems as \cite{loopy-slam, dpvs, orbslam} rely on hand-crafted ORB features \cite{rublee2011orb}, we leverage recent end-to-end features from place recognition tasks \cite{eigen-places}. For each incoming keyframe, we compute it's visual features and insert them in a FAISS \cite{faiss} database on the CPU. We then check for nearest neighbors in all past frames. Similar to \cite{hi-slam}, we only consider a frame pair $\left( i, j \right)$ a loop candidate if i) The feature distance is small enough $d_{f,ij} < \tau_{f}$ ii) The camera orientation distance is small enough $d\left(\mathbf{R}_{i}, \mathbf{R}_{j} \right) < \tau_{r}$ and iii) the frames are far apart enough $|t_{i} - t_{j}| > \tau_{t}$. If a candidate pair is found, we augment the graph by adding a bi-directional edge to the backend. This Process runs in parallel on the CPU with a marginal additional cost. 

\subsection{Differentiable Rendering}
Similar to previous works \cite{monogs, splat-slam, splatam, photoslam} we utilize Gaussian hyperprimitives defined as a set of points $\mathbf{P}\in \mathbb{R}^{3}$ associated to our dense tracking map. Each Gaussian possesses a rotation $\mathbf{r}\in SO\left(3\right)$, scaling $\mathbf{s}\in \mathbb{R}^{3}$, density $\mathbf{\sigma} \in \mathbb{R}^{1}$ and spherical harmonic coefficients $\mathbf{SH}\in \mathbb{R}^{16}$. We initialize the Gaussians similar to \cite{monogs} by downsampling the map by a constant factor after triangulation. Gaussians are optimized via backpropagation on a dense Rendering loss. The rendering process \cite{3dgs} is defined as:
\begin{equation}\label{eqn:rendering}
	C\left( \mathbf{R}, \mathbf{t} \right) = \sum_{i\in \mathcal{N}} \mathbf{c}_{i} \alpha_{i} \prod_{j=1}^{i-1} \left( 1 - \alpha_{i} \right),
\end{equation} where $\mathbf{c}$ denotes the color converted from $\mathbf{SH}$ and $\alpha_{i}=\sigma_{i}\cdot \mathcal{G}\left(\mathbf{R}, \mathbf{t}, \mathbf{P}_{i},\mathbf{r}_{i}, \mathbf{s}_{i} \right)$. This allows us to render our map at given keyframe $G_{i}$ to produce both an image $I'_{i}$ and depth $Z'_{i}$. We follow \cite{monogs} for median depth rendering. Gaussian Splatting \cite{3dgs, monogs, splatam} utilizes a mixed rendering loss 
\begin{align}\label{rnd_loss}
	L_{i} &= \lambda_{1} L_{rgb} + \left(1 - \lambda_{1} \right) L_{depth} \nonumber \\
	&= \lambda_{1} \left[ \left(1 - \lambda_{2}\right) \norm{I'_{i} - I^{*}_{i}} + \lambda_{2} \left( 1 - SSIM\left(I'_{i}, I^{*}_{i}\right) \right) \right] \nonumber \\ 
	&+ \left( 1 - \lambda_{1} \right) \norm{Z'_{i} - Z^{*}_{i}}
\end{align} 
which allows us to perform backpropagation by comparing with a reference $I^{*}, Z^{*}$. Each time we update our renderer, we optimize over a batch of cameras to improve our scene reconstruction. Since every component is differentiable, we can in theory optimize our keyframe poses with the rendering objective and feed them back into the tracker. We therefore want to research the questions: \textit{Which objective is better suited for tracking? Can we improve our system further by finetuning with a dense rendering objetive?}\newline Since we only improve the map by covering the whole 3D space with Gaussians, the original adaptive density control \cite{3dgs} strategy splits and clones Gaussians based on their size and gradient. This strategy was also used in any succesful SLAM application \cite{photoslam, monogs,splat-slam, splatam,gaussian-slam}. It was recently observed, that this strategy is suboptimal and by guiding this process with a Monte Carlo Chain Markov (MCMC) model \cite{mcmc}, we can improve performance. At the same time, this provides a preset upper limit of the total number of primitives. We compare these different strategies for our system and compare the 3D hyperprimitives themselves with the recently proposed 2D surfel Gaussians \cite{2dgs}. 2D Gaussian Splatting approximates surfaces by collapsing the primitives to flat surface disks, which result in more accurate geometry. 

\section{Experiments}
\label{seq:experiments}
We combine our components in a flexible way and ablate these choices in the following. During inference, we synchronize frontend, backend and renderer based on fixed frequencies, 
i.e. we run backend and renderer for every $k_{1}$, $k_{2}$ calls of the leading frontend process. The loop detector is constantly run in the background. If we detect a large tracking 
map update, we record the rel. transformations to reanchor our hyperprimitives. With mostly a stable map, we simply use a rigid body transformation $G \in SE\left(3\right)$ for this purpose. 
Gaussians are then typically in a position where they will quickly reconverge upon a new rendering optimization. For our monocular experiments with a prior, we use Metric3D \cite{metric3d} as it gave the most temporally consistent predictions without any scale optimization. We ablate this choice in the supplementary against multiple SotA models. We run our system on a NVIDIA RTX 4090. Similar to \cite{monogs, splat-slam} we do a refinement stage 
after running online tracking. We refine our map for 2k iterations and report the refined results for our final numbers. As benchmark metrics generally favor slower methods for this task, we report the detailed scaling of speed and performance in 
Fig. \ref{fig:speed}. We give more details on our system configuration, loss balancing and experiment settings in the supplementary.

\textbf{Datasets.}
We evaluate our method  on common SLAM benchmarks Replica \cite{replica} and TUM-RGBD \cite{tum-rgbd}. We additionally showcase the ability of our system on self-recorded outdoor cellphone video.

\begin{table}[h!]
	\centering
	\scriptsize
	\setlength{\extrarowheight}{2mm} 
	\resizebox{0.95\linewidth}{!}
	{
	\begin{tabularx}{\linewidth}{lcccc}
		\toprule
		\multirow{2}{*}{Components} & \makecell[c]{ATE RMSE \\ \textit{KF} [cm]  } & \makecell[c]{ATE RMSE \\ \textit{All} [cm] } & \makecell[c]{ATE RMSE \\ \textit{KF} [cm] } & \makecell[c]{ATE RMSE \\ \textit{All} [cm]} \\
		& \multicolumn{2}{c}{\cellcolor[HTML]{EEEEEE}{TUM RGBD}} & \multicolumn{2}{c}{\cellcolor[HTML]{EEEEEE}{Replica}} \\ 
		\midrule
		\makecell[l]{Frontend +\\ Backend} & 4.88 & 5.22 & 2.51 & 2.47 \\
		\makecell[l]{+ scale opt.} & 1.92 & 1.80 & 0.273 & 0.273 \\
		\makecell[l]{+ Loop Detection} & \textbf{1.88} & \textbf{1.78} & \textbf{0.269} & \textbf{0.268} \\
		[0.5pt] \hdashline \noalign{\vskip 1pt}
		\makecell[l]{w Loop BA \cite{go-slam} \\Backend} & 3.91 & 3.61 & 0.53 & 0.52 \\		
		\bottomrule
	\end{tabularx}
	}
	\caption{\textbf{Ablation Tracking}. We compare in \textit{P-RGBD} mode, but observe that this is mostly consistent across input modes. Optimizing the scale of priors \cite{metric3d} is still important 
		even when they are metric predictions. Using visual cues to find loop edges improves tracking. 
	}
	\label{tab:tracker}
\end{table}
\begin{table}[htb]
	\centering
	\scriptsize
	\setlength{\extrarowheight}{1mm} 
	\resizebox{0.95\linewidth}{!}
	{
	\begin{tabularx}{\linewidth}{lcccccc}
		\toprule
		Technique & PSNR$\uparrow$ & LPIPS$\downarrow$ & L1$\downarrow$ & PSNR$\uparrow$ & LPIPS$\downarrow$ & L1$\downarrow$ \\
		& \multicolumn{3}{c}{\cellcolor[HTML]{EEEEEE}{\textit{KF}}} & \multicolumn{3}{c}{\cellcolor[HTML]{EEEEEE}{\textit{Non-KF}}} \\ 
		\midrule
		3DGS \cite{3dgs} & 23.25 & 0.228 & 0.089 & 22.49 & 0.244 & 0.089 \\
		\makecell[l]{+ Covis. \\ Pruning \cite{monogs}} & 23.26 & 0.227 & 0.091 & 22.46 & 0.245 & 0.092 \\
		[0.5pt] \hdashline \noalign{\vskip 1pt}
		\makecell[l]{MCMC \cite{mcmc}}  & \textbf{23.80} & \textbf{0.211} & \textbf{0.082} & \textbf{22.84} & \textbf{0.232} & 0.0843 \\ 
		\makecell[l]{+ Covis. \\ Pruning \cite{monogs}} & 23.78 & 0.214 & 0.82 & 22.81 & 0.234 & \textbf{0.0841} \\ 
		[0.5pt] \hdashline \noalign{\vskip 1pt}
		\makecell[l]{2DGS \cite{2dgs}} & 20.67 & 0.313 & 0.103 & 19.822 & 0.329 & 0.103 \\
		\makecell[l]{+ Covis. \\ Pruning \cite{monogs}} & 20.71 & 0.310 & 0.102 & 19.838 & 0.329 & 0.103 \\ 
		\bottomrule
	\end{tabularx}
	}
	\caption{\textbf{Ablation Rendering}. We compare a selection of recent advancements in rendering within our framework. Results are averaged over TUM RGBD \cite{tum-rgbd}, as this is one of the most challenging benchmarks. We test this in \textit{P-RGBD} mode. See the supplementary for more techniques we have tried.
	}
	\label{tab:renderer}
\end{table}
\textbf{Baselines.} We compare ourselves to SotA pure Splatting based SLAM frameworks \cite{monogs, gaussian-slam, splatam} 
and hybrid systems \cite{photoslam, glorieslam, splat-slam} like ours. Finally, we also compare to systems based on volume rendering \cite{go-slam} or NeRF's \cite{nerf-slam, q-slam}. 

\textbf{Evaluation Metrics.} For Rendering we report PSNR, SSIM \cite{ssim} and LPIPS \cite{lpips} on the rendered keyframe images 
against the groundtruth. For geometry, we compare the rendered depth L1 $[cm]$ error to the groundtruth sensor as in \cite{nerf-slam}. In case of monocular depth, we compare the scale-aligned depth \cite{zoedepth}. For tracking we compare the ATE RMSE $[cm]$ \cite{tum-rgbd} error on the estimated trajectory. Reported results are averaged over 5 runs for statistical significance.

\textbf{Tracking Ablation.}
Table \ref{tab:tracker} shows the importance of individual tracking components in \textit{P-RGBD} mode. These results are mostly consistent across input modes and datasets, see supplementary. We make the observation that the factor graph building process is of most importance. 

Integrating monocular priors with the scale optimization is crucial. We did not have success with the Loop BA proposed in \cite{go-slam} in our Backend. Instead, we achieve the best results when adding visually similar loop candidates into our graph. 
We also want to highlight, that we achieve SotA results by simply utilizing a more conservative graph building strategy. See the supplementary for more details. 

\begin{table}[h]
	\centering
	\scriptsize
	\begin{tabularx}{\linewidth}{lcccccc}
		\toprule
		Method & \makecell[c]{\texttt{fr1/} \\ \texttt{desk}} & \makecell[c]{\texttt{fr2/} \\ \texttt{xyz}} & \makecell[c]{\texttt{f3/} \\\texttt{off}}  & \makecell[c]{\texttt{f1/} \\\texttt{desk2}} & \makecell[c]{\texttt{f1/} \\\texttt{room}} & \textbf{Avg.} \\
		\midrule
		\multicolumn{7}{l}{\cellcolor[HTML]{EEEEEE}{\textit{\textbf{Mono}}}} \\ 
		DPV-SLAM \cite{dpvs} & 1.8 & 1.0 & - & \rd 2.9 & 9.6 &  \\ 
		GlORIE-SLAM \cite{glorieslam} & \nd 1.6 & \fst 0.2 & \rd 1.4 & \nd 2.8 & \nd 4.2 & \nd 2.1 \\
		GO-SLAM \cite{go-slam} & \nd 1.6 & \nd 0.6 & 1.5 & \nd 2.8 & \rd 5.2 & \rd 2.3 \\
		MonoGS \cite{monogs} & \rd 4.2 & 4.8 & 4.4 & - & - &  \\
		MoD-SLAM \cite{mod-slam} & \fst 1.5 & \rd 0.7 & \fst 1.1 & - & - & \\
		Photo-SLAM \cite{photoslam} & \fst 1.54 & 0.98 & \nd 1.26 & - & - & \\
		Ours Mono & \nd 1.6 & \fst 0.2 & 1.6 & 8.3 & 5.8 & 3.5 \\
		Ours P-RGBD & \nd 1.6 & \fst 0.2 & 1.7 & \fst 2.3 & \fst 3.3 & \fst 1.8 \\
		\midrule
		\multicolumn{7}{l}{\cellcolor[HTML]{EEEEEE}{\textit{\textbf{RGBD}}}} \\ 
		SplaTAM \cite{splatam} & 3.4 & \rd 1.2 & 5.2 & 6.5 & 11.1 & 5.5 \\
		GS-SLAM \cite{gaussian-slam} & \fst 1.5 & 1.6 & 1.7 & - & - & \\
		GO-SLAM \cite{go-slam} & \fst 1.5 & \nd 0.6 & \nd 1.3 & 2.8 & 5.2 & 2.3 \\ 
		Photo-SLAM \cite{photoslam} & \rd 2.6 & \fst 0.35 & \fst 1.0 & - & - & \\
		Ours RGBD & \nd 1.6 & 1.4 & \rd 1.4 & \fst 2.2 & \fst 2.7 & \fst 1.9 \\
		\bottomrule
	\end{tabularx}
	\caption{\textbf{Tracking Performance TUM-RGBD \cite{tum-rgbd}} (ATE RMSE$\downarrow$ [cm]). Results are from respective papers. We achieve SotA tracking with our framework. Best results are highlighted as \colorbox{colorFst}{\bf first}, \colorbox{colorSnd}{second}, \colorbox{colorTrd}{third}.
	}
	\label{tab:sota_tracker}
\end{table}

\textbf{Renderer Ablation.}
Table \ref{tab:renderer} shows an ablation of recently proposed Gaussian Splatting techniques. For this ablation, we compare results without a refinement stage. We detail additional experiments in the supplementary. We want to highlight, that common comparisons should always factor in the total number of Gaussians used. We use 120k Gaussians on average on TUM-RGBD. Of course, using more primitives will improve photo-realism at the cost of memory and compute. We observe, that the covisibility pruning from \cite{monogs} is not necessarily effective on indoor datasets. Naive map building can perform better, however at the price of a few thousand more Gaussians. By far the most effective improvement in our experiments is the MCMC guided densification strategy \cite{mcmc}, which gives a consistent boost in rendering metrics compared to the naive gradient based densification strategy \cite{3dgs}. In order to make a fair comparison we match the total number of Gaussians to be equal with both strategies. 2D Gaussian Splatting turned out to be ineffective on very cluttered indoor scenes such as TUM-RGBD, either over-smoothing details or not building correct surfaces. We also observe, that optimizing geometry typically comes at the cost of worse rendering performance. We can control this trade-off by tuning $\lambda_{1}$.

\begin{table*}[h]
	\centering    
	\scriptsize
	\resizebox{\linewidth}{!}
	{
		\begin{tabular}{lcccccccccccc}
			\toprule
			Metric & \makecell[c]{NeRf-\\SLAM~\cite{nerf-slam}} & \makecell[c]{GO-\\SLAM~\cite{go-slam}} & \makecell[c]{NICER-\\SLAM~\cite{nicer-slam}} & \makecell[c]{MoD-\\SLAM~\cite{mod-slam}} &  
			\makecell[c]{Photo-\\SLAM~\cite{photoslam}}&  \makecell[c]{Mono-\\GS~\cite{monogs}} & \makecell[c]{GlORIE-\\SLAM~\cite{glorieslam}} & \makecell[c]{Q-SLAM\\~\cite{q-slam}}
			& \makecell[c]{Splat-\\SLAM$\textcolor{red}{^*}$~\cite{splat-slam}} & \makecell[c]{\textbf{Ours}\\mono} & \makecell[c]{\textbf{Ours}\\P-RGBD} \\
			\midrule
			PSNR $\uparrow$    					& \fst 41.40 & 22.13 & 25.41 & 27.31 & 33.30 & 31.22 & 31.04 & 32.49 & 36.45 & \rd 39.47 & \nd 39.66 \\
			SSIM $\uparrow$   					& - &  0.73 & 0.83  & 0.85  & 0.93  & 0.91 & \rd 0.97 & 0.89 & \nd 0.95 & \fst 1.0 & \fst 1.0 \\
			LPIPS $\downarrow$ 					& - & - & 0.19 & - & - & 0.21 & \rd 0.12 & 0.17 & \nd 0.06 & \fst 0.03 & \fst 0.03 \\  
			L1 $\downarrow$						& 4.49 & 4.39 & - & \rd 3.23 & - & 27.24 & - & \nd 2.76 & \fst 2.41 & 3.33 & 3.34 \\ 
			\midrule
			\makecell[l]{ATE RMSE} $\downarrow$	& - & \rd 0.39 & 1.88 & \nd 0.35 & 1.09 & 14.54 & \nd 0.35 & - & \nd 0.35 & \fst 0.27 & \fst 0.27 \\  
			\bottomrule
		\end{tabular}
	}
	\caption{
		\textbf{Avg. Rendering and Tracking Results on Replica~\cite{replica}} for monocular methods. Results are from~\cite{slam-survey} and the respective papers.  
	}
	\label{tab:replica_mono_comp}
\end{table*}

\begin{table*}[h]
	\centering    
	\scriptsize
	\resizebox{0.9\linewidth}{!}
	{
		\begin{tabular}{lcccccccccc}
			\toprule
			Metric & \makecell[c]{Vox-\\Fusion~\cite{vox-fusion}} & \makecell[c]{NICE-\\SLAM~\cite{nice-slam}} 
			& \makecell[c]{Mono-\\GS~\cite{monogs}} & \makecell[c]{Point-\\SLAM~\cite{point-slam}} & \makecell[c]{SplatAM\\~\cite{splatam}} & \makecell[c]{Gaussian\\SLAM~\cite{gaussian-slam}}
			& \makecell[c]{Photo-\\SLAM~\cite{photoslam}} & \makecell[c]{GO-\\SLAM~\cite{go-slam}} & \makebox[0.07\linewidth]{\textbf{Ours}} \\
			\midrule
			PSNR $\uparrow$    					& 24.42 & 26.16 & \rd 38.94 & 35.17 & 34.11 & \fst 42.08 & 34.96 & - & \nd 39.66 \\
			SSIM $\uparrow$   					& 0.81 & 0.83 & 0.97 & \nd 0.98 & \rd 0.97 & \fst 1.0 & 0.94 & - & \fst 1.0 \\
			LPIPS $\downarrow$ 					& 0.23 & 0.23 & \rd 0.07 & 0.12 & 0.10 & \fst 0.02 & 0.06 & - & \nd 0.03 \\  
			L1 $\downarrow$						& - & - & - & - & -  & - & - & \nd 3.38 & \fst 0.55 \\
			\midrule
			\makecell[l]{ATE RMSE} $\downarrow$	& 3.09 & 2.35 & \rd 0.32$^*$/0.58 & 0.53 & 0.36 & \nd 0.31 & 0.60 & 0.34 & \fst 0.29 \\  
			\bottomrule
		\end{tabular}
	}
	\caption{
		\textbf{Avg. Rendering and Tracking Results on Replica~\cite{replica}} for RGB-D methods. Results are from the respective papers.$^*$result comes from the slower single-process implementation.
	}
	\label{tab:replica_rgbd_comp}
\end{table*}
\begin{figure*}[b]
	\vspace{0em}
	\centering
	{
		\setlength{\tabcolsep}{1pt}
		\renewcommand{\arraystretch}{1}
		\newcommand{\sz}{0.2}
		\newcommand{\subsz}{0.2}
		\begin{tabular}{ccccc}
			\Large
			& GlORIE-SLAM~\cite{glorieslam}  & MonoGS~\cite{monogs} & \ours (Ours)  & Ground Truth \\[-1pt]
			\raisebox{0.6cm}{\rotatebox{90}{\makecell{\texttt{fr3}\\\texttt{office}}}} &
			\includegraphics[width=\sz\linewidth, trim=10 10 10 10, clip, height=2.5cm]{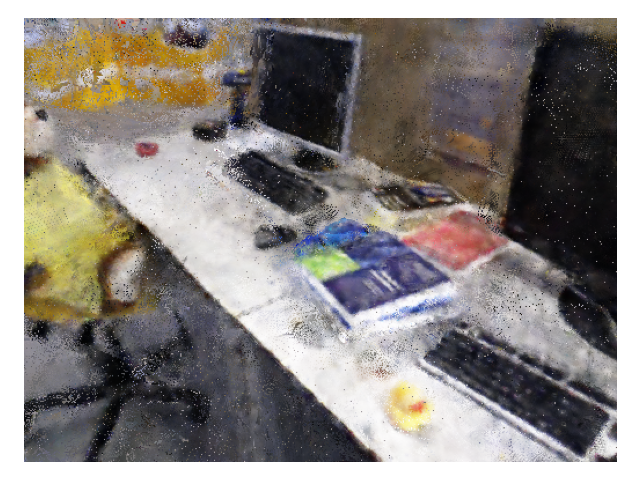} &
			\includegraphics[width=\sz\linewidth, trim=15 10 15 10, height=2.5cm]{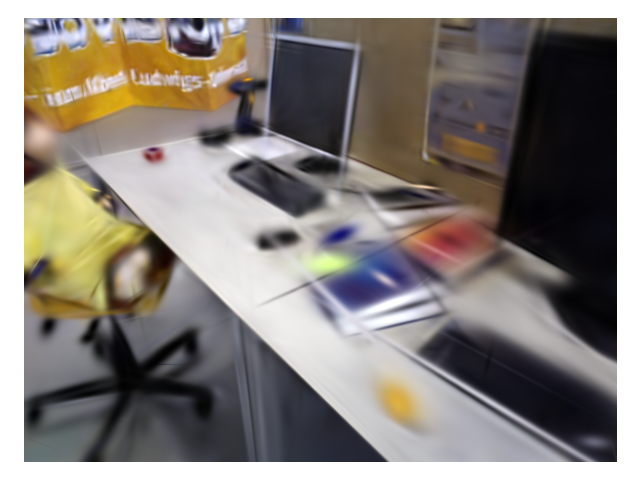} &
			\includegraphics[width=\sz\linewidth, trim=10 10 10 10, clip, height=2.5cm]{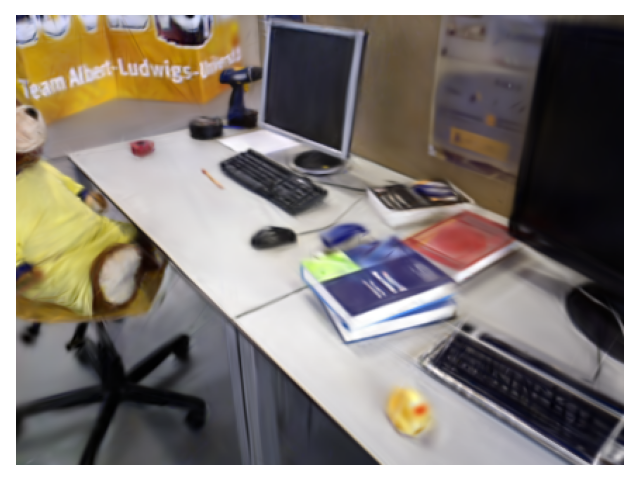} &
			\includegraphics[width=\sz\linewidth, height=2.5cm]{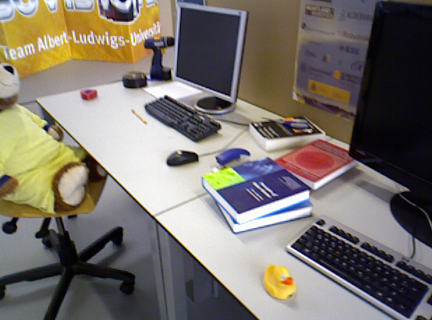} \\ 
			
			& \includegraphics[width=\sz\linewidth, trim=10 12 10 12, clip, height=2.5cm]{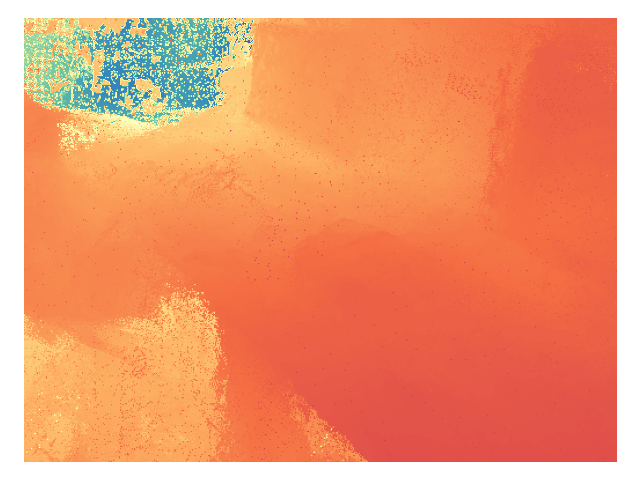} &
			\includegraphics[width=\sz\linewidth, trim=15 12 15 12, height=2.5cm]{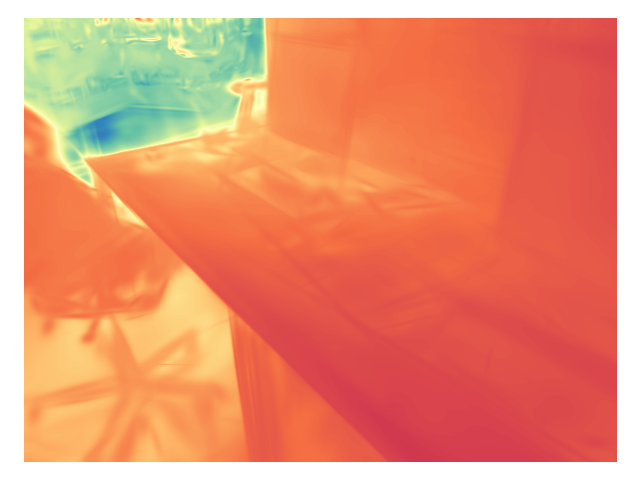} &
			\includegraphics[width=\sz\linewidth, trim=10 10 10 10, clip, height=2.5cm]{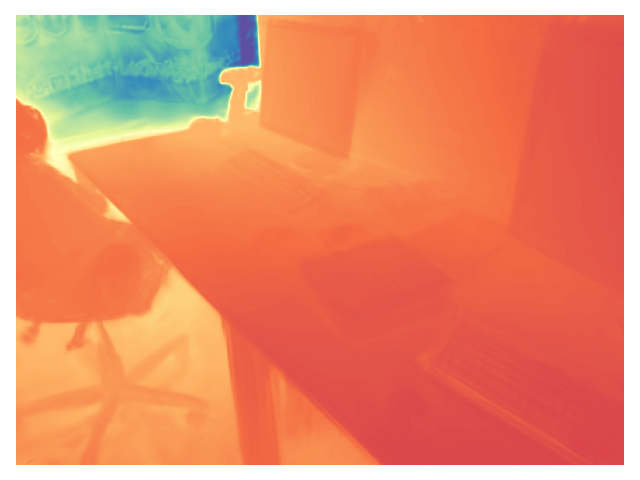} &
			\includegraphics[width=\sz\linewidth, height=2.5cm]{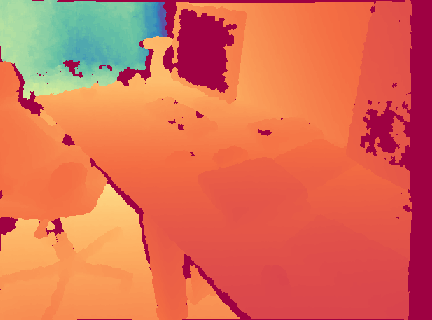} \\
			
			\\[0.05cm]
			
			\raisebox{0.8cm}{\rotatebox{90}{\makecell{\texttt{fr1}\\\texttt{desk}}}}&
			\includegraphics[width=\sz\linewidth, height=2.5cm]{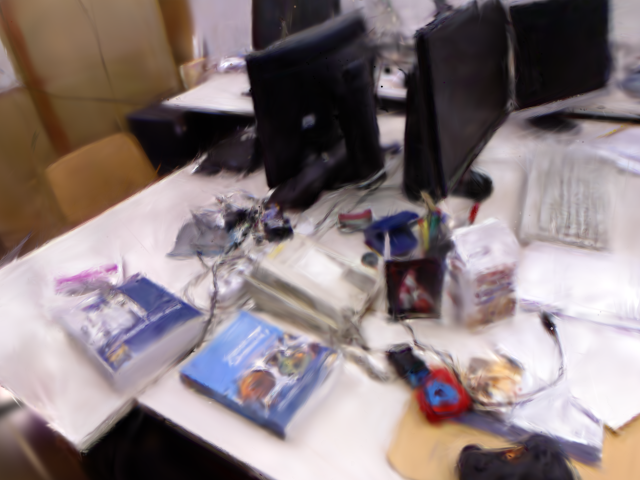} &
			\includegraphics[width=\sz\linewidth, trim=15 12 15 12, height=2.5cm]{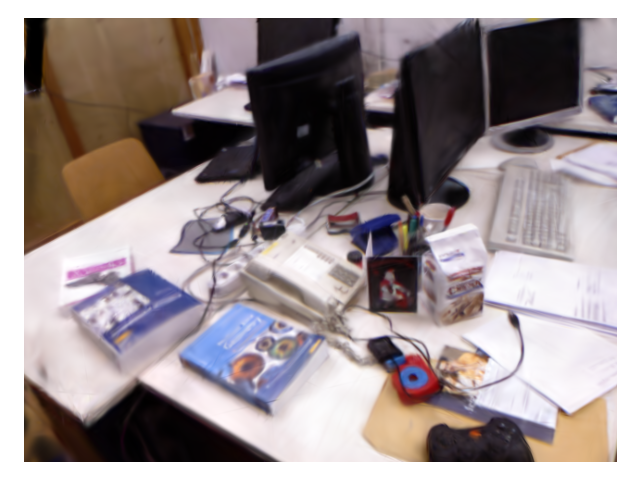} &
			\includegraphics[width=\sz\linewidth, trim=5 10 5 10, clip, height=2.5cm]{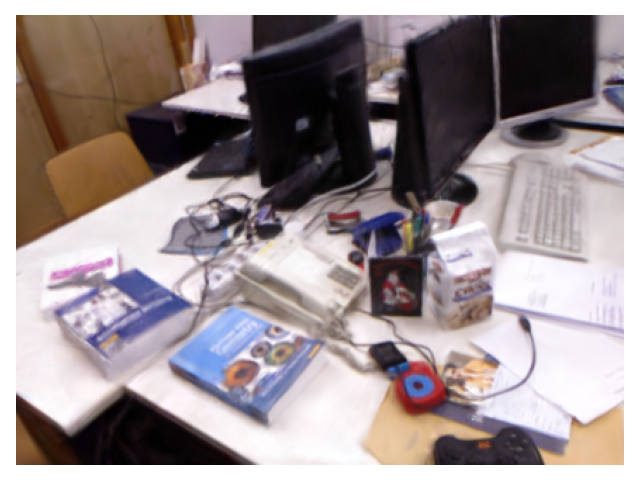} & 
			\includegraphics[width=\sz\linewidth, height=2.5cm]{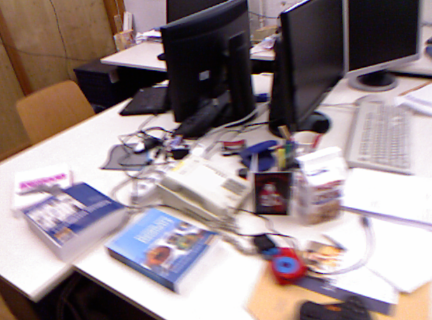} \\
			&
			\includegraphics[width=\sz\linewidth, height=2.5cm]{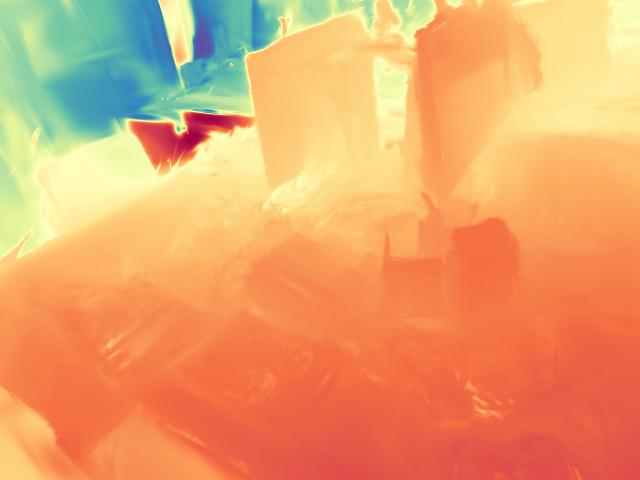} &
			\includegraphics[width=\sz\linewidth, trim=15 12 15 12, height=2.5cm]{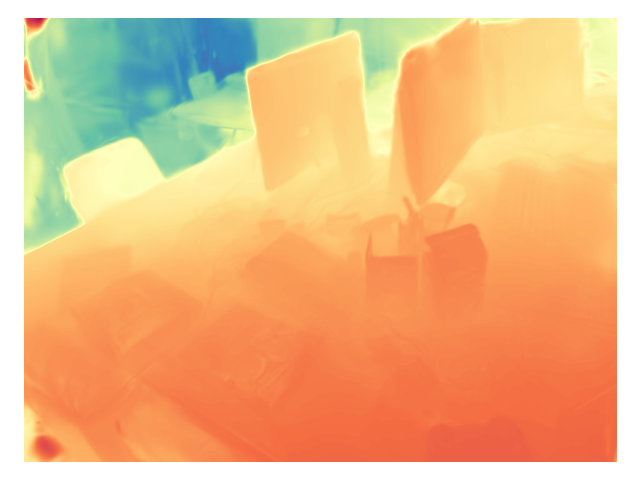} &
			\includegraphics[width=\sz\linewidth, trim=5 10 5 10, clip, height=2.5cm]{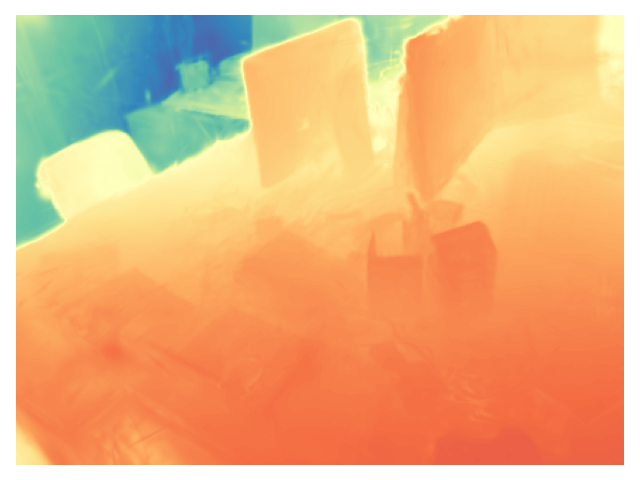} &
			\includegraphics[width=\sz\linewidth, height=2.5cm]{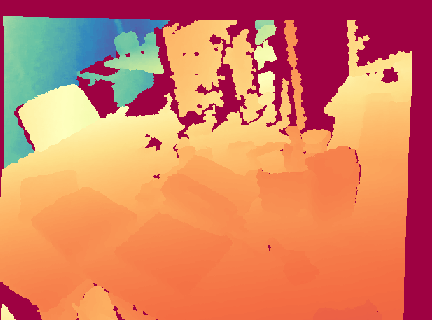} \\
			
			\Large
			& Photo-SLAM~\cite{photoslam} & MonoGS~\cite{monogs} & \ours (Ours)  & Ground Truth \\[-4pt]
			& & & &
		\end{tabular}
	}
	\caption{\textbf{Rendering Results on TUM-RGBD~\cite{tum-rgbd}.} We show views, that were not in the training set, i.e. our keyframe buffer. Top two rows show \textit{monocular} methods, bottom shows \textit{RGBD} (We show the results with prior for ours). We achieve a higher rendering and depth quality than \cite{photoslam, glorieslam, monogs} due to initializing with a dense tracking system and using dense hyperprimitives. Using a monocular prior can even improve upon a sparse laser sensor.}
	\label{fig:render_tum}
	\vspace{0em}
\end{figure*}

\begin{table*}[h!]
	\centering
	\scriptsize
	\setlength{\tabcolsep}{11.16pt}
	\resizebox{0.9\linewidth}{!}
	{
	\begin{tabularx}{0.85\linewidth}{llcccccc}
		\toprule
		Method & Metric  & \texttt{f1/desk} & \texttt{f2/xyz} & \texttt{f3/off}  & \texttt{f1/desk2} & \texttt{f1/room}& \textbf{Avg.}\\
		\midrule
		\multicolumn{8}{l}{\cellcolor[HTML]{EEEEEE}{\textit{\textbf{RGB-D}}}} \\ 
		\multirow{3}{*}{\makecell[l]{SplaTaM~\cite{splatam}}} 
		&PSNR$\uparrow$ & \rd 22.00 & \rd 24.50 & 21.90 & - & - & \\
		&SSIM $\uparrow$ & \rd 0.86 & \nd 0.95 & \rd 0.88 & - & - & \\
		&LPIPS $\downarrow$ & \rd 0.23 & \nd 0.10 & 0.20 & - & - & \\
		[0.8pt] \hdashline \noalign{\vskip 1pt}
		\multirow{3}{*}{\makecell[l]{Gaussian-SLAM~\cite{gaussian-slam}}} 
		&PSNR$\uparrow$ & \nd 24.01 & \nd 25.02 & 26.13 & \nd 23.15 & \nd 22.98 & \nd 24.26\\
		&SSIM $\uparrow$ & \nd 0.92 & \rd 0.92 & \nd 0.94 & \nd 0.91 & \nd 0.89 & \nd 0.92 \\
		&LPIPS $\downarrow$ & \nd 0.18 & 0.19 & \nd 0.14 & \nd 0.20 & \nd 0.24 & \nd 0.19\\
		[0.8pt] \hdashline \noalign{\vskip 1pt}
		\multirow{3}{*}{\makecell[l]{Photo-SLAM~\cite{photoslam}}} 
		&PSNR$\uparrow$ & 20.87 & 22.09 & 22.74 & - & - &  \\
		&SSIM $\uparrow$ & 0.74 & 0.77 & 0.78 & - & - &  \\
		&LPIPS $\downarrow$ & 0.24 & \rd 0.17 & \rd 0.15 & - & - & \\
		[0.8pt] \hdashline \noalign{\vskip 1pt}
		\multirow{3}{*}{\makecell[l]{\textbf{Ours}}} 
		&PSNR$\uparrow$ & \fst 26.45 & \fst 28.45 & \fst 27.83 & \fst 25.13 & \fst  26.16 & \fst  26.81 \\
		&SSIM $\uparrow$ & \fst 0.99 & \fst 0.99 & \fst 0.99 & \fst  0.99 & \fst  0.99 & \fst  0.99 \\
		&LPIPS $\downarrow$ & \fst 0.12 & \fst 0.07 & \fst 0.10 & \fst 0.15 & \fst  0.14 & \fst  0.12 \\
		
		\midrule
		\multicolumn{8}{l}{\cellcolor[HTML]{EEEEEE}{\textit{\textbf{Mono}}}} \\ 
		\multirow{3}{*}{\makecell[l]{Photo-SLAM~\cite{photoslam}}} 
		&PSNR$\uparrow$& 20.97 & 21.07&  19.59& - & - & \\
		&SSIM $\uparrow$& 0.74 & \rd 0.73 & 0.69 & - & - &  \\
		&LPIPS $\downarrow$& \rd 0.23 &  \rd 0.17 & 0.24 & - & - & \\
		[0.8pt] \hdashline \noalign{\vskip 1pt}
		
		\multirow{3}{*}{\makecell[l]{MonoGS~\cite{monogs}}} 
		&PSNR$\uparrow$& 19.67 &  16.17 &  20.63 & 19.16& 18.41 &  18.81\\
		&SSIM $\uparrow$& 0.73 &  0.72 & 0.77 &  0.66 &  0.64 & 0.70\\
		&LPIPS $\downarrow$& 0.33 & 0.31 & 0.34 & 0.48 & 0.51 &  0.39\\
		[0.8pt] \hdashline \noalign{\vskip 1pt}
		
		\multirow{3}{*}{\makecell[l]{GlORIE-SLAM~\cite{glorieslam}}} 
		&PSNR$\uparrow$& 20.26 &  25.62 & 21.21 & 19.09 & 18.78 & 20.99\\
		&SSIM $\uparrow$& \rd 0.79 & 0.72 & 0.72 & \nd 0.92 & 0.73 & 0.77\\
		&LPIPS $\downarrow$&  0.31 & 0.09 & 0.32 & 0.38 & 0.38 & 0.30\\
		[0.8pt] \hdashline \noalign{\vskip 1pt}
		
		\multirow{3}{*}{\makecell[l]{Splat-SLAM$\textcolor{red}{^*}$~\cite{splat-slam}}}
		&PSNR$\uparrow$ & \rd 25.61 & \fst 29.53 & \rd 26.05 &  \rd 23.98 & \rd 24.06 & \rd 25.85\\
		&SSIM $\uparrow$ & \nd 0.84 & \nd 0.90 & \nd 0.84 & \rd 0.81 & \nd 0.80 & \nd 0.84\\
		&LPIPS $\downarrow$ & \nd 0.18 & \nd 0.08 & \rd 0.20 & \rd 0.23 & \rd 0.24 &  \rd 0.19\\
		[0.8pt] \hdashline \noalign{\vskip 1pt}
		
		\multirow{3}{*}{\makecell[l]{\textbf{Ours Mono}}}
		&PSNR$\uparrow$ & \fst 26.72 & \nd 29.35 & \fst 27.92 & \nd 24.58 & \fst 25.64 & \fst 26.84 \\
		&SSIM $\uparrow$ & \fst 0.99 & \fst 0.99 & \fst 0.99 & \fst 0.99 & \fst 0.99 & \fst 0.99 \\
		&LPIPS $\downarrow$ & \fst 0.12 & \fst 0.07 & \fst 0.11 & \nd 0.18 & \fst 0.17 & \fst 0.13 \\
		[0.8pt] \hdashline \noalign{\vskip 1pt}
		\multirow{3}{*}{\makecell[l]{\textbf{Ours P-RGBD}}}
		&PSNR$\uparrow$ & \nd 26.42 & \rd 28.08 & \nd 27.84 & \fst 25.21 & \nd 25.11 & \nd 26.53 \\
		&SSIM $\uparrow$ & \fst 0.99 & \fst 0.99 & \fst 0.99 & \fst 0.99 & \fst 0.99 & \fst 0.99 \\
		&LPIPS $\downarrow$ & \fst 0.12 & \nd 0.08 & \fst 0.11 & \fst 0.17 & \nd 0.18 & \fst 0.13 \\

		\bottomrule
	\end{tabularx}
	}
	\caption{\textbf{Rendering Performance on TUM-RGBD~\cite{tum-rgbd}.} Numbers are from respective papers. Our method performs competitively or better than related \textit{RGB-D} methods with monocular input. We achieve SotA results across all modes.}
	\label{tab:render_tum}
\end{table*}

\subsection{Comparison with the State-of-the-Art}
We evaluate on synthetic and real-world scenes and compare with the State-of-the-Art. As can be seen in Table \ref{tab:sota_tracker}, we achieve competitive tracking performance on real world scenes. We want to highlight, that mostly \textit{fr1/desk2} and \textit{fr1/room} are challenging and therefore account biggest in the average. The performance of most frameworks seems similar on easier ones. We are also SotA on Replica across different modes in Table \ref{tab:replica_mono_comp} and \ref{tab:replica_rgbd_comp}. It can be seen, that traditional and end-to-end tracking systems are still the best once perfect supervision is missing. However with perfect synthetic data, direct methods achieve strong results. Monocular methods, that utilize a depth prior \cite{glorieslam, mod-slam, splat-slam} generally perform better on rendering and tracking due to the extra information. Our method consistently ranks across the highest in rendering due to the dense representation both in tracker and renderer. Even though Photo-SLAM \cite{photoslam} utilizes a robust tracking system \cite{orbslam}, the sparse hyperprimitive optimization does not allow indistinguishable renders. Figure \ref{fig:render_tum} shows rendered images and depth maps. We achieve highly detailed geometry on monocular video. Our monocular prior provides dense guidance even when laser sensors have holes. Table \ref{tab:render_tum} showcases the SotA on TUM-RGBD. We consistently achieve strong rendering metrics even on more challenging scenes like \textit{fr1/room}. 
We provide a more detailed overlook in the supplementary with an evaluation protocol of non-training frames. We observe in our experiments, that the benefit of both monocular and sensor depth priors can be observed mainly in the $L1$ reconstruction metric and on non-training frames. The current evaluation protocol rewards overfitting the scene. We also want to highlight from Table \ref{tab:replica_rgbd_comp}, that the synthetic Replica \cite{replica} benchmark is saturated in \textit{RGBD} mode. Predicted images and groundtruth are already indistinguishable at a PSNR of $\geq 39 [dB]$. The same can be said about the predicted depths at L1 $\leq 0.6 [cm]$. We did not achieve the same geometry quality on Replica as related work \cite{splat-slam, mod-slam, q-slam}, however the comparison is not fair since the number of hyperprimitives and inference time was not published in addition. We believe, that with more refinement iterations, different input filter thresholds and loss hyperparameters, we could reach the same metrics. More details can be found in the supplementary.

\textbf{Can our renderer improve tracking?} We can feedback outputs of our rendering pipeline back into the tracking system and verify this idea on Replica. We can backpropagate gradients from rendering into our pose graph by finetuning poses during the render updates. We then use the finetuned poses in our next tracking update during bundle adjustment. 
\begin{table}[h]
	\centering
	\scriptsize
	\begin{tabularx}{0.96\linewidth}{lccccc}
		\toprule
		\multirow{2}{*}{Feedback} & \makecell[c]{ATE RMSE \\ \textit{KF} [cm]  } & \makecell[c]{ATE RMSE \\ \textit{All} [cm] } & PSNR$\uparrow$ & LPIPS$\downarrow$ & L1$\downarrow$ \\
		& \multicolumn{5}{c}{\cellcolor[HTML]{EEEEEE}{\textbf{Replica}}} \\ 
		\multicolumn{6}{l}{\textbf{RGBD}} \\
		\midrule
		None & 0.293 & 0.289 & 36.03 & 0.06 & 0.0076 \\
		Disparity & 0.294 & 0.304 & 35.99 & 0.06 & 0.0076 \\
		Poses & \textbf{0.277} & \textbf{0.273} & \textbf{36.26} & 0.06 & \textbf{0.0074} \\
		Both & 0.28 & 0.289 & 36.19 & 0.06 & 0.0075 \\
		
		\multicolumn{6}{l}{\textbf{P-RGBD}} \\
		\midrule
		None & \textbf{0.269} & \textbf{0.268} & 32.92 & 0.134 & \textbf{0.0374} \\
		Poses & 0.356 & 0.348 & \textbf{32.95} & 0.134 & 0.0394 \\
		Both & 0.34 & 0.349 & 32.9 & 0.135 & 0.04 \\
		& \multicolumn{5}{c}{\cellcolor[HTML]{EEEEEE}{\textbf{TUM RGBD}}} \\
		
		\multicolumn{6}{l}{\textbf{RGBD}} \\
		\midrule
		None & \textbf{1.94} & \textbf{1.87} & \textbf{23.76} & \textbf{0.194} & \textbf{0.054} \\
		Disparity & 1.99 & 1.92 & 23.65 & 0.199 & 0.055 \\
		Poses & 1.98 & 1.89 & 23.72 & 0.197 & 0.056 \\
		Both & 2.0 & 2.12 & 23.61 & 0.199 & 0.056 \\
		\bottomrule
	\end{tabularx}
	\caption{\textbf{Feedback}. Since all components are differentiable, we can finetune the pose graph based on the rendering objective and feed this back into the tracking system. This requires a well initialized rendering system and good dense groundtruth supervision. We did not achieve stable results on more noisy monocular data. We compare the average result across the dataset over 5 runs without refinement.
	}
	\label{tab:feedback}
\end{table}

We can also go one step further and use the rendering objective to densify the disparity state of the tracker. In practice we perform a check to make sure that the difference between rendered depth and tracker disparity is not too large in order to not confuse the update network. Table \ref{tab:feedback} shows different variations of this experiment in both \textit{RGBD} mode with perfect groundtruth and in \textit{P-RGBD} mode. We observe, that this actually works as long as perfect groundtruth is available. Results worsen when using a monocular prior or trying this idea on real data. We do not report monocular experiments on TUM-RGBD, because the stability of the tracker was severely affected.

\begin{figure}[t]
	\centering
	\includegraphics[width=0.9\linewidth]{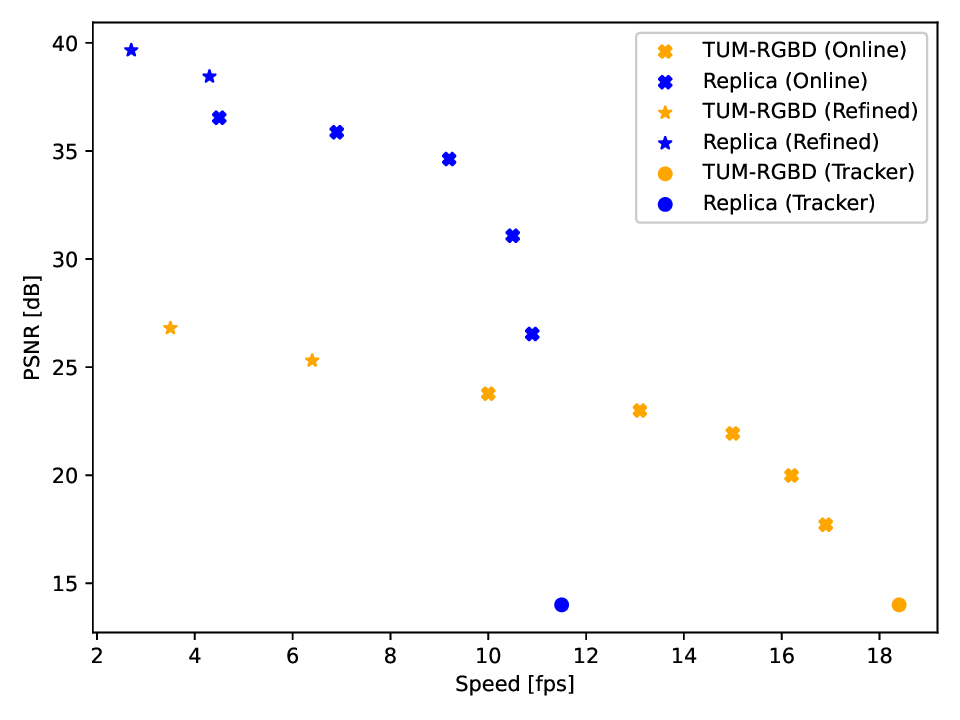}
	\caption{
		\textbf{Compute-Performance Trade-off}. We take the average across TUM RGBD \cite{tum-rgbd} and Replica \cite{replica} in \textit{RGBD} mode. 
		We added the baseline Tracker at the bottom for perspective, this does not have a meaningful Metric attached to it.
	}
	\label{fig:speed}
\end{figure}

\textbf{Runtime analysis.} The rendering performance heavily depends on how much compute is spend on optimizing the Gaussian primitives. We can give a scaling curve by adjusting the optimization iterations of the renderer or the frequency with which we use it. We also distinguish between online mode and offline refinement and give a baseline speed of our standalone tracker in Figure \ref{fig:speed}. We report results on the rendered keyframes. We want to note that in case of monocular depth priors, we are bottlenecked by the inference speed of the depth prediction network.

\begin{figure}[h]
	\centering
	\includegraphics[width=1.0\linewidth, trim=0 10 0 10, clip]{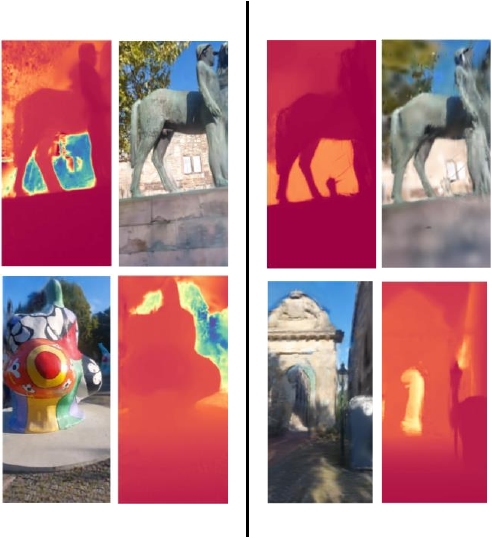}
	\caption{
		\textbf{Results on hand-captured cellphone videos.} In-the-wild outdoor scenes pose different challenges than benchmarks. \newline \textit{Left}: 3D Gaussian Splatting. 
		\textit{Right}: 2D Gaussian Splatting. 
		While 2DGS is more resistant to floaters due to its surface optimization, it struggles with rendering quality. 
		Both methods cannot deal well with strong lighting changes and reflections without extensions. 
	}
	\label{fig:wild}
\end{figure}

\textbf{In-the-wild Reconstruction.} For in the-wild reconstruction, we tested both 3DGS \cite{3dgs} and 2DGS \cite{2dgs} on challenging outdoor scenes. We qualitatively analyzed both methods on self-recorded videos. The difference in rendering quality seems to prevail on unbounded outdoor scenes with no 360-degree camera trajectory. Due to the challenging lighting conditions and much more unreliable monocular depth priors, we accumulate many floater Gaussians. 2D Gaussian Splatting does not suffer as much from these, as it generates smooth surfaces. See Figure \ref{fig:wild} for a visual comparison. 

\textbf{Failure Cases.} Our method fails to handle challenging lighting changes and lens flares without additional modifications. In general, we perform much worse in sparser scenarios or when our priors are unreliable. We also observe, that although our tracking system is robust, 
it can still drift on much more challenging scenes and trajectories. Since we optimize our hyperprimitives in batches, 
we are prune to catastrophic forgetting as other methods and need to reoptimize.

\section{Conclusion}
\label{sec:conclusion}
We combined a dense end-to-end SLAM system with a photo-realistic renderer. We systematically ablated common design choices and achieve SotA results with our framework on common benchmarks. 
The integration of recent monocular depth priors allowed to close the gap between monocular and RGBD SLAM both for odometry and rendering. Our experiments show, that photorealistic rendering and accurate geometry 
can be complementary objectives at this level, where improving rendering performance comes at a cost of worse geometry. At the same time, we did not see an improvement of our tracker based 
on the rendering objective for natural scenes. Our framework is flexible and can seamlessly reconstruct even in-the-wild video with unknown intrinsics.

\textbf{Outlook.} We hope that our Python framework enables rapid experimentation and further research in combining neural networks and SLAM. 
Recent foundation models \cite{geolrm} allow to infer 3D scenes from images directly without test-time optimization. The integration of such models poses an exciting avenue for future research. Extending the system to larger complex scenes would be another interesting direction.
\paragraph{ACKNOWLEDGMENT.}
We would like to say a special thank you to Andrei Prioteasa for helping setup the experiments and our colleagues at the Institute of Applied Mathematics, Heidelberg for fruitful discussions. 

{\small
	\bibliographystyle{ieeenat_fullname}
	\bibliography{draft.bib}
}

\clearpage
\setcounter{section}{0} 
\section*{Supplementary Material}
In this supplementary material, we provide more details on our approach, experiment settings and further experimental results. We encourage readers to take a look at our open-sourced implementation upon publication for more detailed configuration. All experiments are performed on a desktop computer with an \textit{AMD Ryzen Threadripper PRO 5955WX} CPU and \textit{NVIDIA RTX4090} GPU.

\section{Inference settings and hyperparameters}
\label{sup:settings}
We run our system at resolution $320\times 432$ on TUM-RGBD~\cite{tum-rgbd} and $360\times 640$ on Replica~\cite{replica}. 

\subsection{Tracking}
Tracking is configured by the \textit{frontend} and \textit{backend} parameters for graph building, optimization and our loop detector. Since the configuration system is quite complex, we will only brush across the critical parameter settings. We will release full configurations upon publication.

\paragraph{Frontend.}
We use a motion threshold of $ 3.0 $ for adding new keyframes. During scale optimization we use the objective
\begin{equation}\label{eq:reg_weight}
	E = E + \alpha E_{reg}
\end{equation} with $\alpha=0.001 $. We found it important to keep keyframes longer in the frontend bundle adjustment window. This can be controlled with the \textit{max age} variable in \cite{droid}. We increase this value from $25$ to $ 30 $. We use a weight $\beta = 0.5 $ on TUM-RGBD~\cite{tum-rgbd} and $\beta=0.7$ on Replica~\cite{replica} for measuring frame distance, see \cite{droid}.

\paragraph{Backend.}
We run the backend every $ 8 $ frontend passes in our experiments. We build our global graph more conservatively by using a window of max. $ 150 $ frames, using up to $ 1500 $ edges. 

\paragraph{Loop Detector.}
We compute visual features by using the EigenPlaces \cite{eigen-places} ResNet50 network. We found qualitatively, that a feature threshold $\tau_{f} = 0.5$ works well in practice. We make the assumption that loop candidates are at least $ \tau_{t} = 10$ frames apart. For our orientation threshold, we set $\tau_{r}=15^{\circ}$. This assumes that during a loop closure we have a very similar orientation, but due to drift a very distinct translation. 

\subsection{Rendering}
We run our mapper every $20$ frontend calls and optimize for $100 $ iterations at a small delay of 5 frames. We found that in practice, this can be arbitrarily tuned, i.e. we could also run more frequent with less training iterations. We anneal a 3D positional learning rate between $\left[1e-4, 1e-6\right]$ during our optimization, the other parameters are similar to \cite{monogs}. During each iteration, we optimize newly added frames and add the $ 10 $ last frames and $ 20 $ random past frames simmilar to \cite{monogs}. Since we run a test-time optimization, we are also prone to catastrophic forgetting. Using enough random frames ensures that this does not happen. Our system is not yet designed to handle large-scale scenes or unbounded scenes, where smarter strategies may be needed. 

After filtering the tracking map with a covisibility check \cite{droid}, we downsample the point cloud with factor $64$ on Replica~\cite{replica} and 16 on TUM-RGBD~\cite{tum-rgbd}.
We use the same thresholds for densification across datasets. We made the experience that balancing these parameters can result in similar results as long as the total number of Gaussians is similar. We weight $L1$-error and $SSIM$ \cite{ssim} in our appearance loss with $\lambda_{2} = 0.2 $. We balance depth and appearance supervision with $\lambda_{1} = 0.9$ on TUM-RGBD~\cite{tum-rgbd} and $0.8$ on Replica~\cite{replica}. Balancing these two terms, can shift metrics slightly in favor of either appearance or geometry. Similar to \cite{monogs} we encourage isotropic Gaussians with their scale regularization term. Our officially reported metrics are for dense depth supervision when a prior exists. For monocular video, we use the filtered tracking map as depth guidance.

\paragraph{Monte Carlo Markov Chain Gaussian Splatting.}
MCMC Gaussian Splatting \cite{mcmc} has additional hyperparameters for the noise level and the max. number of Gaussians in a scene. This poses an upper limit beyond which cannot be densified. We use $lr_{noise}=1e4$ and use a slightly lower number of Gaussians from the runs with vanilla 3D Gaussian Splatting \cite{3dgs}.

\paragraph{2D Gaussian Splatting}
2D Gaussian Splatting has a slightly different objective function \cite{2dgs}. On top of the default rendering objective, we also have a normal consistency $L_{normal}$ and depth distortion loss $L_{dist}$. We also found, that this representation has a different learning dynamic than 3D Gaussians. We therefore tuned the weighting to the best of our ability (without extensive parameter sweeps).  

\subsection{Feedback}
In our feedback experiment, we perform backpropagation on the local pose graph of our rendering batch as is done in \cite{monogs}. We used vanilla 3D Gaussian Splatting with the adaptive density control densification strategy \cite{3dgs}. As a sanity check, we only feedback the poses and/or disparity of rendered frames that have a decently similar disparity to the tracking map. The reason for this lies in the fact that our renderer has a small delay behind the leading tracker. If the rendering map is yet not covering enough pixels for some reason, we could potentially feedback a much sparser frame than we initially used during tracking. This could potentially disturb the update network \cite{droid}. We therefore check that the abs. rel. error between rendered disparity and tracking disparity is $\leq 0.2$. If at least 50\% of pixels satisfy this condition, then the frame is considered good.

\section{Extended Evaluation}
\label{sup:eval}
\begin{table*}[htb]
	\centering
	\begin{tabular}{lrcccccc}
		\toprule
		Technique & \#Gaussians  & PSNR$\uparrow$ & LPIPS$\downarrow$ & L1$\downarrow$ & PSNR$\uparrow$ & LPIPS$\downarrow$ & L1$\downarrow$ \\
		& & \multicolumn{3}{c}{\cellcolor[HTML]{EEEEEE}{\textit{KF}}} & \multicolumn{3}{c}{\cellcolor[HTML]{EEEEEE}{\textit{Non-KF}}} \\ 
		\multicolumn{8}{l}{\textit{\textbf{TUM-RGBD}}} \\[2pt] 
		Monocular & 118 889 & \textbf{26.84} & 0.129 & 16.67 & 24.62 & 0.156 & 17.37 \\
		P-RGBD & 119 100 & 26.53 & 0.131 & 8.50 & 24.81 & 0.155 & 8.38 \\
		RGBD & 123 232 & 26.81 & \textbf{0.110} & \textbf{4.26} & \textbf{24.89} & \textbf{0.144} & \textbf{4.63} \\
		& & & & & & & \\[-5pt]
		
		\multicolumn{8}{l}{\textit{\textbf{Replica}}} \\[2pt]
		Monocular & 246 637 & 39.47 & 0.031 & 3.33 & 38.42 & 0.032 & 3.47 \\
		P-RGBD & 248 175 & \textbf{39.66} & 0.029 & 3.34 & 38.27 & 0.031 & 3.53 \\
		RGBD & 235 825 & \textbf{39.66} & \textbf{0.028} & \textbf{0.55} & \textbf{38.87} & \textbf{0.029} & \textbf{0.61} \\
		& & & & & & & \\[-5pt]
		\bottomrule
	\end{tabular}	
	\caption{\textbf{Full Rendering results}. We report our overall best results with MCMC densification \cite{mcmc} averaged over 5 runs on TUM-RGBD~\cite{tum-rgbd} and Replica~\cite{replica} with refinement.
	}
	\label{tab:supp_full_render}
\end{table*}
\begin{table*}[htb]
	\centering
	\begin{tabular}{lrcccccc}
		\toprule
		Supervision & \#Gaussians & PSNR$\uparrow$ & LPIPS$\downarrow$ & L1$\downarrow$ & PSNR$\uparrow$ & LPIPS$\downarrow$ & L1$\downarrow$ \\
		& & \multicolumn{3}{c}{\cellcolor[HTML]{EEEEEE}{\textit{KF}}} & \multicolumn{3}{c}{\cellcolor[HTML]{EEEEEE}{\textit{Non-KF}}} \\ 
		\multicolumn{8}{l}{\textit{\textbf{TUM-RGBD}}} \\[2pt] 
		dense & 88 280 & 25.98 & 0.140 & \textbf{8.2} & \textbf{24.48} & 0.161 & \textbf{8.2} \\
		sparse & 100 156 & \textbf{26.67} & \textbf{0.129} & 15.6 & 24.37 & \textbf{0.155} & 16.7 \\
		& & & & & & & \\[-5pt]
		
		\multicolumn{7}{l}{\textit{\textbf{Replica}}} \\[2pt]
		dense & 264 343 & 38.79 & 0.0361 & 3.55 & \textbf{37.84} & 0.0371  & 3.64 \\
		sparse & 275 997 & \textbf{38.95} & \textbf{0.0347} & \textbf{2.96} & 37.82 & \textbf{0.0361} & \textbf{3.11} \\
		\bottomrule
	\end{tabular}	
	\caption{\textbf{Sparse vs. dense supervision of vanilla 3D Gaussian Splatting~\cite{3dgs} with monocular prior}. Geometry reconstruction depends heavily on the degree and quality of supervision. TUM-RGBD~\cite{tum-rgbd} does not have enough redundancy in frames for the filtered map to cover the scene. Replica~\cite{replica} on the other hand will produce enough reliable covisible 3D points, such that the filtered tracking map provides strong supervision for each Gaussian. For this reason, we can achieve better results when using the sparser, filtered map on Replica. This closes the gap to related work \cite{splat-slam}. We believe that with different priors and hyperparameters, we would achieve the same $L1$ error.
	}
	\label{tab:supp_dense_v_sparse}
\end{table*}

In this section, we want to provide more insights into how our system performs quantitatively and show more qualitative results. The reported rendering metrics for our comparison with related work 
are computed on the \textit{keyframe} images based on the estimated poses, as is standard. However, this can give a warped view on the quality of a method. We want to highlight several key points:
\begin{itemize}
	\item Every method has a \textit{different keyframe management} or builds their graph based on different thresholds.
	\item \textit{Not all metrics are reportedly available on all datasets.} We omitted an extensive evaluation of related work due to time constraints. Example: $L1$ metric is only readily available for Replica~\cite{replica}, however due to being a virtual dataset this metric is already quite saturated. The TUM-RGBD~\cite{tum-rgbd} benchmark is much more interesting. 
	\item \textit{Performance should be measured both on training and other frames!} Generalization of our test-time optimization is what normally counts, which is why we report results on non-training frames. 
	\item The \textit{difference between  modes} only becomes apparent when considering both geometry and predicted images for both training and other frames.
\end{itemize}

\begin{figure*}[b]
	\vspace{0em}
	\centering
	{
		\setlength{\tabcolsep}{1pt}
		\renewcommand{\arraystretch}{1}
		\newcommand{\sz}{0.2}
		\newcommand{\subsz}{0.2}
		\begin{tabular}{ccccc}
			\Large
			& GlORIE-SLAM~\cite{glorieslam}  & MonoGS~\cite{monogs} & \ours (Ours)  & Ground Truth \\[-1pt]
			\raisebox{0.8cm}{\rotatebox{90}{\makecell{\texttt{fr1}\\\texttt{room}}}} & 
			\includegraphics[width=\sz\linewidth, trim=10 10 10 10, clip, height=2.4cm]{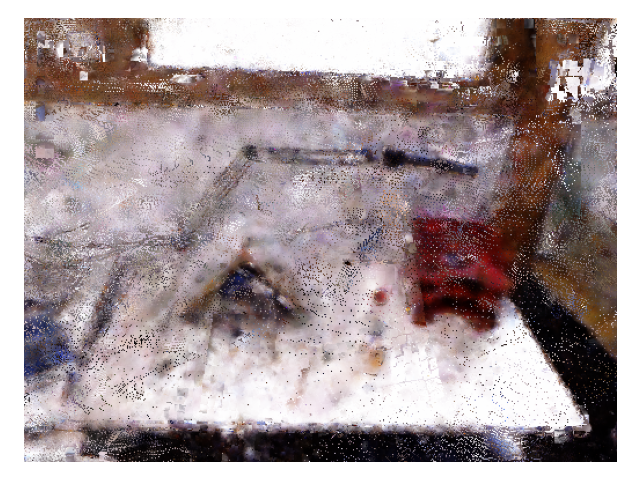} &
			\includegraphics[width=\sz\linewidth, trim=15 10 15 10, height=2.4cm]{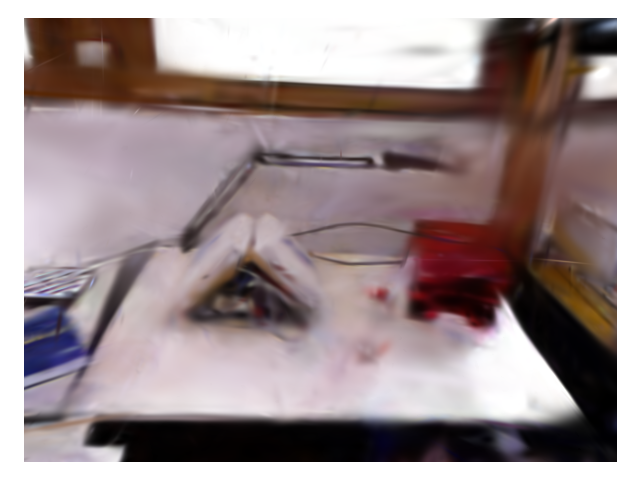} &
			\includegraphics[width=\sz\linewidth, trim=10 10 10 10, clip, height=2.4cm]{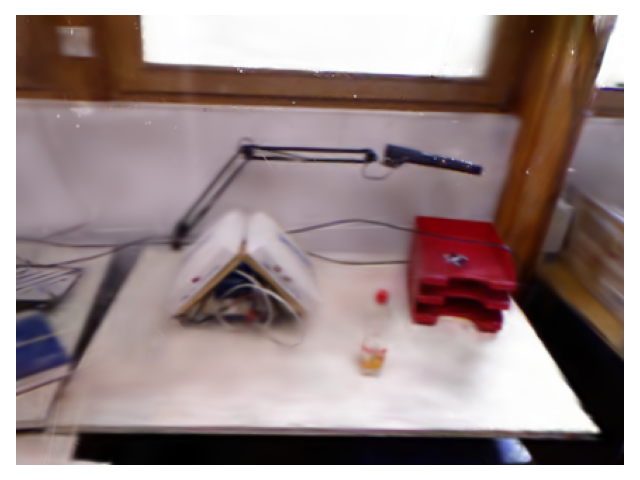} &
			\includegraphics[width=\sz\linewidth, clip, height=2.4cm]{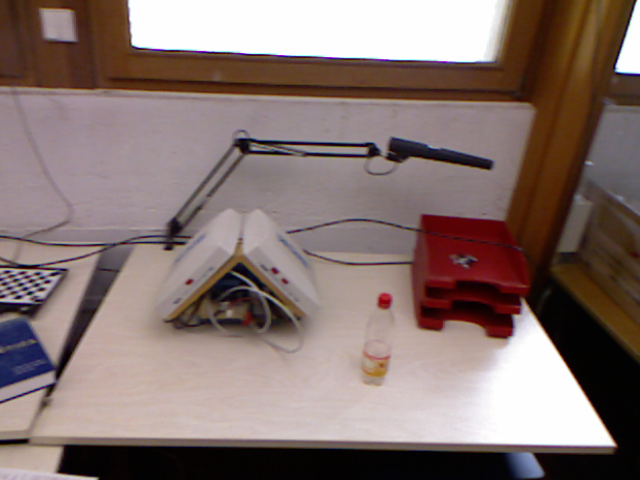} \\ 
			
			& \includegraphics[width=\sz\linewidth, trim=10 12 10 12, clip, height=2.4cm]{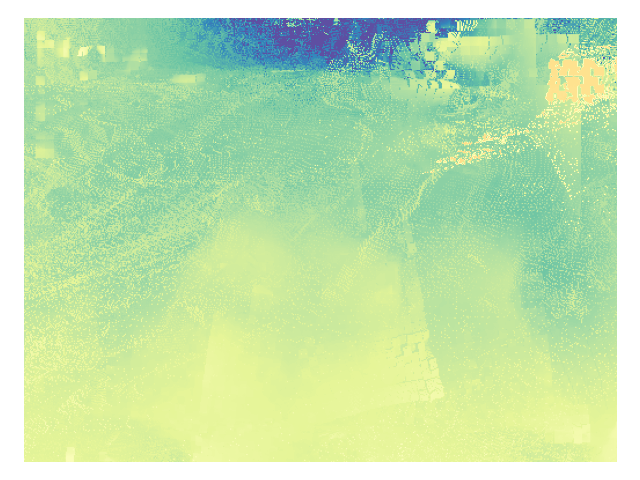} &
			\includegraphics[width=\sz\linewidth, trim=15 12 15 12, height=2.4cm]{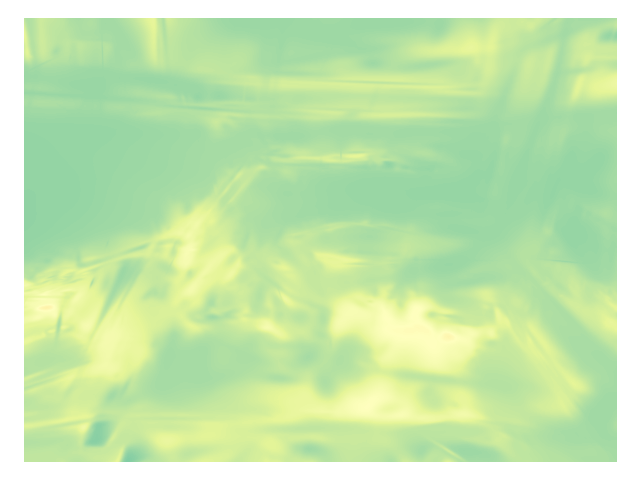} &
			\includegraphics[width=\sz\linewidth, trim=10 10 10 10, clip, height=2.4cm]{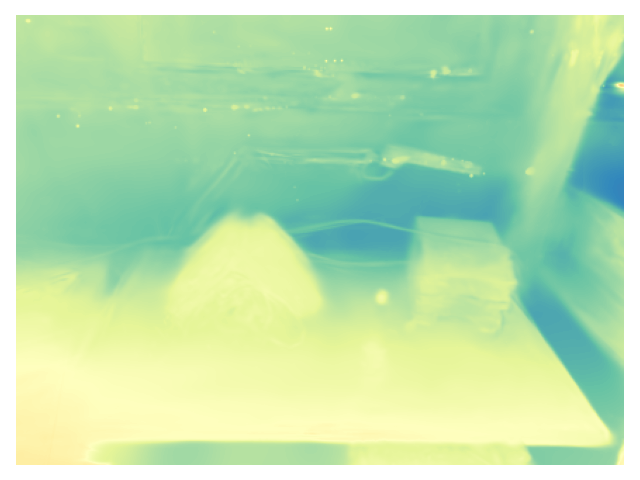} &
			\includegraphics[width=\sz\linewidth, clip, height=2.4cm]{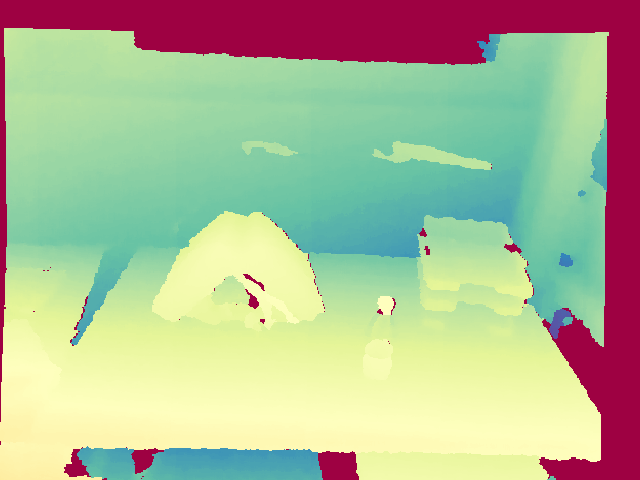} \\
			
			\raisebox{0.8cm}{\rotatebox{90}{\makecell{\texttt{fr2}\\\texttt{xyz}}}} & 
			\includegraphics[width=\sz\linewidth, trim=10 10 10 12, clip, height=2.4cm]{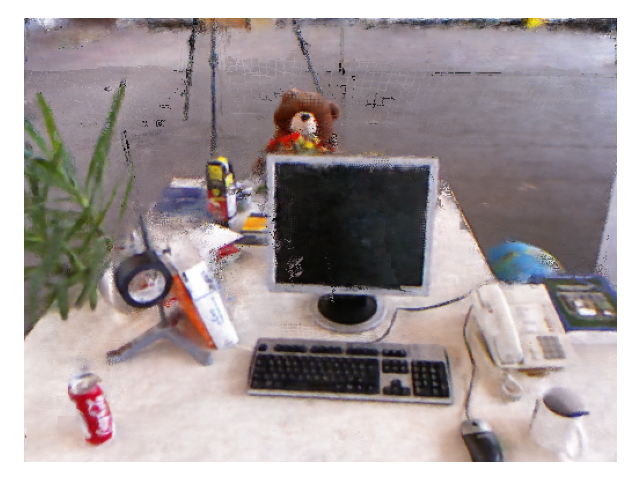} &
			\includegraphics[width=\sz\linewidth, trim=15 10 15 10, height=2.4cm]{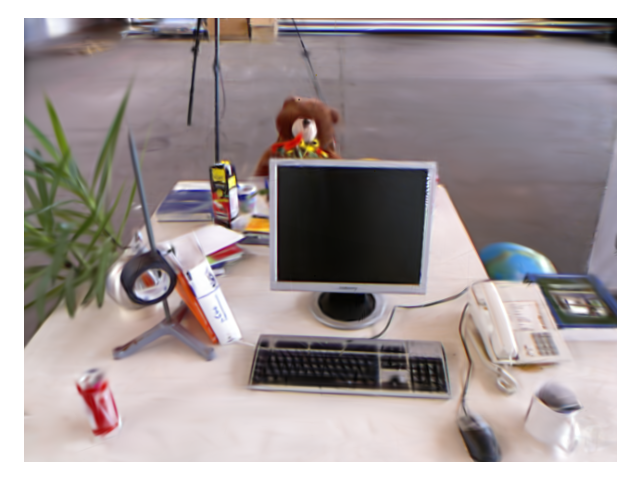} &
			\includegraphics[width=\sz\linewidth, trim=10 10 10 10, clip, height=2.4cm]{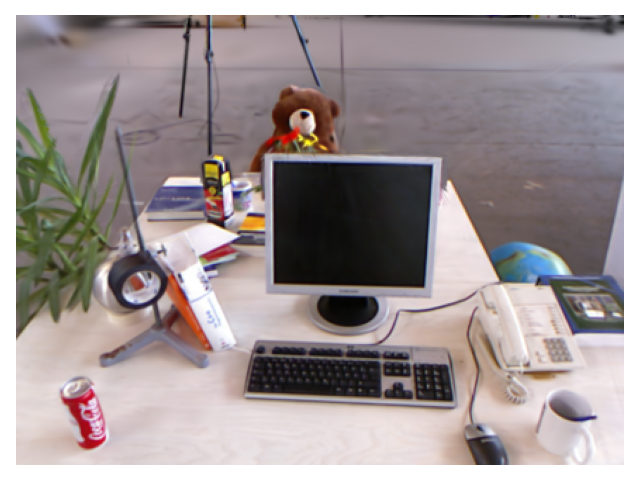} &
			\includegraphics[width=\sz\linewidth, clip, height=2.4cm]{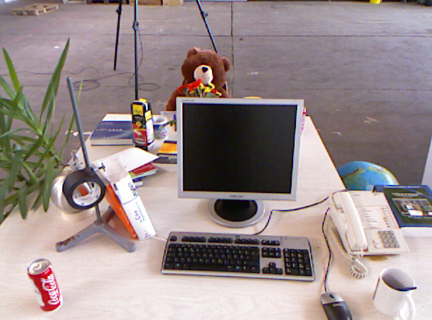} \\ 
			
			& \includegraphics[width=\sz\linewidth, trim=10 12 10 12, clip, height=2.4cm]{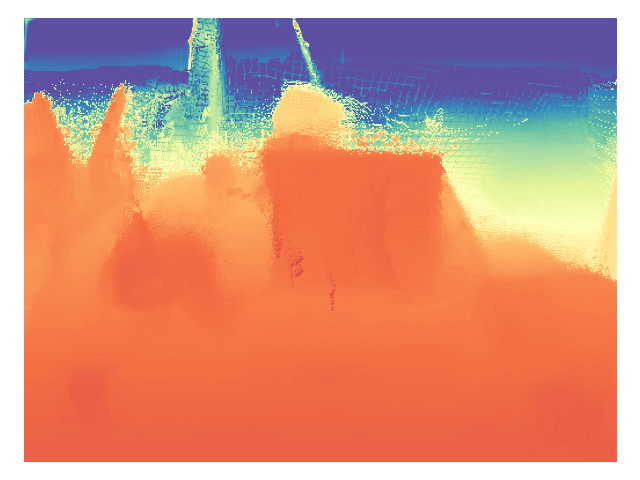} &
			\includegraphics[width=\sz\linewidth, trim=15 12 15 12, height=2.4cm]{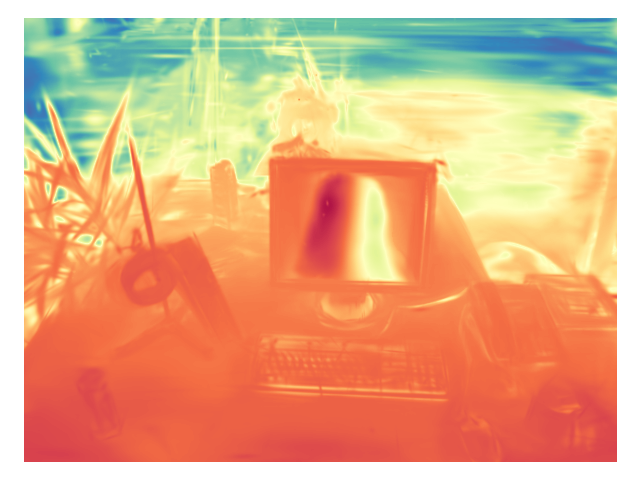} &
			\includegraphics[width=\sz\linewidth, trim=10 10 10 10, clip, height=2.4cm]{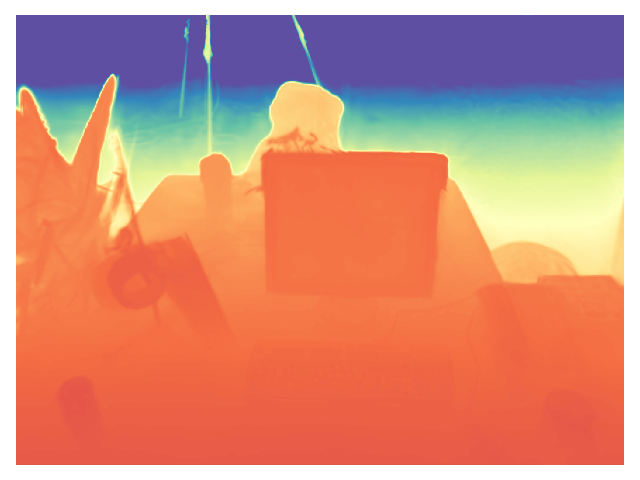} &
			\includegraphics[width=\sz\linewidth, clip, height=2.4cm]{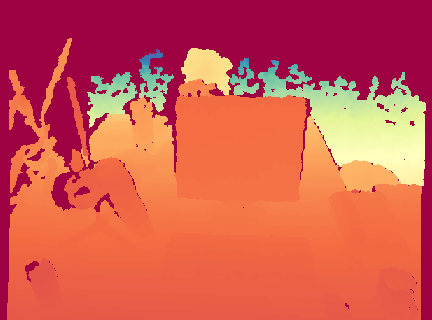} \\
			
			\\[0.01cm]
	
			\raisebox{0.8cm}{\rotatebox{90}{\makecell{\texttt{fr1}\\\texttt{room}}}} &
			\includegraphics[width=\sz\linewidth, height=2.4cm]{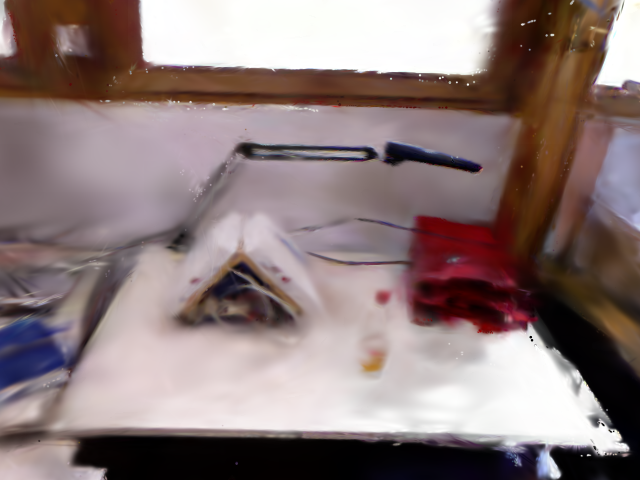} &
			\includegraphics[width=\sz\linewidth, trim=15 12 15 12, height=2.4cm]{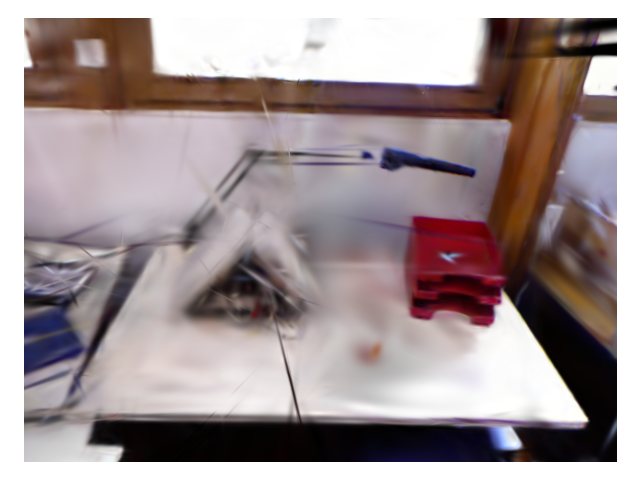} &
			\includegraphics[width=\sz\linewidth, trim=5 10 5 10, clip, height=2.4cm]{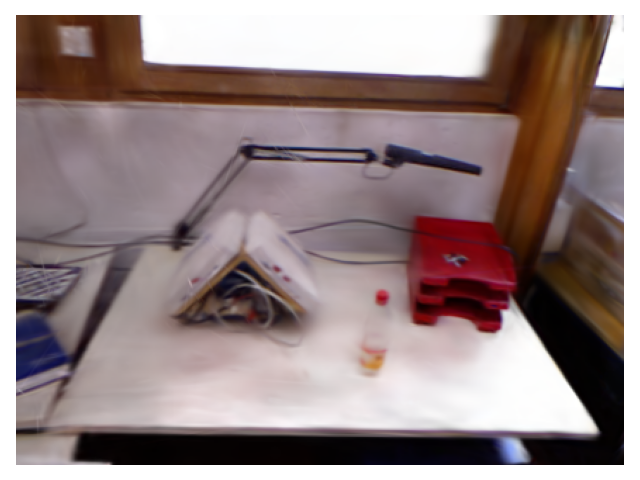} &
			\includegraphics[width=\sz\linewidth, clip, height=2.4cm]{figures/comparison/groundtruth/fr1_room/0338.png} \\
			&
			\includegraphics[width=\sz\linewidth, height=2.4cm]{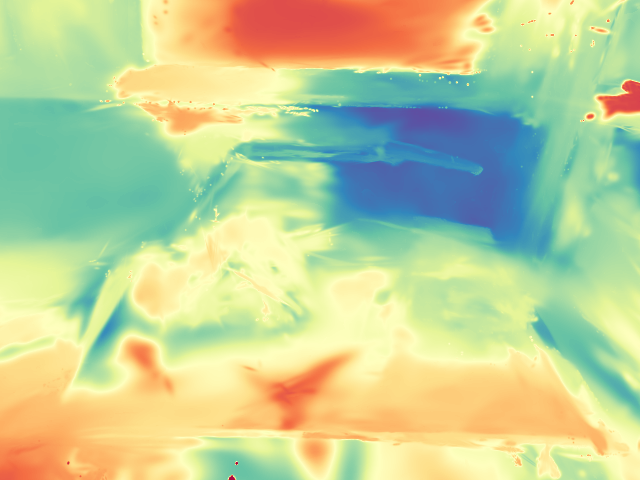} &
			\includegraphics[width=\sz\linewidth, trim=15 12 15 12, height=2.4cm]{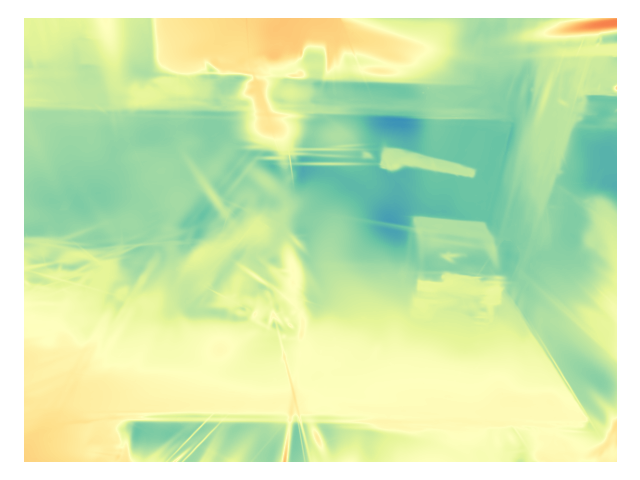} &
			\includegraphics[width=\sz\linewidth, trim=5 10 5 10, clip, height=2.4cm]{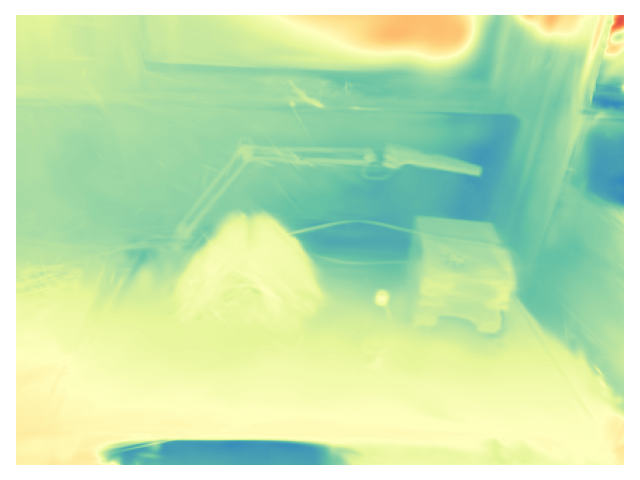} &
			\includegraphics[width=\sz\linewidth, clip, height=2.4cm]{figures/comparison/groundtruth/fr1_room/depth_0338.png} \\	
			
			\raisebox{0.8cm}{\rotatebox{90}{\makecell{\texttt{fr2}\\\texttt{xyz}}}} & 
			\includegraphics[width=\sz\linewidth, height=2.4cm]{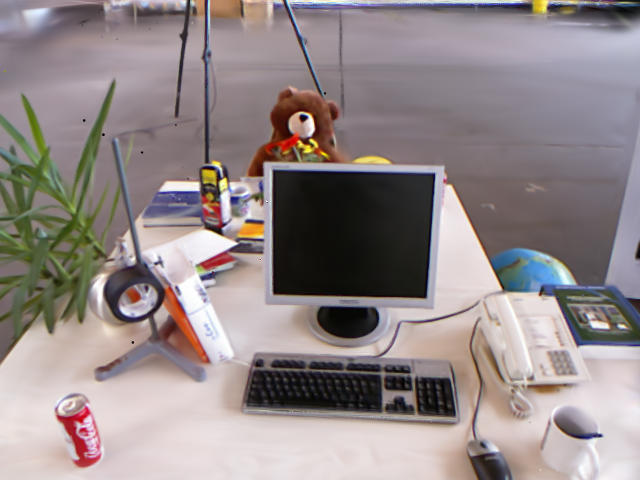} &
			\includegraphics[width=\sz\linewidth, trim=15 12 15 12, height=2.4cm]{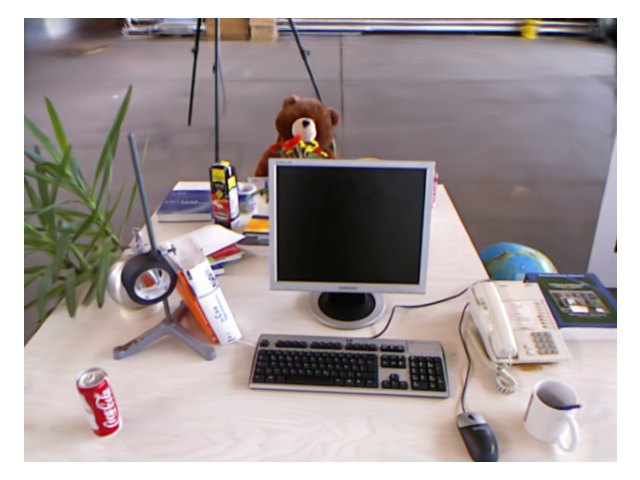} &
			\includegraphics[width=\sz\linewidth, trim=5 10 5 10, clip, height=2.4cm]{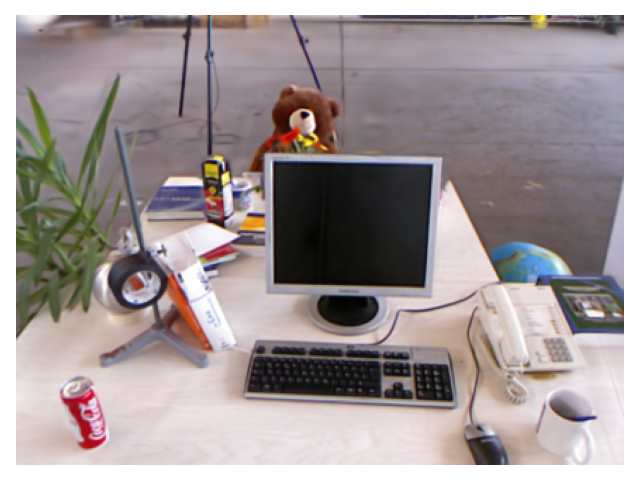} &
			\includegraphics[width=\sz\linewidth, clip, height=2.4cm]{figures/comparison/groundtruth/fr2_xyz/0976.png} \\
			&
			\includegraphics[width=\sz\linewidth, height=2.4cm]{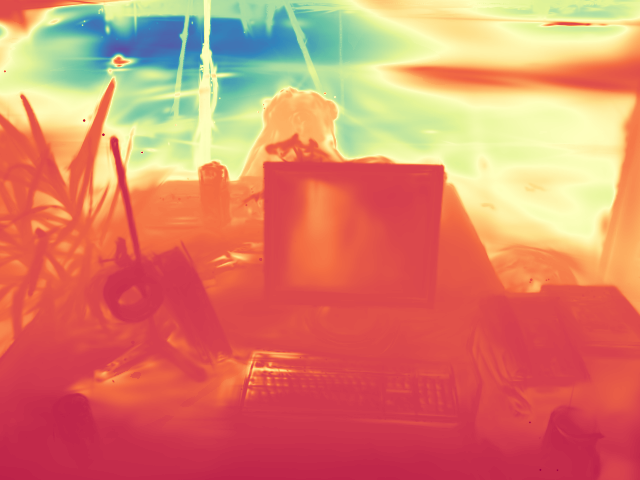} &
			\includegraphics[width=\sz\linewidth, trim=15 12 15 12, height=2.4cm]{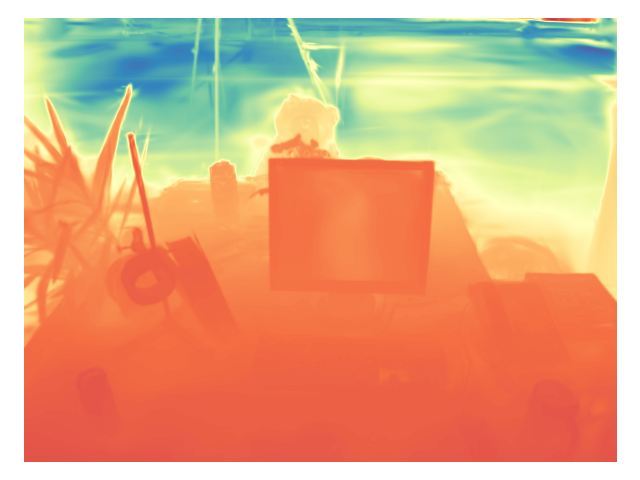} &
			\includegraphics[width=\sz\linewidth, trim=5 10 5 10, clip, height=2.4cm]{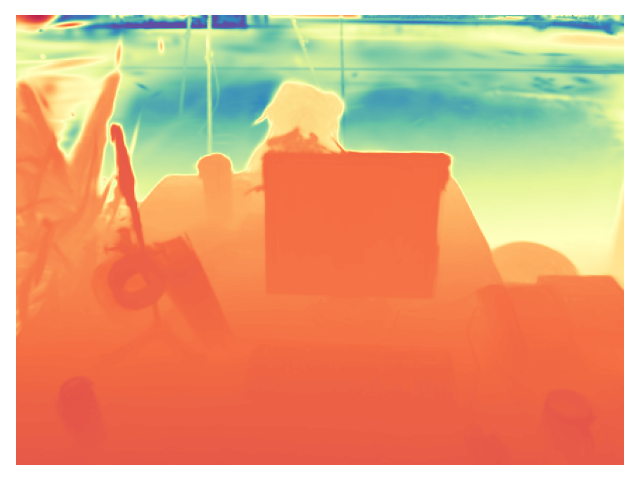} &
			\includegraphics[width=\sz\linewidth, clip, height=2.4cm]{figures/comparison/groundtruth/fr2_xyz/depth_0976.png} \\	
			
			\Large
			& Photo-SLAM~\cite{photoslam} & MonoGS~\cite{monogs} & \ours (Ours)  & Ground Truth \\[-4pt]
			& & & &
		\end{tabular}
	}
	\caption{\textbf{More Rendering Results on TUM-RGBD~\cite{tum-rgbd}}. Top four rows are from \textit{monocular} input, bottom from \textit{RGBD}.}
	\label{fig:supp_render_tum}
	\vspace{0em}
\end{figure*}

\begin{figure*}[b]
	\vspace{0em}
	\centering
	{
		\setlength{\tabcolsep}{1pt}
		\renewcommand{\arraystretch}{1}
		\newcommand{\sz}{0.2}
		\newcommand{\subsz}{0.2}
		\begin{tabular}{ccccc}
			\Large
			& Mono & P-RGBD & RGBD  & Ground Truth \\[-1pt]
			\raisebox{0.4cm}{\rotatebox{90}{\makecell{\texttt{office 2}}}} & 
			\includegraphics[width=\sz\linewidth, trim=10 50 10 50, clip, height=2.4cm]{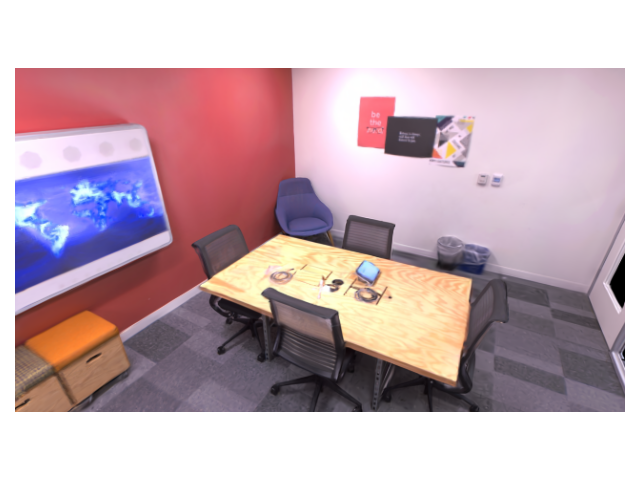} &
			\includegraphics[width=\sz\linewidth, trim=10 50 10 50, clip, height=2.4cm]{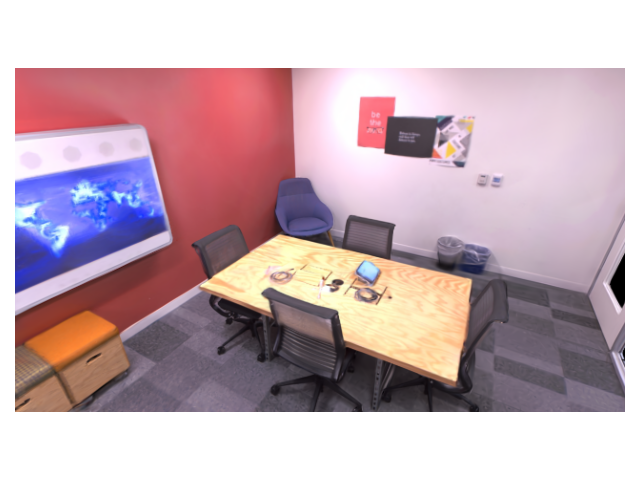} &
			\includegraphics[width=\sz\linewidth, trim=10 50 10 50, clip, height=2.4cm]{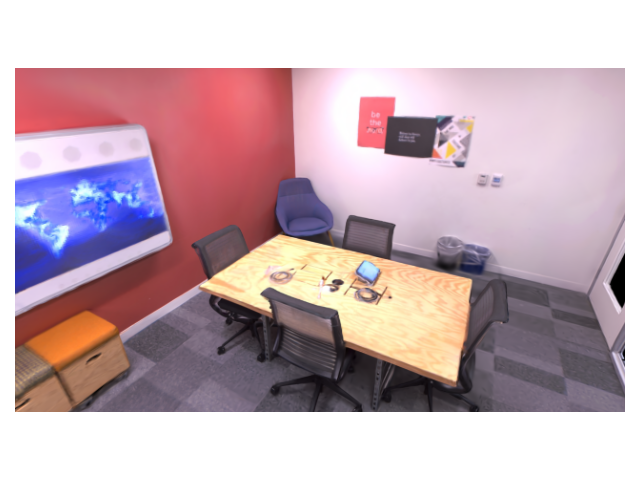} &
			\includegraphics[width=\sz\linewidth, height=2.4cm]{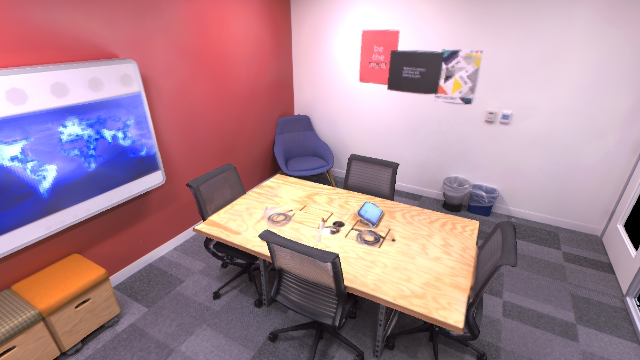} \\ 
			
			& \includegraphics[width=\sz\linewidth, trim=10 50 10 50, clip, height=2.4cm]{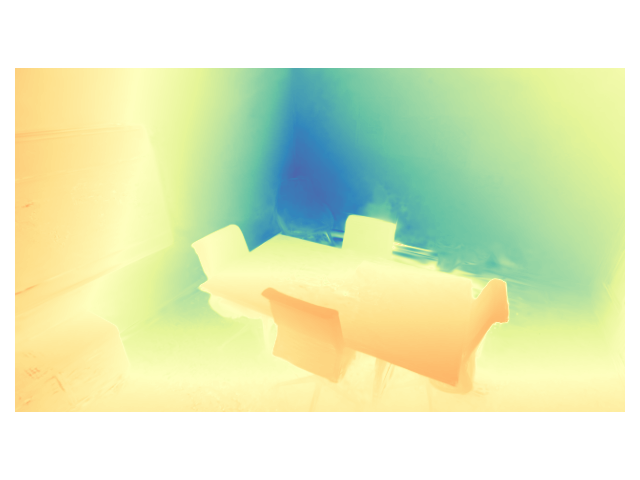} &
			\includegraphics[width=\sz\linewidth, trim=10 50 10 50, clip, height=2.4cm]{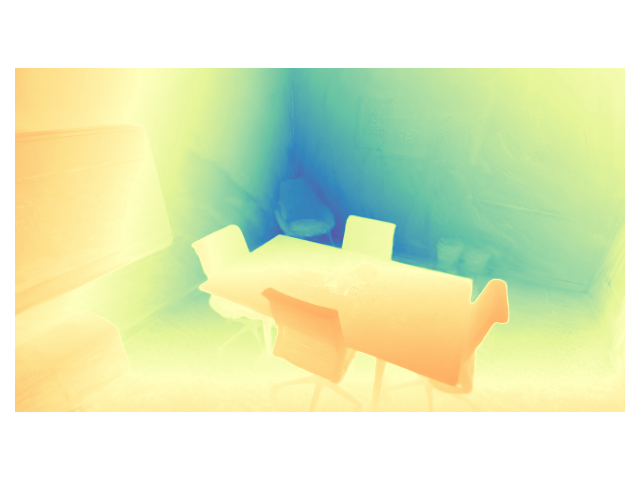} &
			\includegraphics[width=\sz\linewidth, trim=10 50 10 50, clip, height=2.4cm]{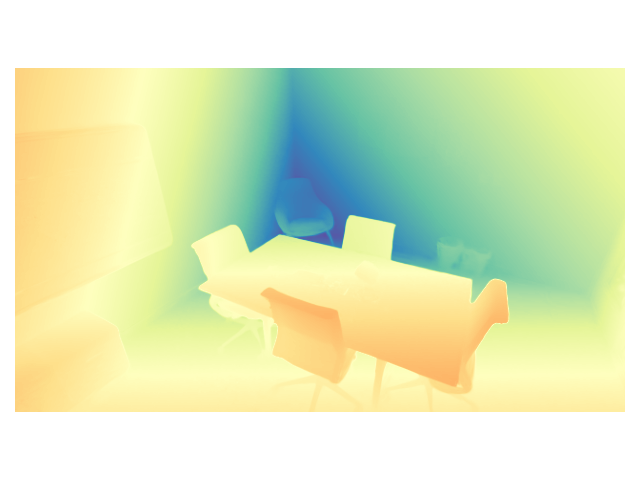} &
			\includegraphics[width=\sz\linewidth, height=2.4cm]{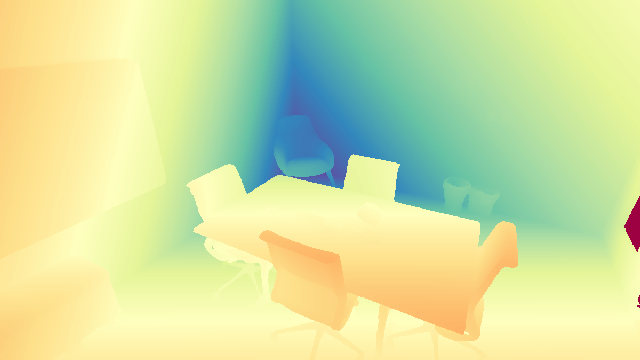} \\
			
			\raisebox{0.5cm}{\rotatebox{90}{\makecell{\texttt{room 0}}}} & 
			\includegraphics[width=\sz\linewidth, trim=10 50 10 50, clip, height=2.4cm]{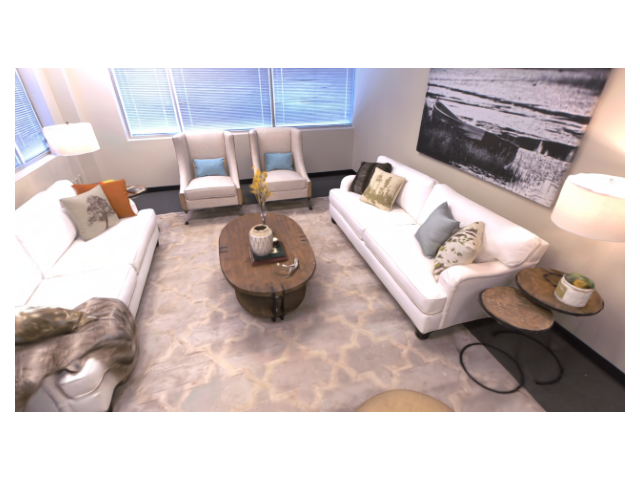} &
			\includegraphics[width=\sz\linewidth, trim=10 50 10 50, clip, height=2.4cm]{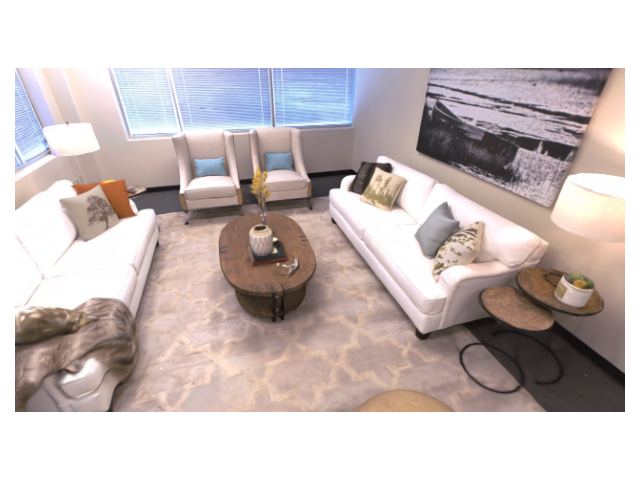} &
			\includegraphics[width=\sz\linewidth, trim=10 50 10 50, clip, height=2.4cm]{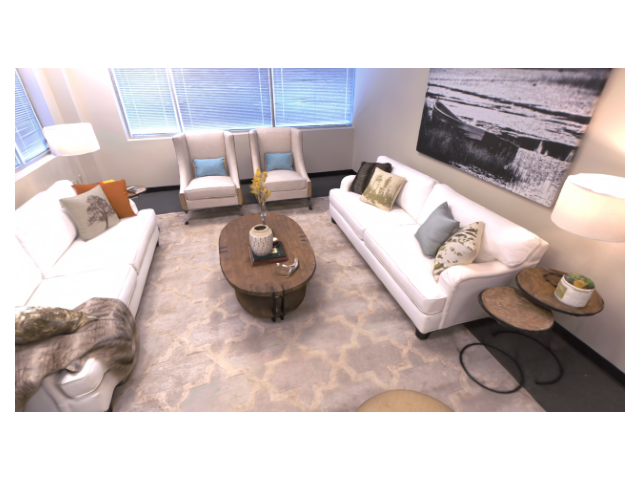} &
			\includegraphics[width=\sz\linewidth, height=2.4cm]{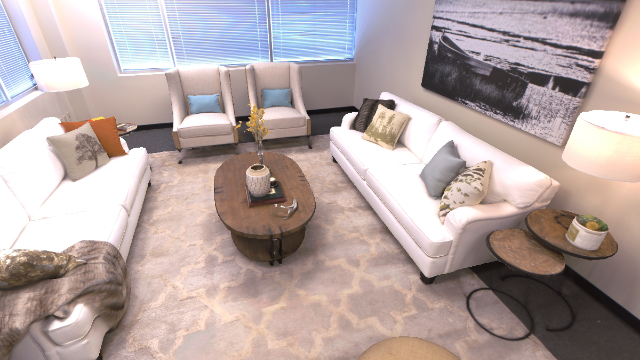} \\ 
			
			& \includegraphics[width=\sz\linewidth, trim=10 50 10 50, clip, height=2.4cm]{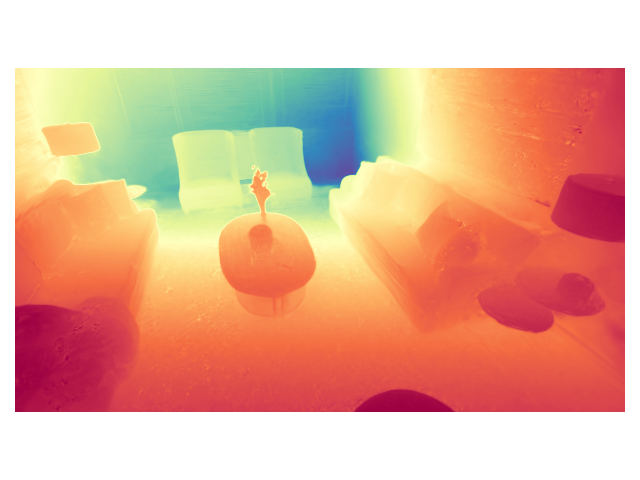} &
			\includegraphics[width=\sz\linewidth, trim=10 50 10 50, clip, height=2.4cm]{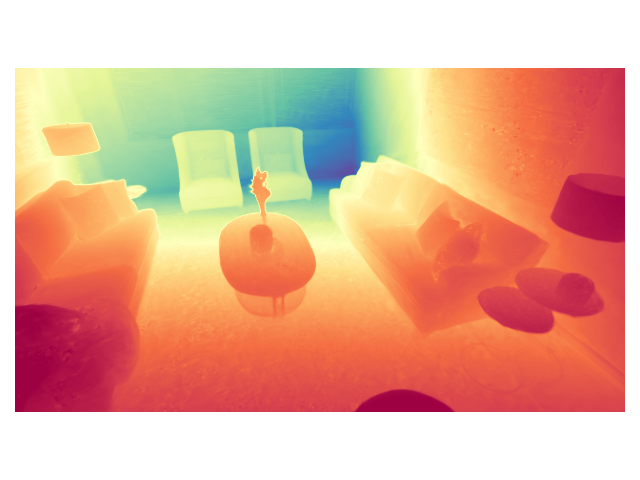} &
			\includegraphics[width=\sz\linewidth, trim=10 50 10 50, clip, height=2.4cm]{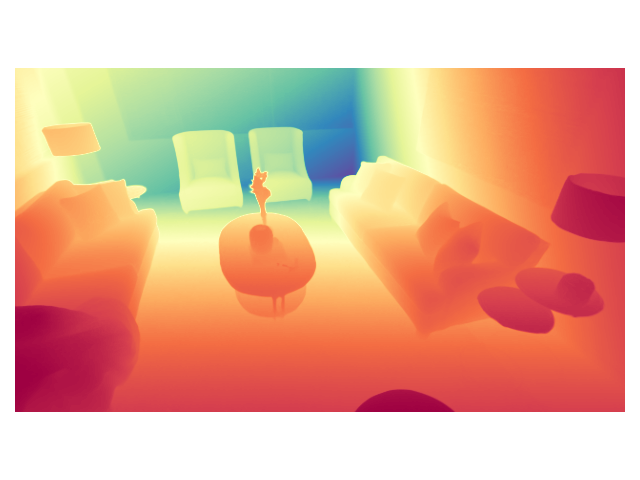} &
			\includegraphics[width=\sz\linewidth, height=2.4cm]{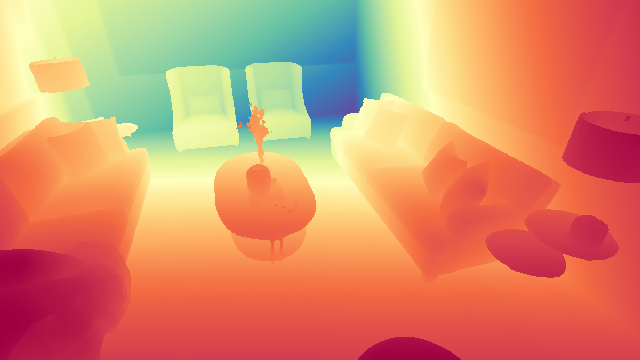} \\
			
			\\[0.01cm]
			
			\raisebox{0.5cm}{\rotatebox{90}{\makecell{\texttt{room 1}}}} & 
			\includegraphics[width=\sz\linewidth, trim=10 50 10 50, clip, height=2.4cm]{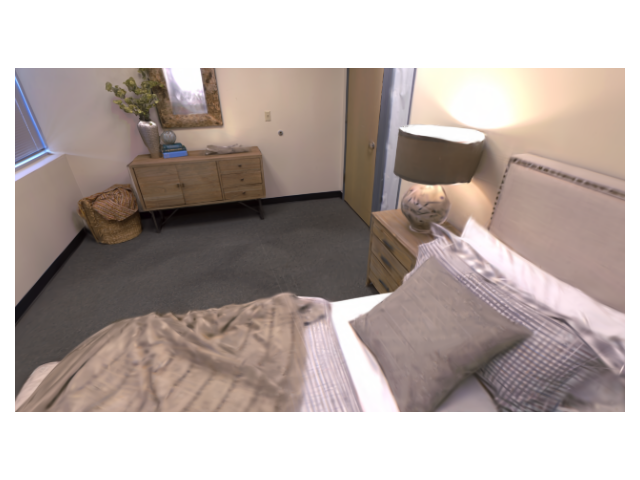} &
			\includegraphics[width=\sz\linewidth, trim=10 50 10 50, clip, height=2.4cm]{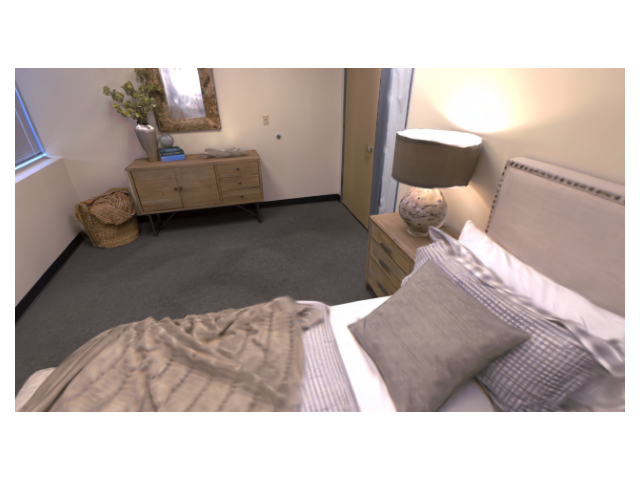} &
			\includegraphics[width=\sz\linewidth, trim=10 50 10 50, clip, height=2.4cm]{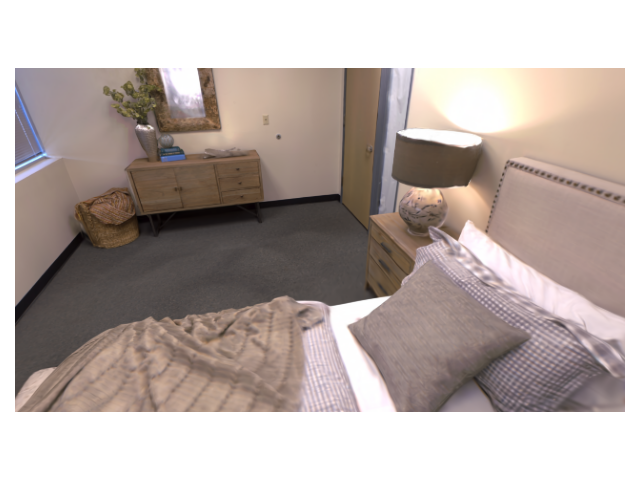} &
			\includegraphics[width=\sz\linewidth, height=2.4cm]{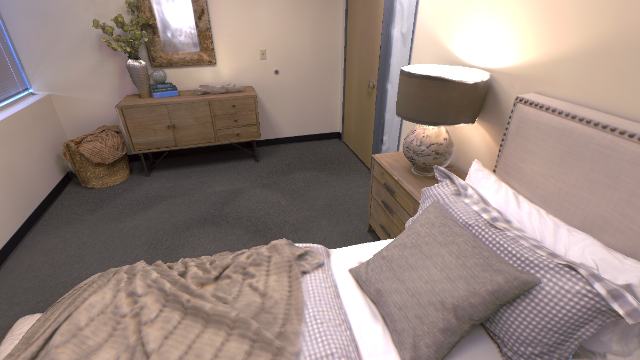} \\
			&
			\includegraphics[width=\sz\linewidth, trim=10 50 10 50, clip, height=2.4cm]{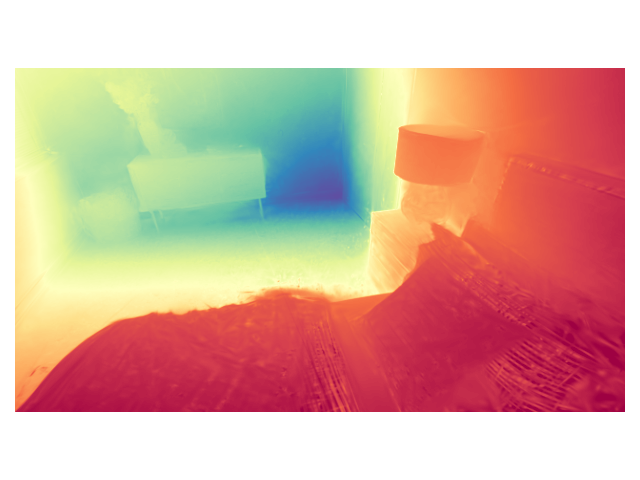} &
			\includegraphics[width=\sz\linewidth, trim=10 50 10 50, clip, height=2.4cm]{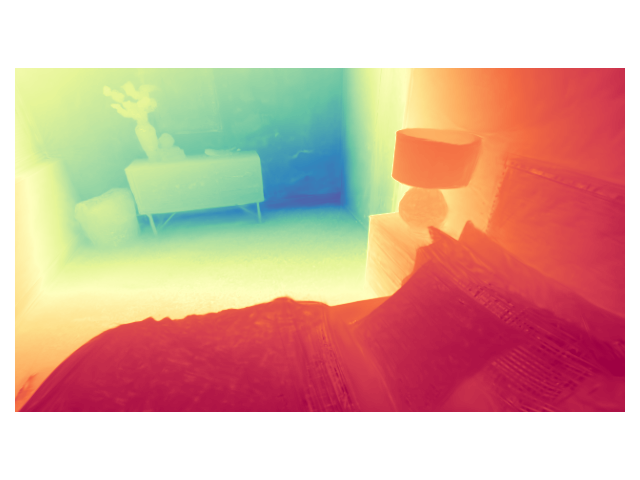} &
			\includegraphics[width=\sz\linewidth, trim=10 50 10 50, clip, height=2.4cm]{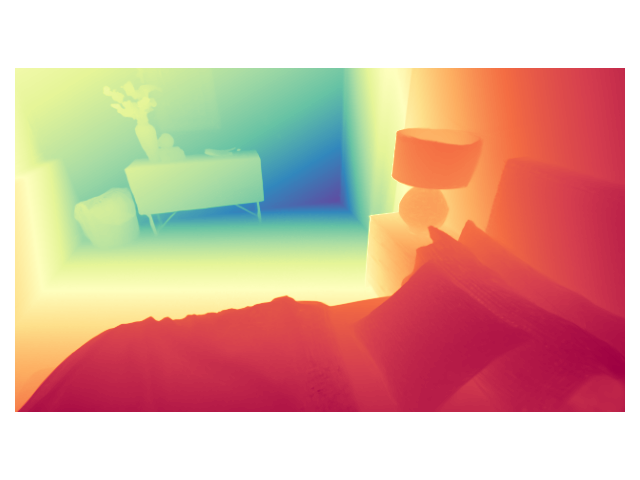} &
			\includegraphics[width=\sz\linewidth, height=2.4cm]{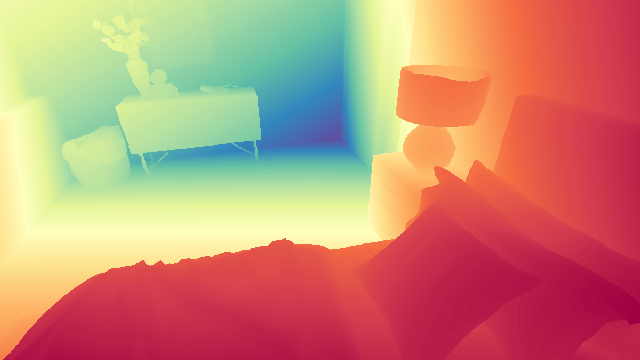} \\	
	
			\raisebox{0.5cm}{\rotatebox{90}{\makecell{\texttt{room 2}}}} & 	 
			\includegraphics[width=\sz\linewidth, trim=10 50 10 50, clip, height=2.4cm]{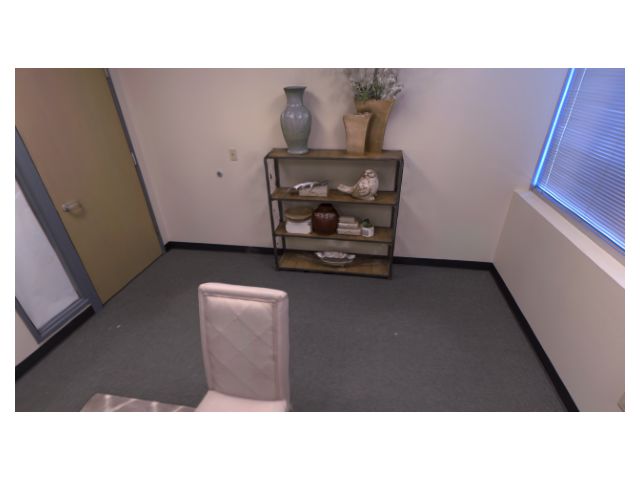} &
			\includegraphics[width=\sz\linewidth, trim=10 50 10 50, clip, height=2.4cm]{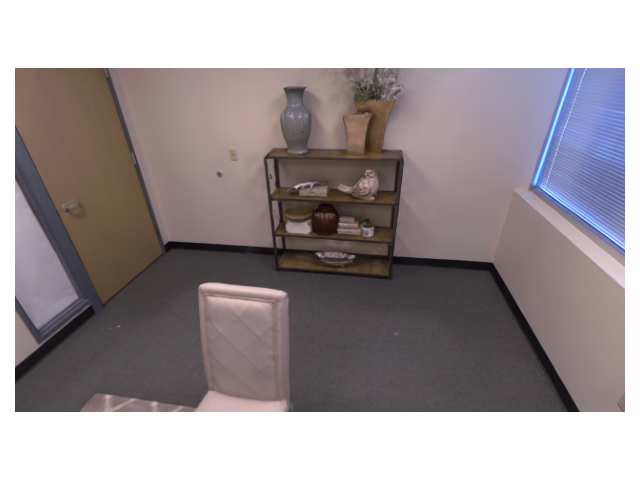} &
			\includegraphics[width=\sz\linewidth, trim=10 50 10 50, clip, height=2.4cm]{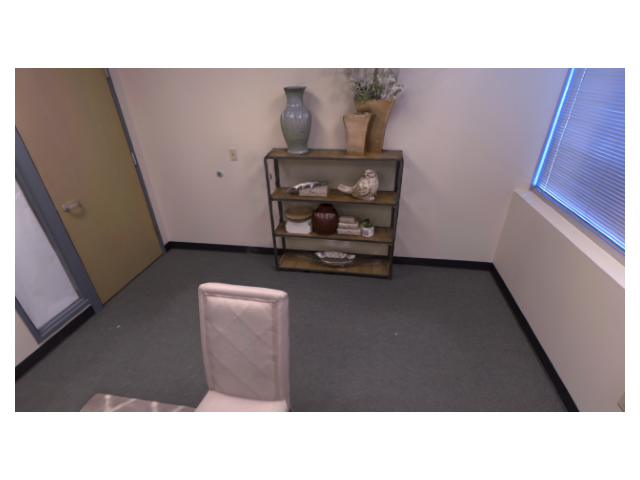} &
			\includegraphics[width=\sz\linewidth, height=2.4cm]{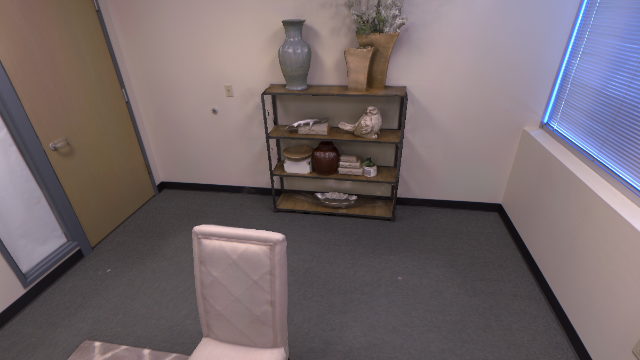} \\
			&
			\includegraphics[width=\sz\linewidth, trim=10 50 10 50, clip, height=2.4cm]{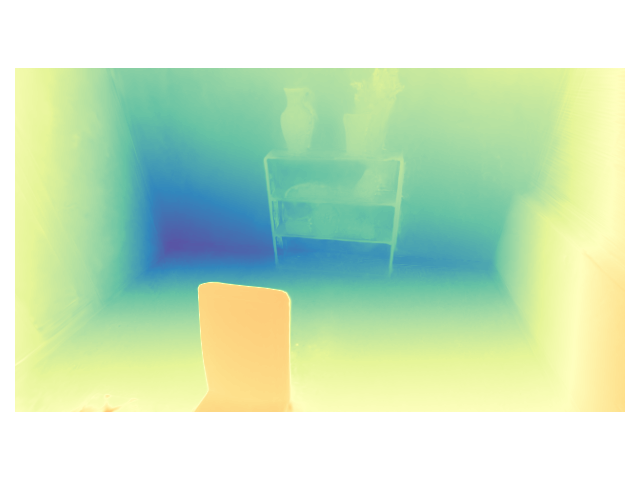} &
			\includegraphics[width=\sz\linewidth, trim=10 50 10 50, clip, height=2.4cm]{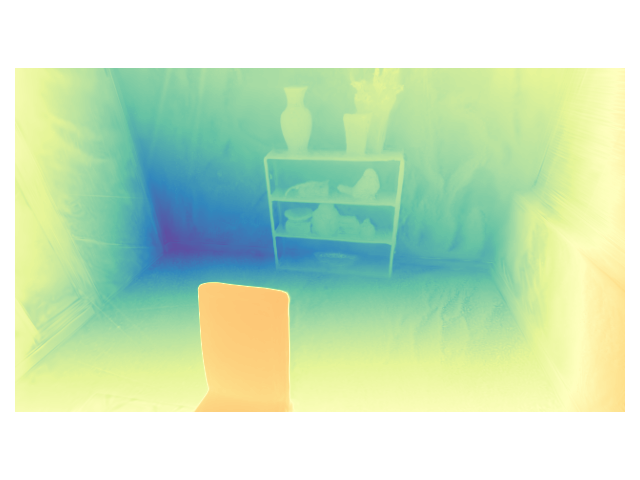} &
			\includegraphics[width=\sz\linewidth, trim=10 50 10 50, clip, height=2.4cm]{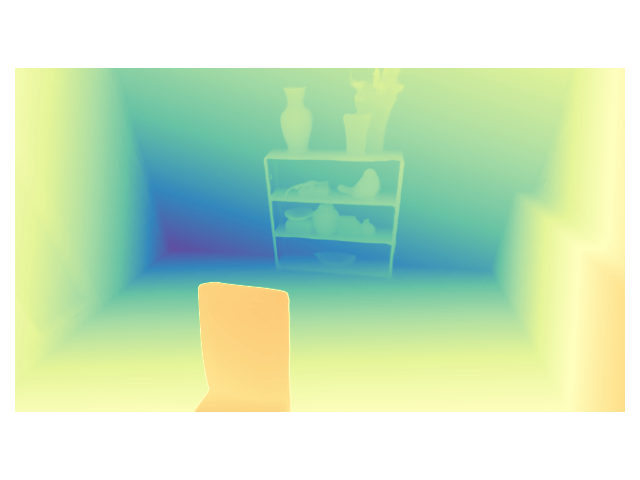} &
			\includegraphics[width=\sz\linewidth, height=2.4cm]{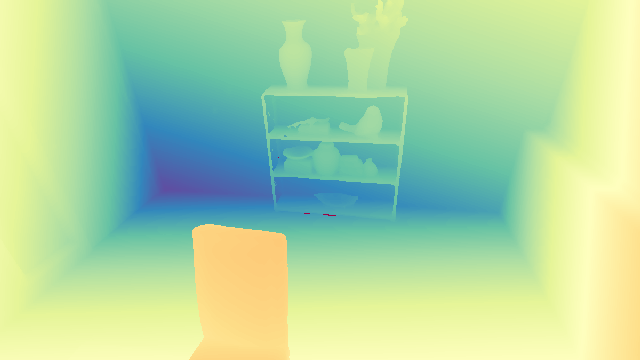} \\	
		\end{tabular}
	}
	\caption{\textbf{Rendering Results on Replica~\cite{replica}}. We show non-training frames in multiple input modalities. Note how visually close the predictions are to the groundtruth.}
	\label{fig:supp_render_replica}
	\vspace{0em}
\end{figure*}

We show our full evaluation metrics of the overall best results in Table \ref{tab:supp_full_render}. We can only see a clean progression from monocular to \textit{RGBD} inputs on the challenging TUM-RGBD~\cite{tum-rgbd} benchmark. We want to highlight, that strict monocular methods can overfit the appearance of training frames very well independent of tracking accuracy or geometric accuracy. However, we can generalize better and achieve much more accurate geometry when using additional depth priors. The benefit of a monocular prior \cite{metric3d} seems to be much smaller on Replica~\cite{replica}. We found out in Table \ref{tab:supp_dense_v_sparse}, that depending on the depth supervision signal this result changes. We also suspect concurrent work \cite{splat-slam} to supervise with a filtered depth map for this reason. Figure \ref{fig:supp_render_tum} and \ref{fig:supp_render_replica} show qualitative examples on top to get a feeling for how good methods work. We specifically chose non-training frames, which might put us at a disadvantage. We can observe clear improvements on fine-structured details, such as the lamp or background. 

Due to our dense map both in tracking and rendering we can achieve better reconstructions than related work. For monocular reconstructions, we specifically show our results with a depth prior \cite{metric3d}, which achieves much more accurate geometric reconstruction and better photo-realism on non-training frames than the monocular counter-part. This holds true even for slightly worse $L1$ metrics on Replica, as can be seen in the qualitative images. Results on Replica are already so accurate, that slight scale differences across time can create slightly non-flat walls. 
\begin{table*}[htb]
	\centering
	\begin{tabular}{lrcccccc}
		\toprule
		Technique & \# Gaussians  & PSNR$\uparrow$ & LPIPS$\downarrow$ & L1$\downarrow$ & PSNR$\uparrow$ & LPIPS$\downarrow$ & L1$\downarrow$ \\
		& & \multicolumn{3}{c}{\cellcolor[HTML]{EEEEEE}{\textit{KF}}} & \multicolumn{3}{c}{\cellcolor[HTML]{EEEEEE}{\textit{Non-KF}}} \\ 
		\multicolumn{8}{l}{\textit{\textbf{no refinement}}} \\[2pt] 
		2DGS~\cite{2dgs} & 173 309 & 20.71 & 0.31 & 10.2 & 19.84 & 0.33 & 10.3 \\
		3DGS~\cite{3dgs} & 111 878 & 23.26 & 0.23 & 9.1 & 22.46 & 0.25 & 9.2 \\
		+ MCMC~\cite{mcmc} & 113 060 & \textbf{23.78} & \textbf{0.21} & \textbf{8.2} & \textbf{22.81} & \textbf{0.23} & \textbf{8.4} \\
		& & & & & & & \\[-5pt]
		
		\multicolumn{8}{l}{\textit{\textbf{with refinement}}} \\[2pt]
		2DGS~\cite{2dgs} & 131 576 & 22.87 & 0.21 & 8.8 & 21.73 & 0.23 & 8.7 \\
		3DGS~\cite{3dgs} &  88 280 & 25.98 & 0.14 & \textbf{8.2} & 24.47 & 0.16 & \textbf{8.2} \\
		+ MCMC~\cite{mcmc} & 119 100 & \textbf{26.53} & \textbf{0.13} & 8.5 & \textbf{24.81} & \textbf{0.15} & 8.4 \\
		& & & & & & & \\[-5pt]
		\bottomrule
	\end{tabular}	
	\caption{\textbf{Ablation Rendering Techniques}. We report results averaged over 5 runs on TUM-RGBD~\cite{tum-rgbd} in \textit{P-RGBD} mode using \cite{metric3d} as a prior. We show a small progression with and without refinement. While 2D Gaussian Splatting \cite{2dgs} quickly produces smooth surfaces, this is not rewarded in the $L1$ error metric. 
	}
	\label{tab:supp_ablation_render}
\end{table*}
\begin{figure*}[b]
	\centering
	\includegraphics[width=0.8\linewidth]{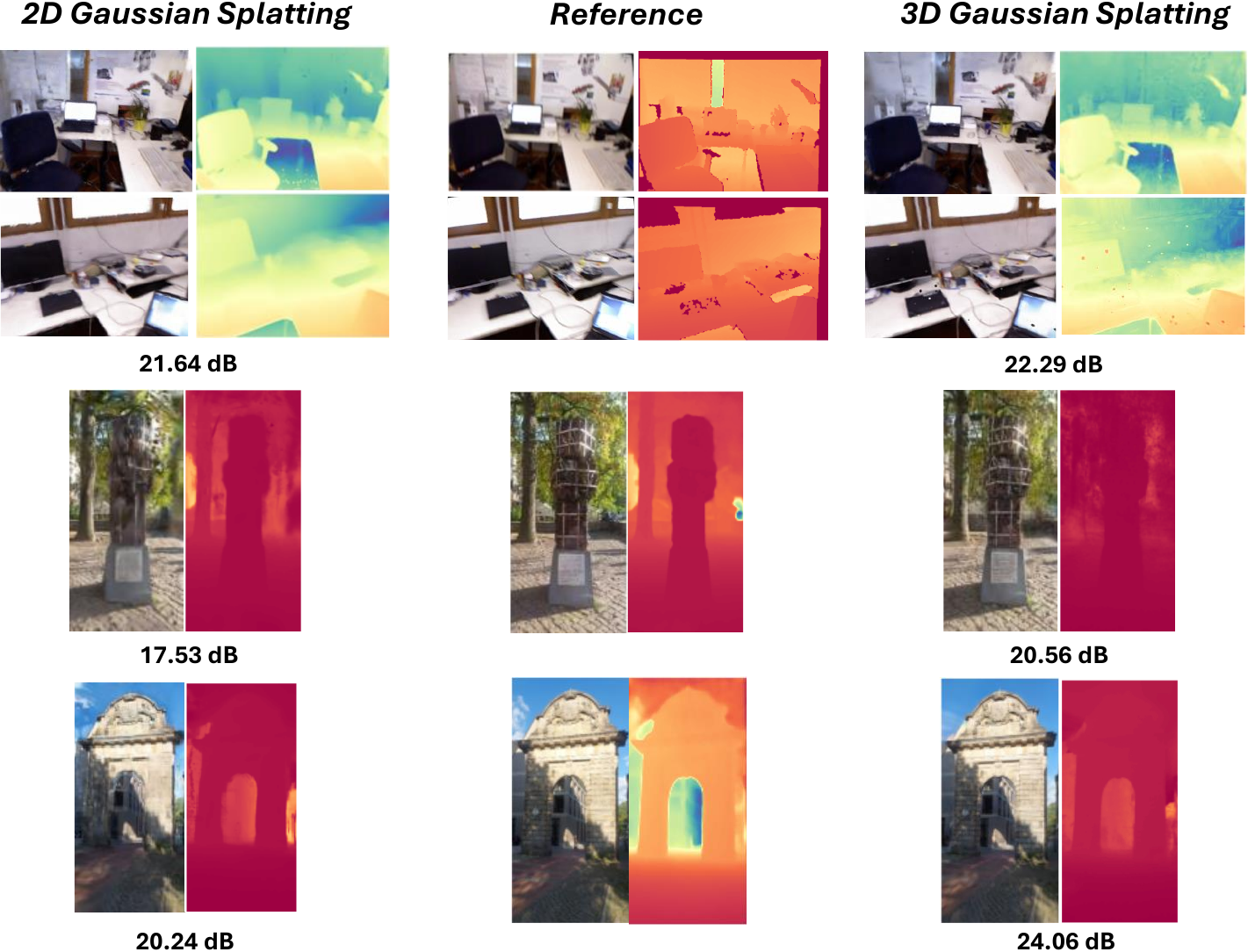}
	\caption{
		\textbf{Geometry vs. appearance}. We found, that 2D Gaussian Splatting \cite{2dgs} can quickly create smooth surfaces and does not accumulate many floaters in outdoor scenes. However, the rendering quality lacks behind 3D Gaussian Splatting \cite{3dgs} and as long as good supervision exists we can achieve better $L1$ metrics with 3D Gaussian Splatting.
	}
	\label{fig:2dgs_vs_3dgs}
\end{figure*}
Table \ref{tab:supp_ablation_render} shows a detailed ablation of Rendering techniques. We did not combine 2D Gaussian Splatting with the improved densification strategy \cite{mcmc}, however we expect this to gain a similar improvement. We did not succeed in achieving better reconstructions for 2D Gaussian Splatting on TUM-RGBD~\cite{tum-rgbd}. However, we observe a clear benefit of this representation similar to the results in the respective paper, see examples in Figure \ref{fig:2dgs_vs_3dgs}. We can quickly converge to flat surfaces, which helps to avoid many floaters in outdoor-scenarios. On the used indoor datasets, vanilla 3D Gaussians perform better. 
\begin{figure*}[b]
	\vspace{0em}
	\centering
	{
		\setlength{\tabcolsep}{1pt}
		\renewcommand{\arraystretch}{1}
		\newcommand{\sz}{0.19}
		\newcommand{\subsz}{0.2}
		\begin{tabular}{cccc}
			Lotus \cite{lotus-d} & DepthAnything \cite{depthanything} & Metric3D \cite{metric3d}  & Reference \\[2pt]
			
			\includegraphics[width=\sz\linewidth, height=3.0cm]{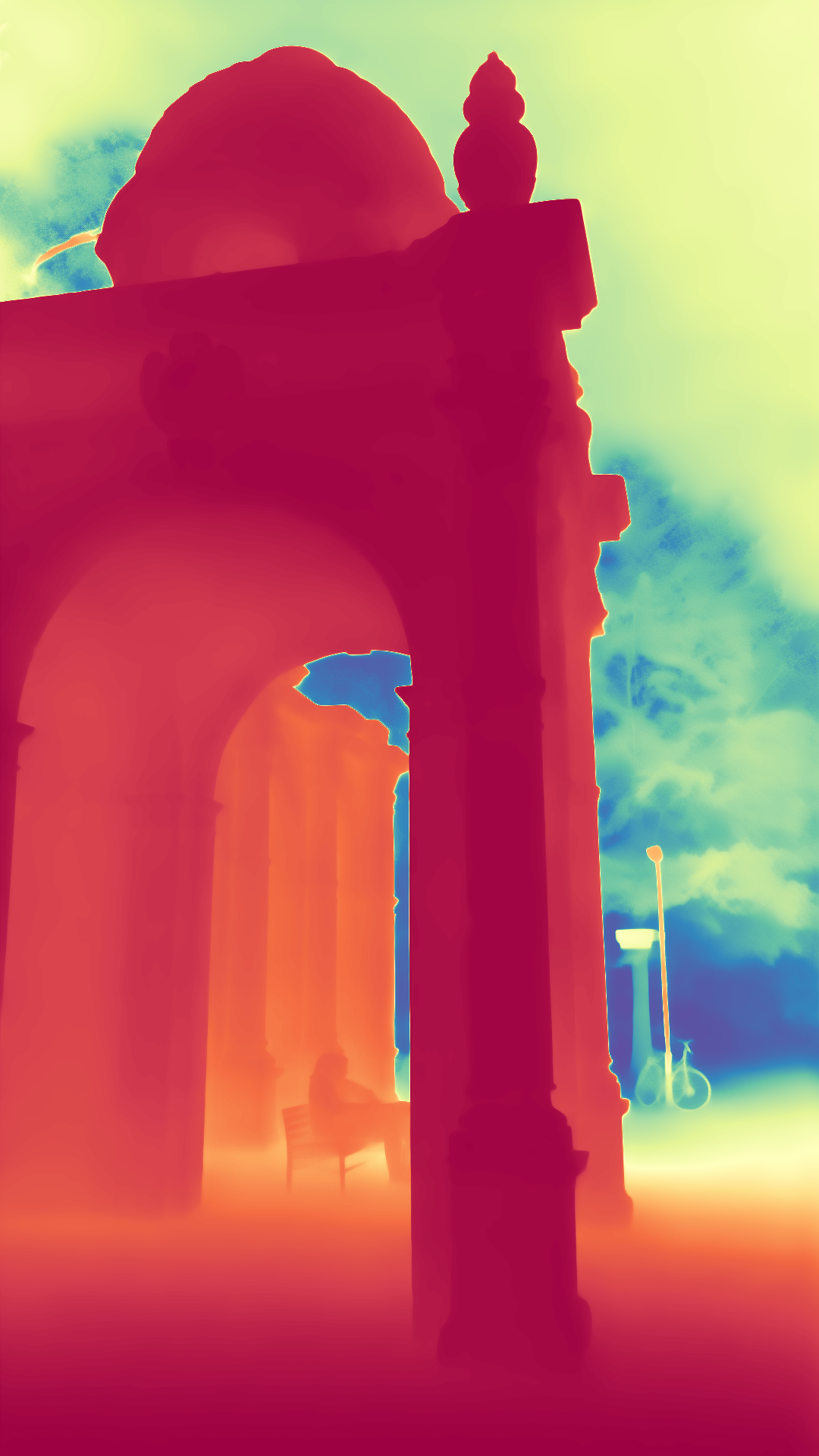} &
			\includegraphics[width=\sz\linewidth, height=3.0cm]{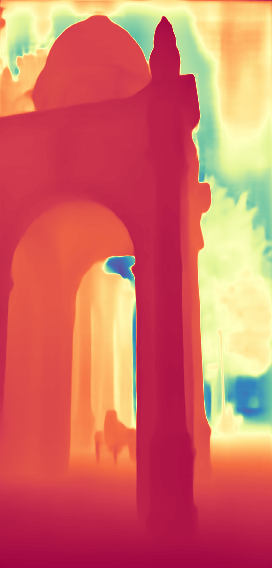} &
			\includegraphics[width=\sz\linewidth, height=3.0cm]{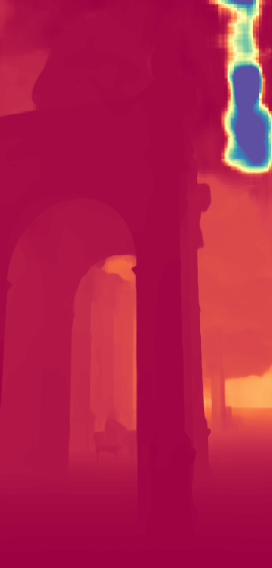} &
			\includegraphics[width=\sz\linewidth, height=3.0cm]{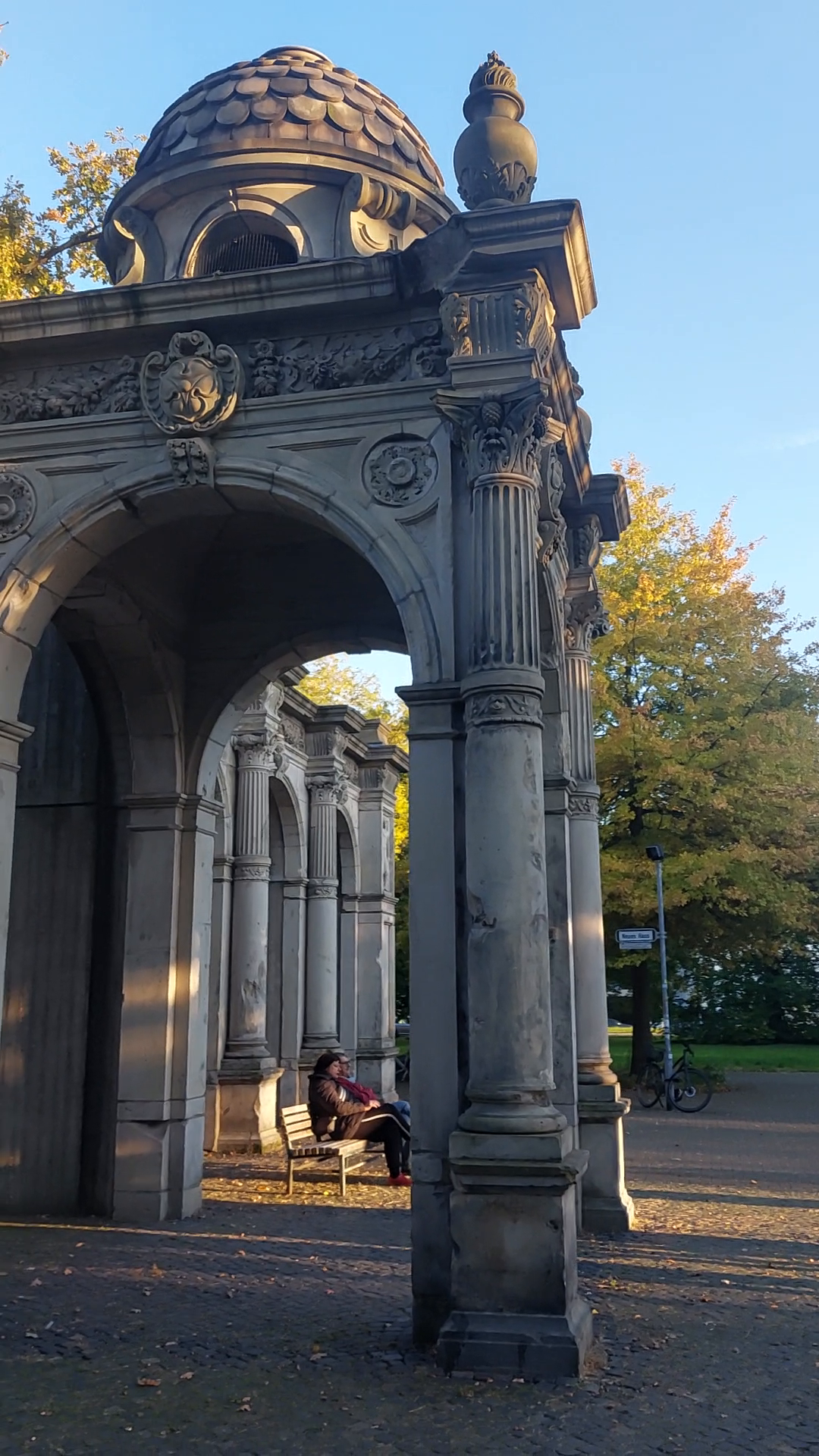} \\
			
			\includegraphics[width=\sz\linewidth, height=3.0cm]{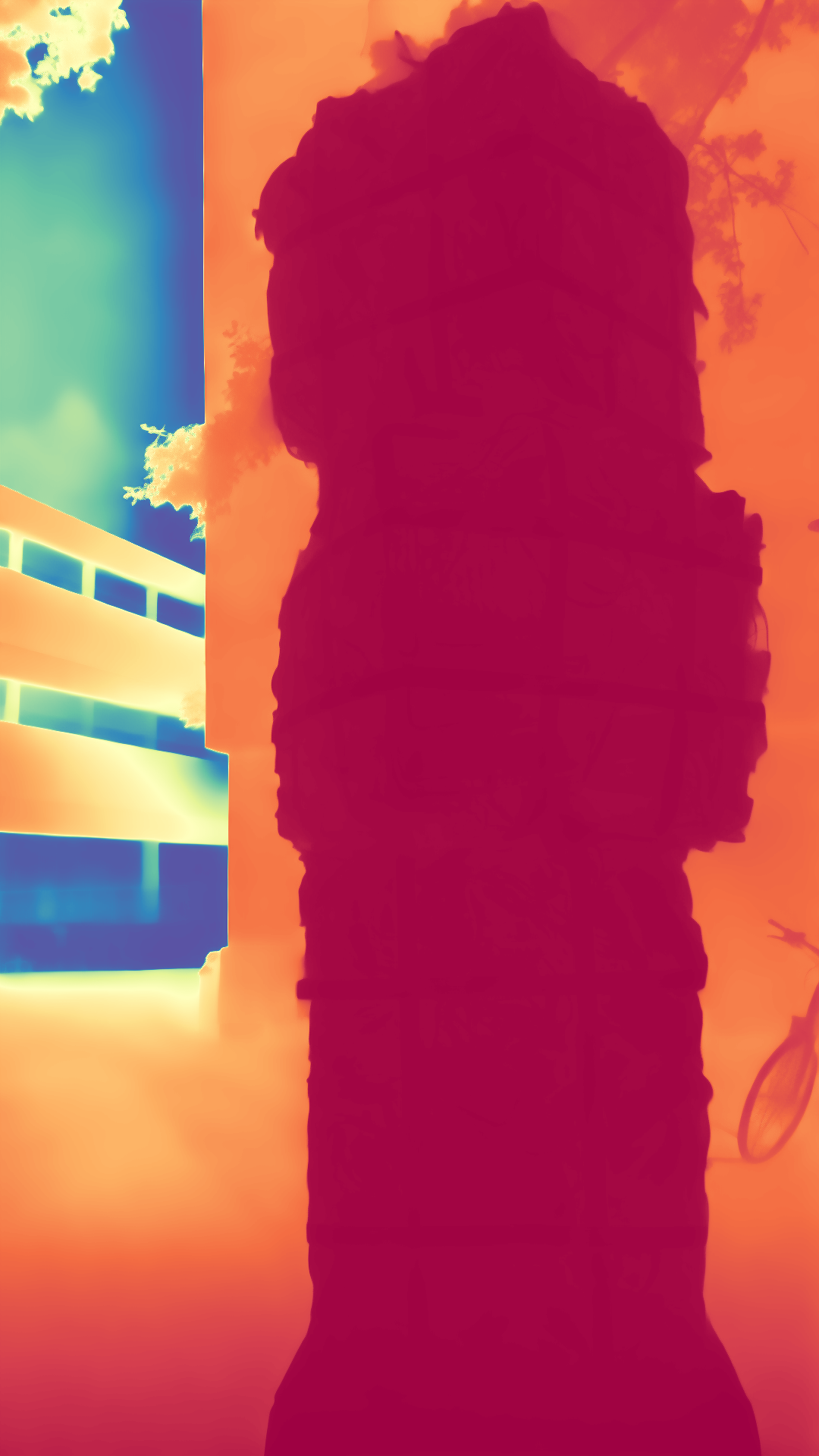} &
			\includegraphics[width=\sz\linewidth, height=3.0cm]{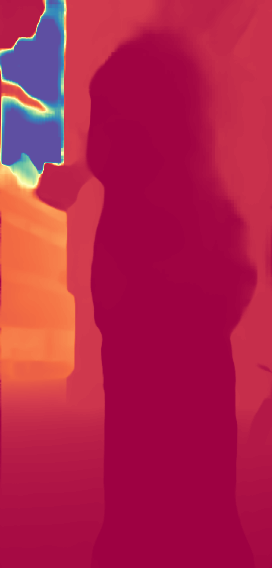} &
			\includegraphics[width=\sz\linewidth, height=3.0cm]{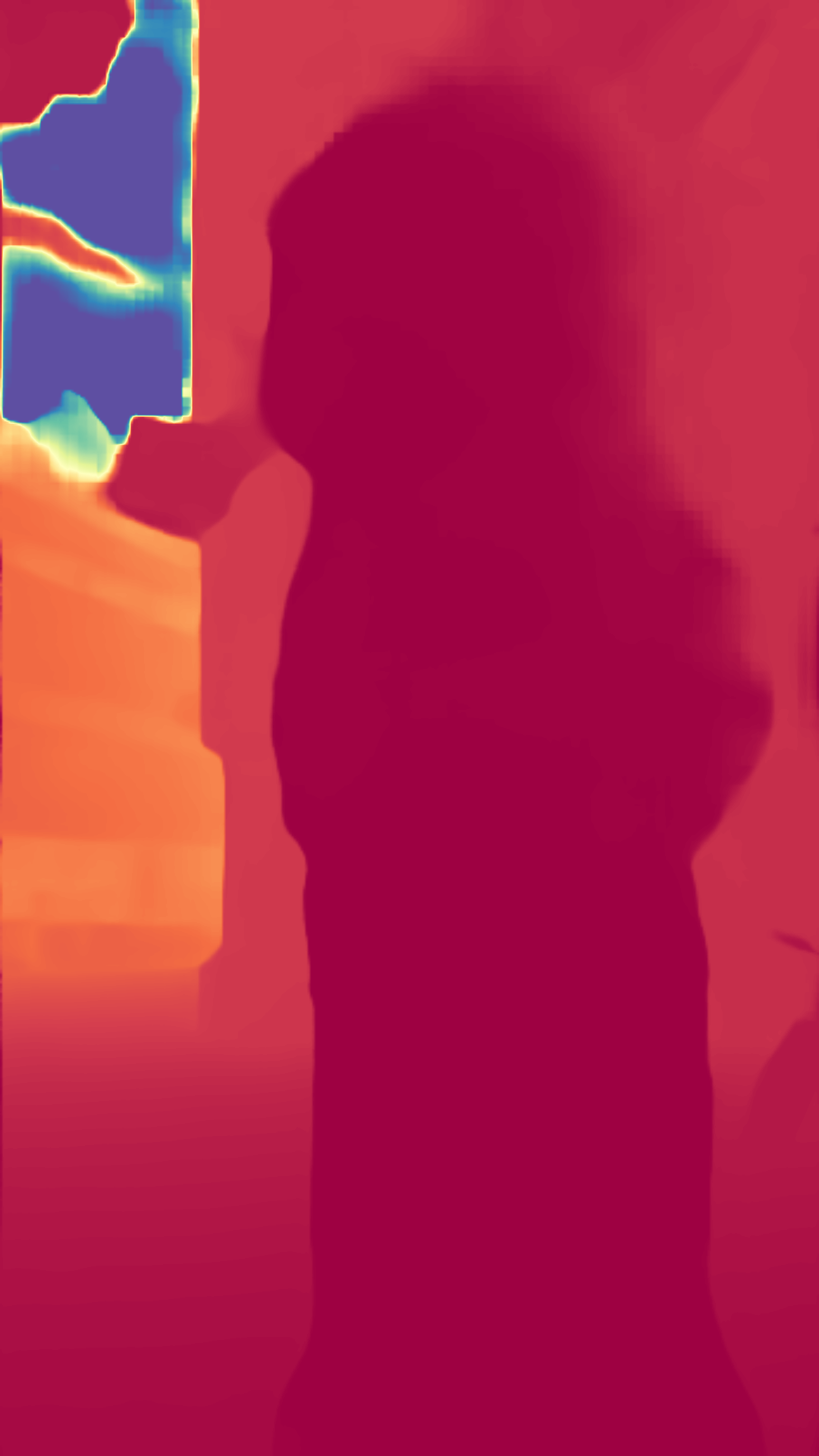} &
			\includegraphics[width=\sz\linewidth, height=3.0cm]{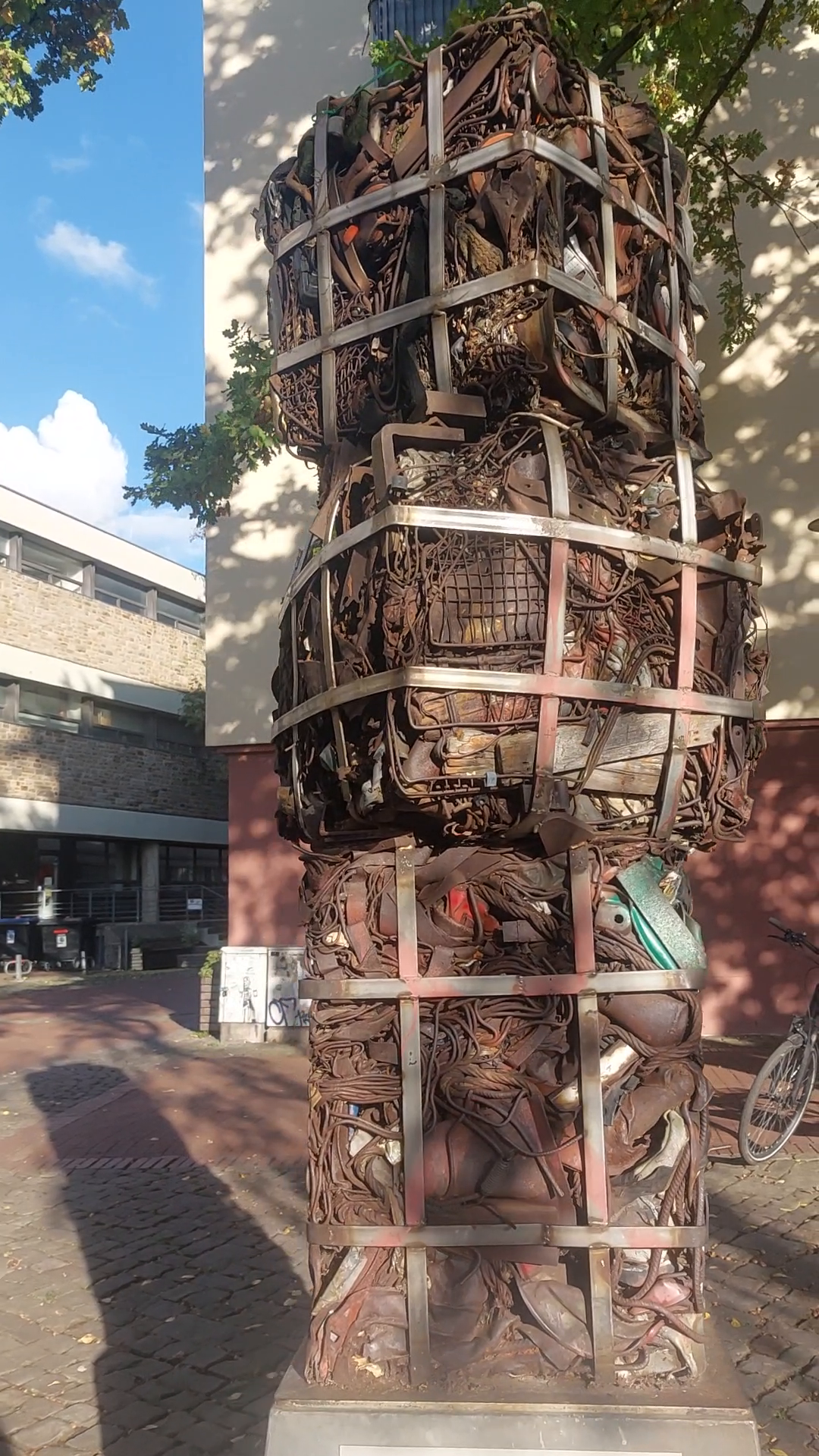} \\
			
			\includegraphics[width=\sz\linewidth, height=3.0cm]{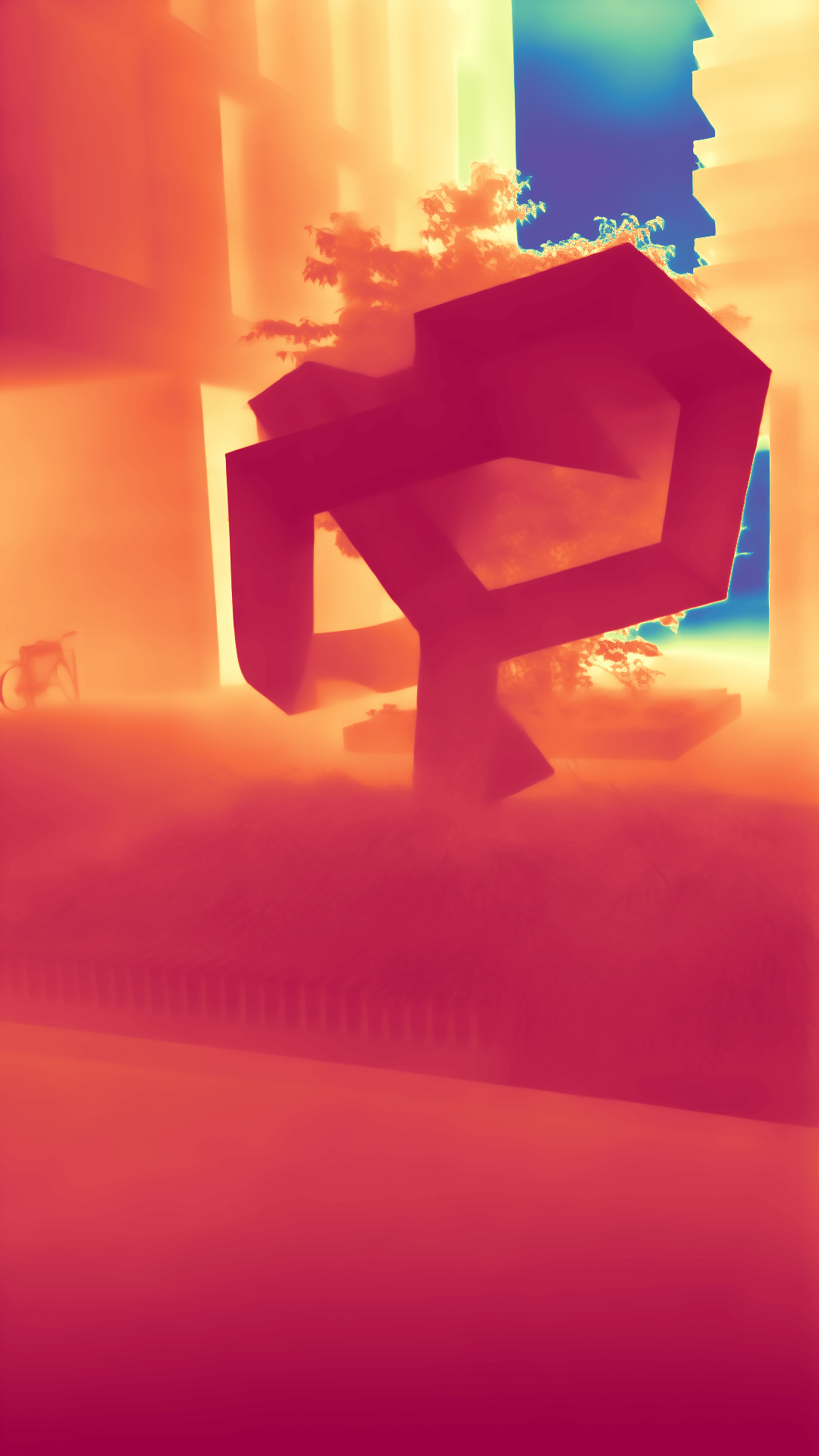} &
			\includegraphics[width=\sz\linewidth, height=3.0cm]{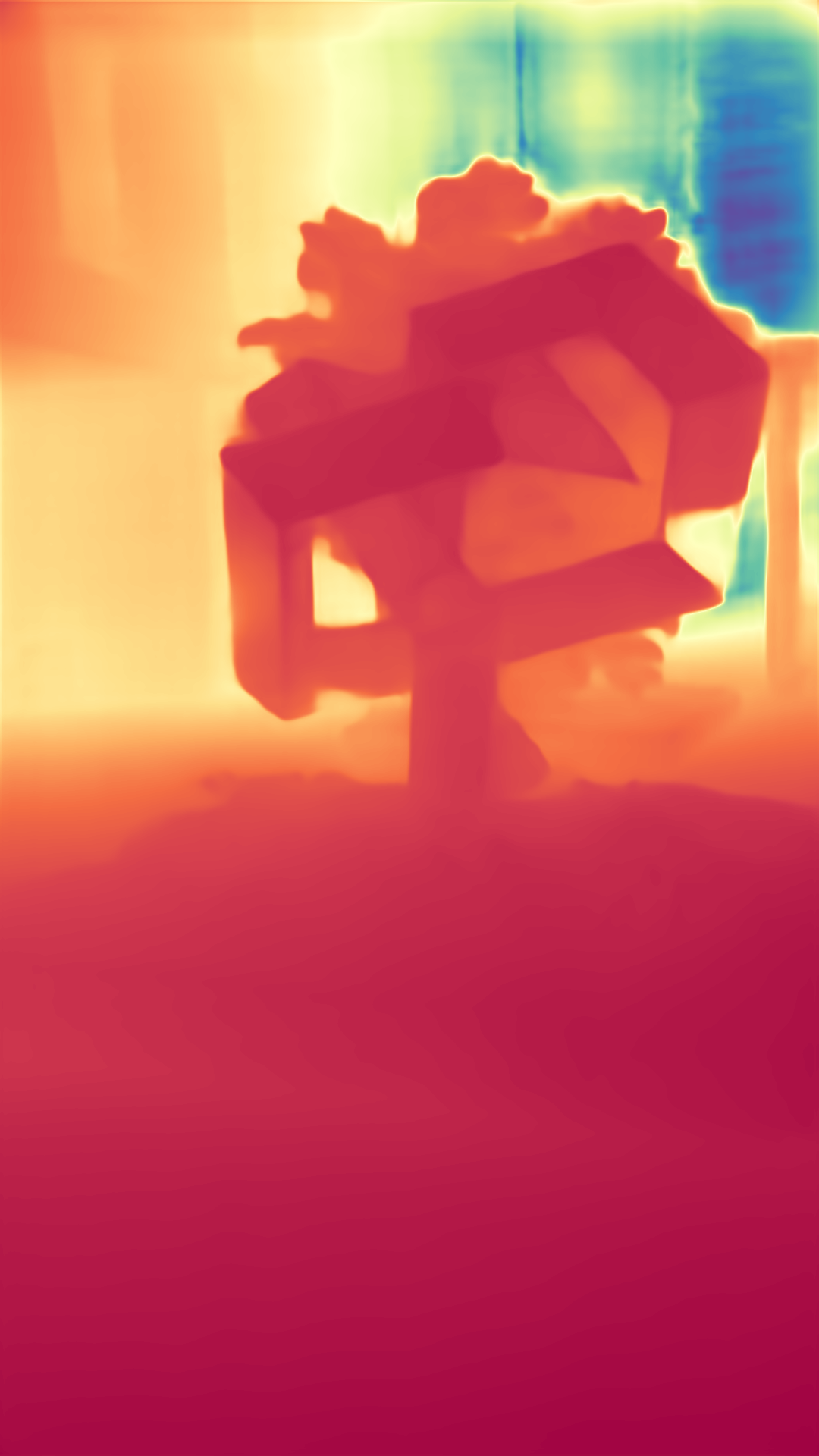} &
			\includegraphics[width=\sz\linewidth, height=3.0cm]{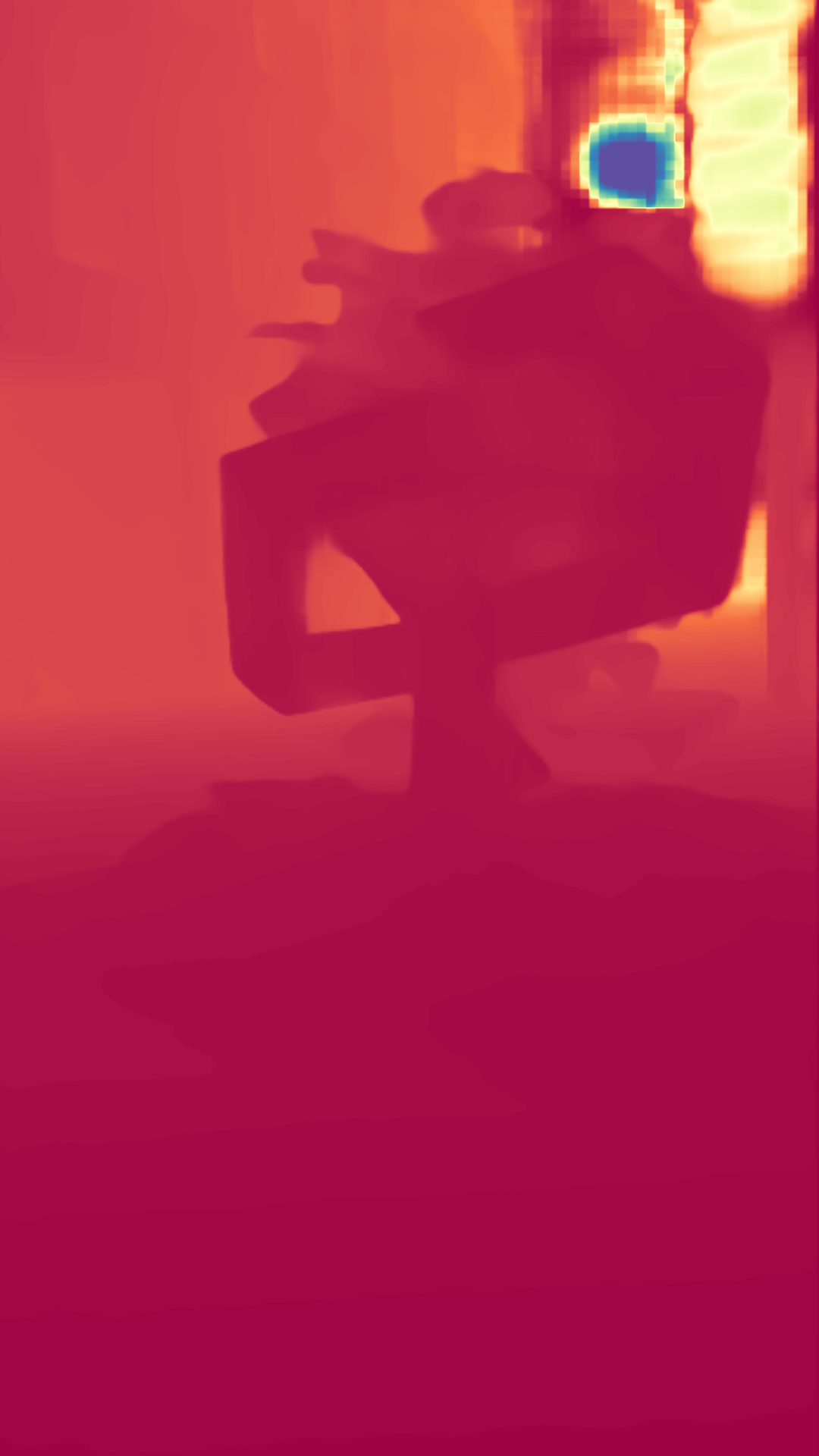} &
			\includegraphics[width=\sz\linewidth, height=3.0cm]{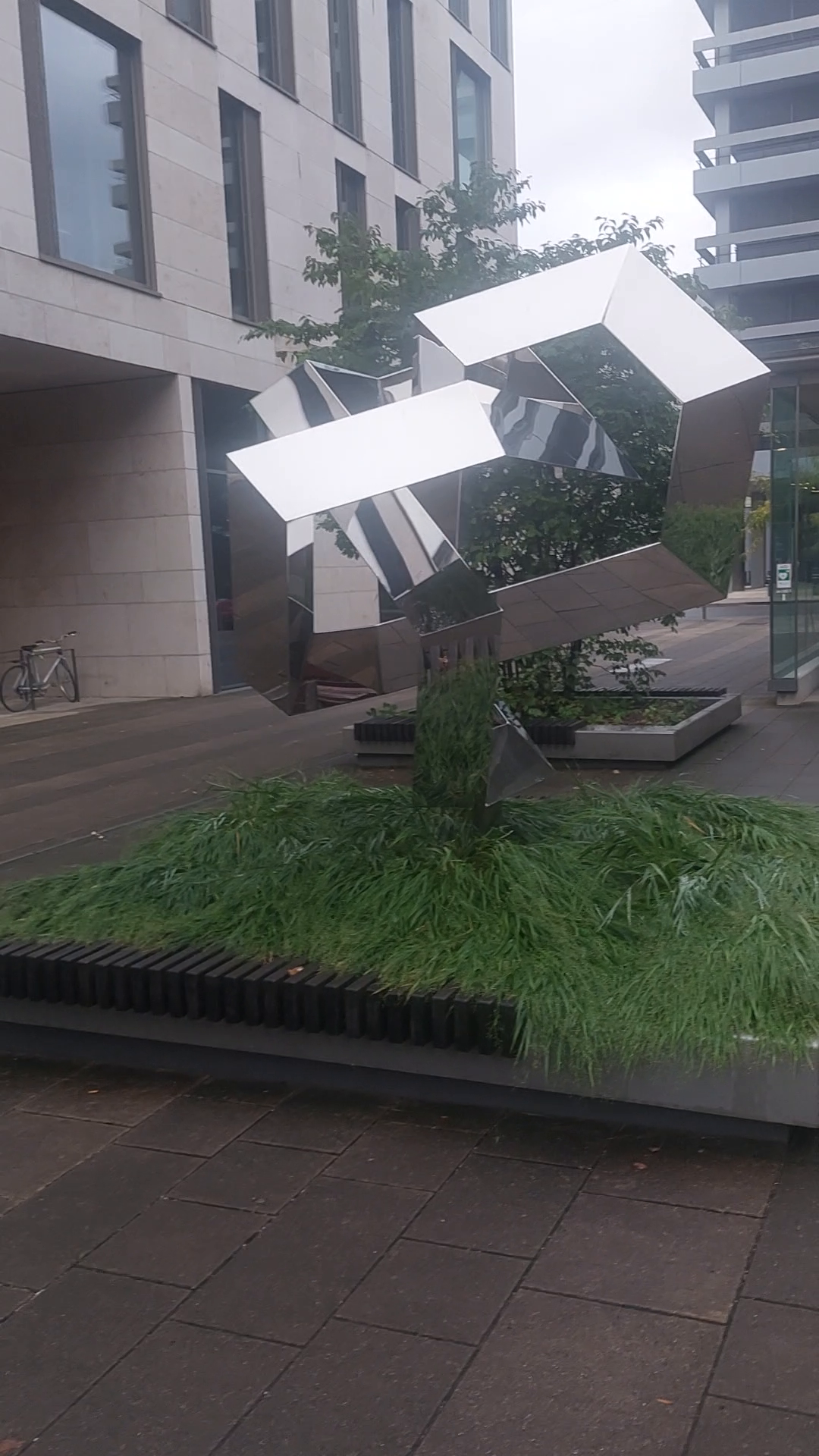} \\
			
			\includegraphics[width=\sz\linewidth, height=3.0cm]{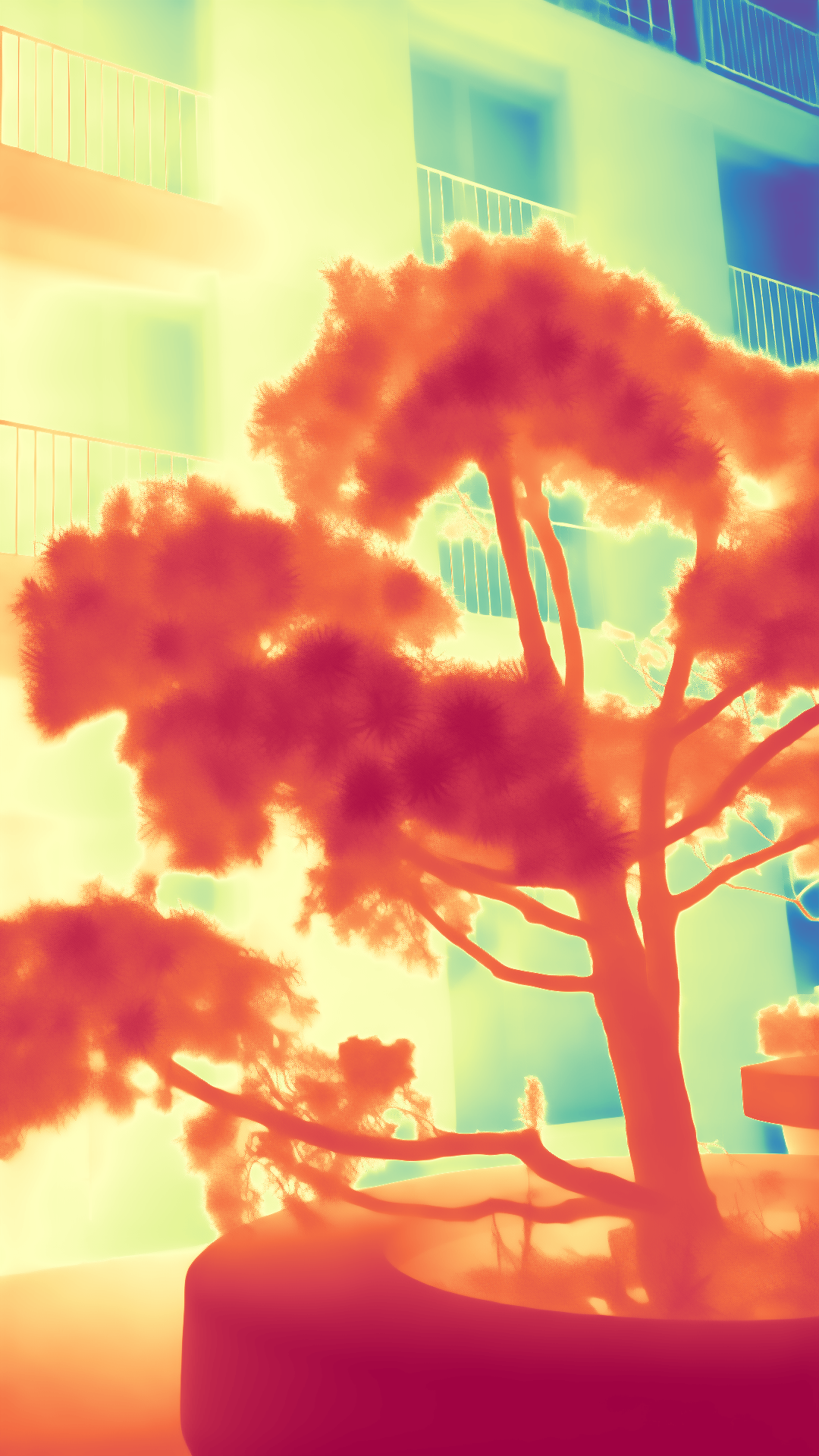} &
			\includegraphics[width=\sz\linewidth, height=3.0cm]{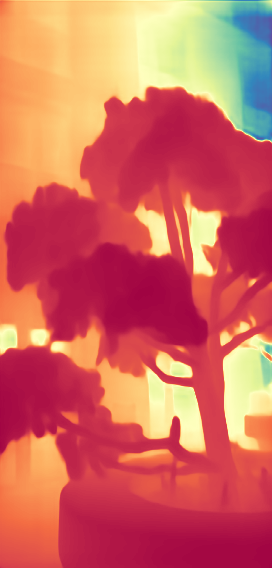} &
			\includegraphics[width=\sz\linewidth, height=3.0cm]{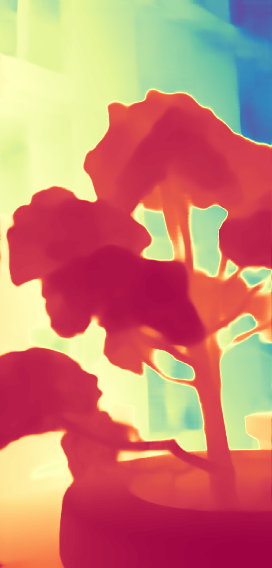} &
			\includegraphics[width=\sz\linewidth, height=3.0cm]{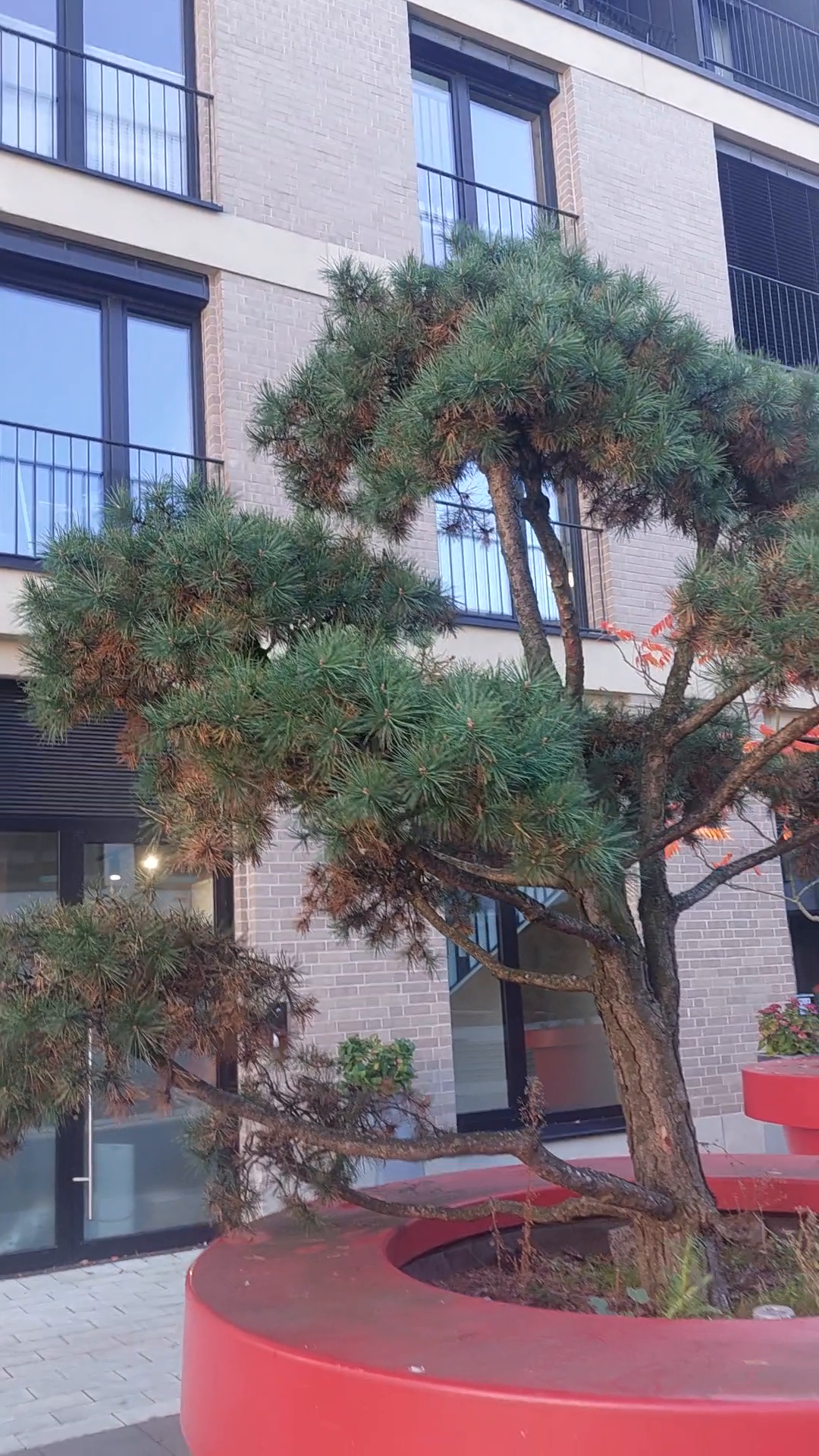} \\
			
			\includegraphics[width=\sz\linewidth, height=3.0cm]{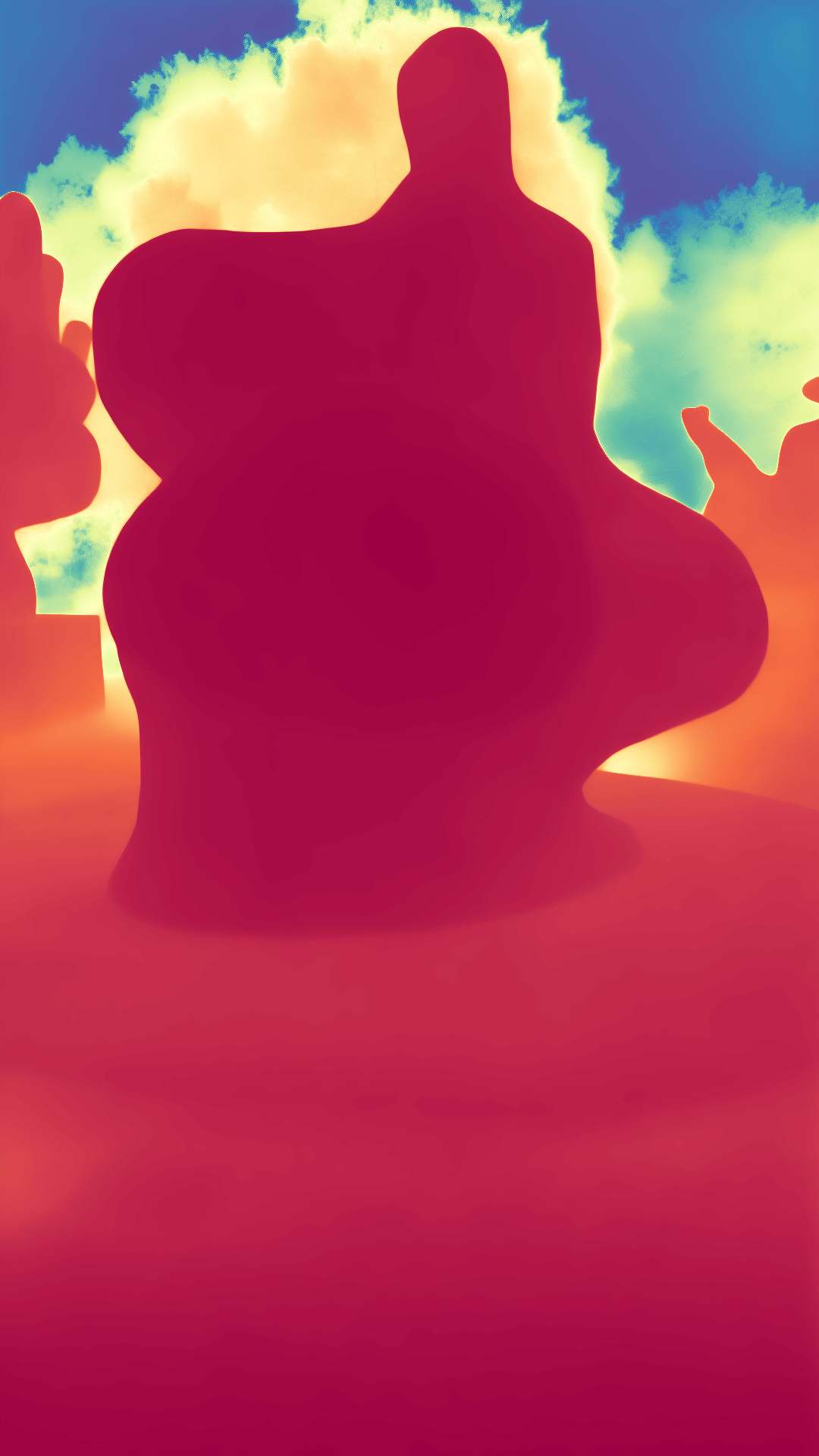} &
			\includegraphics[width=\sz\linewidth, height=3.0cm]{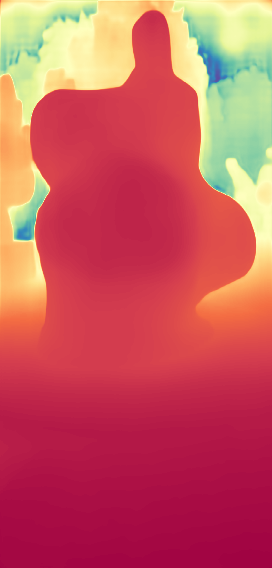} &
			\includegraphics[width=\sz\linewidth, height=3.0cm]{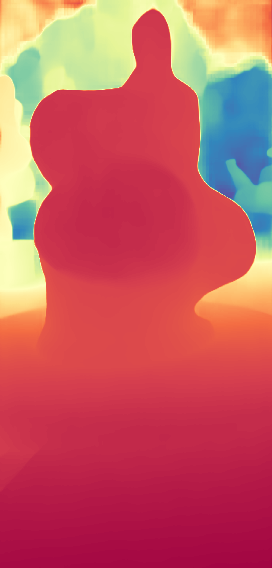} &
			\includegraphics[width=\sz\linewidth, height=3.0cm]{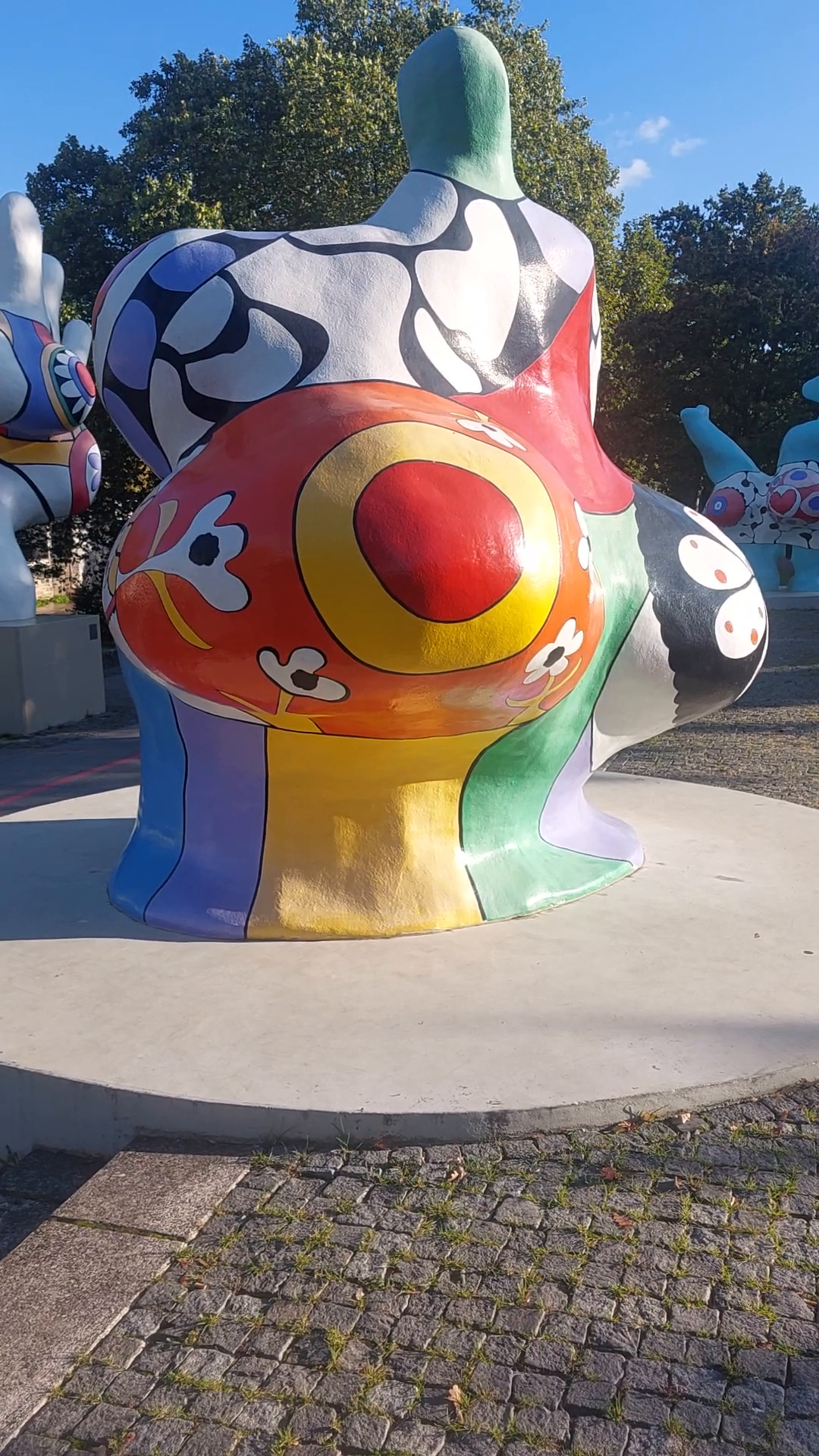} \\		
		\end{tabular}
	}
	\caption{\textbf{Monocular depth prediction in-the-wild}. Models exhibit very clear differences w.r.t captured details and scale consistency on self-captured video. Problems can arise in particular for reflective surfaces or paintings.}
	\vspace{0em}
	\label{fig:supp_depth}
\end{figure*}
\section{Monocular Depth Prediction}
\label{sup:prior}
Monocular depth prediction is a longstanding task with very impressive in-the-wild results of recent SotA models \cite{zoedepth, metric3d,depthanything, lotus-d}. We show some qualitative comparisons between selected models in Figure \ref{fig:supp_depth}. Due to training on massive datasets, current single-image depth predictions can recover fine-structured details. Nonetheless, the accuracy of rel. depth on a single frame is not the only thing that matters for SLAM. We want to highlight: 
\begin{table*}[h!]
	\centering
	\begin{tabular}{lcccccccc}
		\toprule
		Prior & ATE RMSE$\downarrow$ & PSNR$\uparrow$ & LPIPS$\downarrow$ & L1$\downarrow$ & ATE RMSE$\downarrow$ & PSNR$\uparrow$ & LPIPS$\downarrow$ & L1$\downarrow$ \\
		& \multicolumn{4}{c}{\cellcolor[HTML]{EEEEEE}{\textit{KF}}} & \multicolumn{4}{c}{\cellcolor[HTML]{EEEEEE}{\textit{Non-KF}}} \\ 
		\multicolumn{9}{l}{\textit{\textbf{TUM-RGBD}}} \\[2pt] 
		Metric3D \cite{metric3d} & \nd 1.93 & \fst 23.27 & \fst 0.226 & \fst 0.091 & \nd 1.83 & \fst 22.48 & \fst 0.242 & \fst 0.089 \\
		ZoeDepth \cite{zoedepth} & \rd 1.97 & \rd 23.21 & \rd 0.233 & \rd 0.132 & \rd 1.87 & \rd 22.34 & \rd 0.249 & \rd 0.136 \\
		DepthAnything \cite{depthanything} & \fst 1.91 & \nd 23.24 & \nd 0.229 & \nd 0.098 & \fst 1.79 & \nd 22.43 & \nd 0.246 & \nd 0.099 \\
		Lotus \cite{lotus-d} & 2.45 & 22.84 & 0.256 & 0.297 & 2.39 & 21.84 & 0.273 & 0.313 \\
		& & & & & & & & \\[-5pt]
		
		\multicolumn{9}{l}{\textit{\textbf{Replica}}} \\[2pt]
		Metric3D \cite{metric3d} & \rd 0.269 & \rd 32.92 & 0.134 & \fst 0.037 & \nd 0.268 & \rd 32.62 & 0.134 & \fst 0.038 \\
		ZoeDepth \cite{zoedepth} & \fst 0.266 & \fst 33.24 & \nd 0.123 & \rd 0.088 & \fst 0.265 & \fst 32.89 & \nd 0.123 & \rd 0.091 \\
		DepthAnything \cite{depthanything} & \nd 0.268 & \nd 33.06 & \rd 0.131 & \nd 0.063 & \nd 0.268 & \nd 32.73 & \rd 0.131 & \nd 0.066 \\
		Lotus \cite{lotus-d} & 0.275 & 32.23 & \fst 0.116 & 0.295 & \rd 0.278 & 31.72 & \fst 0.118 & 0.318 \\ 
		\bottomrule
	\end{tabular}	
	\caption{\textbf{Ablation Prior Depth on Replica~\cite{replica} and TUM-RGBD~\cite{tum-rgbd}}. Recent SotA depth prediction networks \cite{metric3d, zoedepth, depthanything, lotus-d} have different qualities for SLAM. Good temporal consistency allows accurate geometry reconstruction. However, rendering quality and tracking does not necessarily correlate with it. Results are after online mapping without any refinement using vanilla 3D Gaussian Splatting \cite{3dgs} and averaged over 5 runs. 
	}
	\label{tab:supp_ablation_depth}
\end{table*}
\begin{itemize}
	\item \textit{The rel. depth error on a single image should be minimal.} This is obvious, however most recent models are only evaluated on specific benchmarks such as e.g. KITTI~\cite{kitti} or NYU~\cite{nyu-depth}. Even though model predictions can look qualitatively very different, their abs. rel. error does not seem to be that different on untypical depth prediction benchmarks. 
	\item \textit{Temporal consistency matters a lot.} Even though we optimize scale $ \mathbf{s}_{i} $ and shift $\mathbf{o}_{i}$ parameters to match our perceived optical flow, models result in differently consistent integrated maps. It is still very beneficial to have high temporal scale consistency in a depth model. 
\end{itemize}\vspace{0.5em}

Recent diffusion models \cite{lotus-d, marigold} can leverage \textit{billion-scale} text-to-image pretraining to achieve strong depth prediction results with little finetuning. As can be seen in Figure \ref{fig:supp_depth}, the qualitative difference and recovered fine-structured details compared to models trained only on \textit{million-scale} depth prediction datasets seems obvious. However, diffusion models exhibit strong scale differences across a video. This seems to create a lot of floaters, in part enhanced due to the high-frequency details. We did not see an improvement for SLAM by integrating these models for this reason. Table \ref{tab:supp_ablation_depth} shows the performance of our system with vanilla 3D Gaussian Splatting \cite{3dgs}. We observe that \textit{Metric3D} \cite{metric3d} consistently optimizes the best geometry. However, other metrics are not always consistent.

\section{How important is camera calibration really?}
\label{sup:calib}
In this section we want to show some qualitative examples of in-the-wild footage with unknown intrinsics. As stated in the main paper, we perform a two-stage reconstruction: 
\begin{enumerate}
	\item Run the system without scale-optimization and optimize the camera intrinsics $\mathbf{\theta}$. 
	\item Use the now calibrated camera to run in \textit{P-RGBD} mode and additionally optimize $\mathbf{s}_{i}$ and $\mathbf{o}_{i}$
\end{enumerate}
Since we need an initial estimate of the intrinsics, we assume a heuristic where for a pinhole camera
\begin{align}
	fx = fy = \left( H + W \right) / 2 \nonumber \\
	cx = W / 2 \quad cy = H / 2 \quad .
\end{align}
\begin{figure*}[h]
	\centering
	\includegraphics[width=1.0\linewidth, clip]{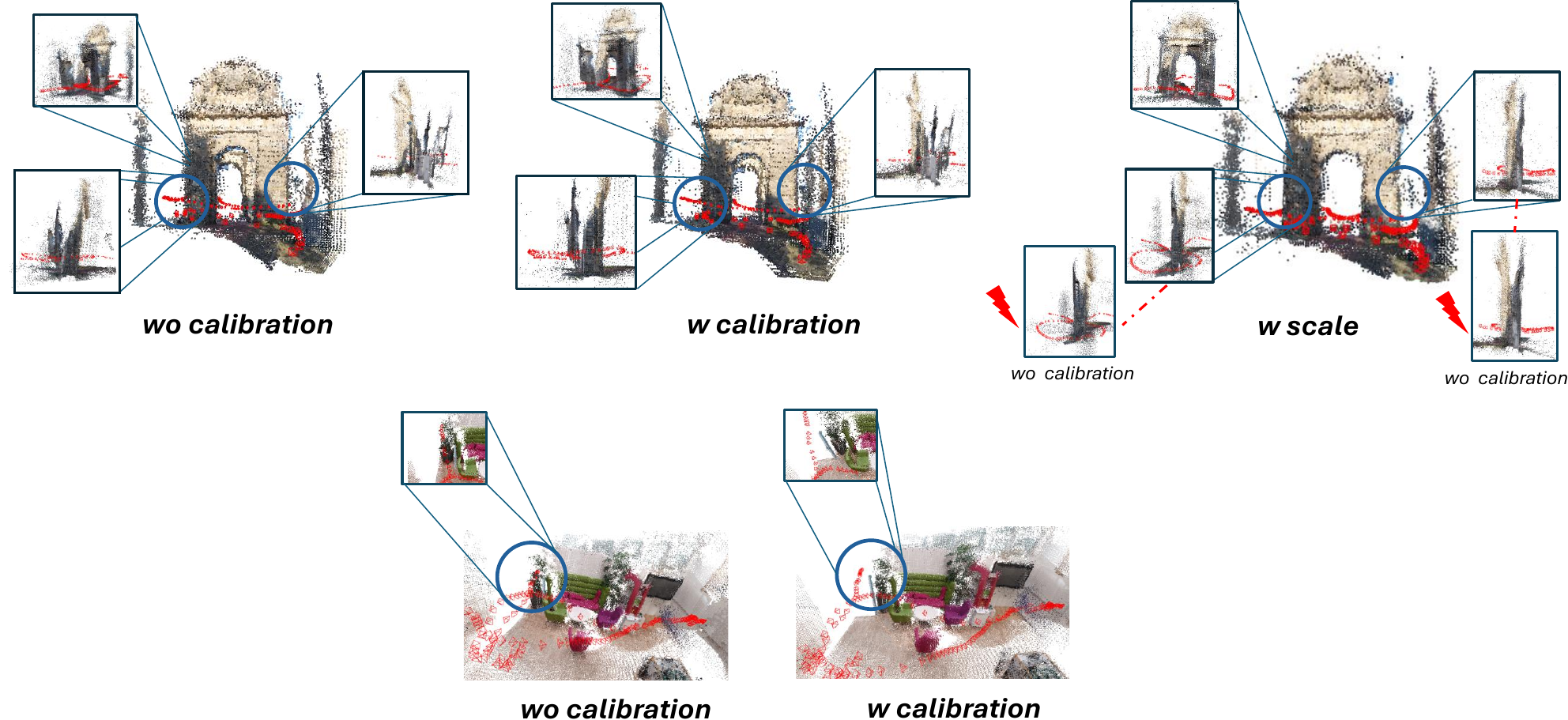}
	\vspace{2pt}
	\caption{
		\textbf{Camera calibration and prior integration matter.} Distortion effects and artifacts both on the map and camera odometry can be observed without calibration. Using our strategy, we can get rid of distortions. The scale-optimized prior integration allows accurate structure reconstruction. Outdoor scenes require all together due to scale inconsistencies of common depth prediction models.
	}
	\label{fig:supp_calib}
\end{figure*}
The benefit of camera calibration was quantitatively shown in \cite{droidcalib}. We report qualitative results on self-recorded scenes and show the robustness when initializing from a heuristic. It can be seen in Figure \ref{fig:supp_calib}, that both intrinsics calibration and scale optimization are beneficial for in-the-wild reconstruction. With wrong intrinsics, we observe distorted odometry and structure. With scale optimization, we can generate globally consistent maps. All together forms a good basis for rendering. 

\newpage
\section{Failure Cases}
\label{sup:fail}
Due to the challenging unbounded outdoor setting on uncalibrated cameras, we quickly observed common limitations of our framework. We notice that even though monocular depth prediction networks allow highly detailed single-frame predictions, their usage on in-the-wild video is limited. Scale inconsistencies and inaccurate predictions make us accumulate floaters over time. We therefore have to use the following: We limit depth supervision to consistent 3D points using the covisibility check \cite{droid} and pixels with confidence $ \mathbf{\sigma}_{i} \geq 0.1 $. This removes the sky and many floaters, but can also underconstraint the scene.   

\begin{figure}[h]
	\centering
	\includegraphics[width=\linewidth]{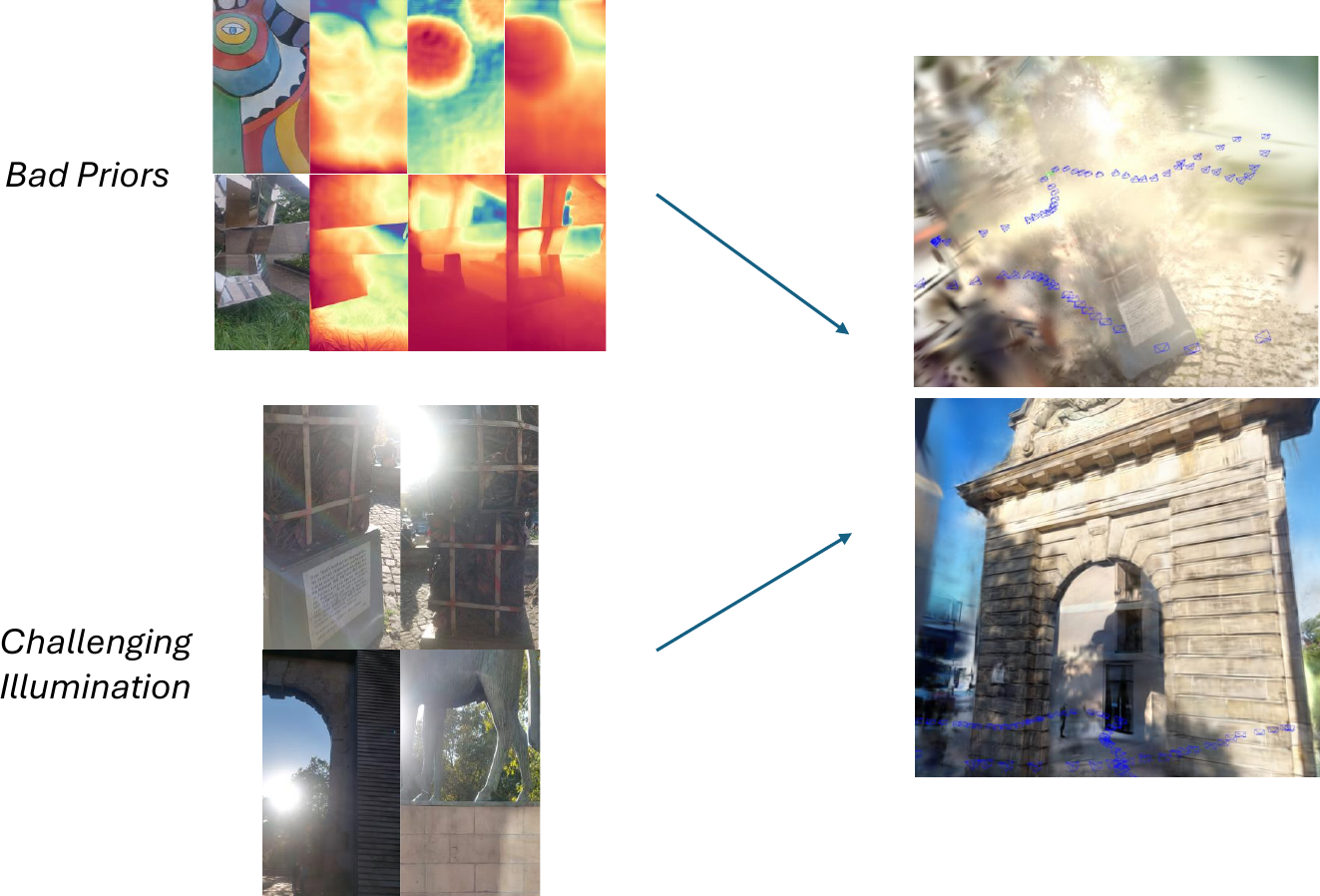}
	\caption{
		\textbf{Common failure cases}. Since we are heavily dependent on depth priors on in-the-wild video, our method can fail when priors get unreliable. Similarly, if the geometry supervision is not good enough, we accumulate floaters on outdoor scenes. Challenging lighting conditions can enhance this effect, since our model will overfit the scene and create additional Gaussians for modeling lighting effects (see Gaussians surrounding object).
	}
	\label{fig:fail}
\end{figure}

\section{What did not work?}
\label{sup:non-working}
We tried the following things unsuccessfully: 
\begin{itemize}
	\item Multi-View Gaussian Splatting \cite{mvgs} backprojects crops of 2D appearance error into 3D by using the camera ray. We can then perform an intersection test to carve out a 3D volume across multiple views. This test identified new Gaussians, that cause a high 2D error, but were not identified in the original densification strategy \cite{3dgs}. However, we did not manage to improve densification this way within our framework.
	\item \cite{hf-slam} uses a regularization term to battle catastrophic forgetting. We did not succeed on improving our metrics this way. We further tried to simply scale the gradients of optimized Gaussians by the number of times its frame has been already optimized by the renderer. 
	\item Sparse GS \cite{sparse-gs} uses a \textit{softmax} for rendering depths. We can identify floaters on outdoor scenes by analyzing the modality of the depth distribution. Since we compute an integrated absolute depth and supervise with priors, we were not able to converge quickly to the correct values due to the used \textit{logarithm} function. Since Sparse GS was created with rel. depth supervision, we did not pursue this further.
\end{itemize}

\newpage







\end{document}